\documentclass[lettersize,journal]{IEEEtran}
\usepackage{amsmath,amsfonts}
\usepackage{algorithmic}
\usepackage{algorithm}
\usepackage{array}
\usepackage{textcomp}
\usepackage{stfloats}
\usepackage{url}
\usepackage{verbatim}
\usepackage{graphicx}
\usepackage{subfigure}
\usepackage{cite}
\usepackage{multirow}
\usepackage{makecell}
\usepackage{authblk}
\usepackage{tcolorbox}
\tcbuselibrary{theorems}

\newtcolorbox[auto counter]{keyinsight}[1][]{
  title=Key Insight~\thetcbcounter,
  colback=gray!10,
  colframe=gray,
  #1
}
\begin{document}

\title{A Comprehensive Evaluation of Deep Learning Object Detection Models on Heterogeneous Edge Devices}

\author{
Daghash K. Alqahtani,
Muhammad Aamir Cheema,
Maria A. Rodriguez,
Adel N. Toosi
\thanks{Daghash K. Alqahtani, Maria A. Rodriguez, and Adel N. Toosi are with the DisNet Lab, School of Computing and Information Systems, The University of Melbourne, Melbourne, VIC, Australia (e-mail: daghash.alqahtani@student.unimelb.edu.au; marodriguez@unimelb.edu.au; adel.toosi@unimelb.edu.au).}
\thanks{Muhammad Aamir Cheema and Adel N. Toosi are also with Monash University, Melbourne, VIC, Australia (e-mail: aamir.cheema@monash.edu).}
\thanks{This research was partially supported by the Australian Research Council (ARC) through funded projects DP230100081 and LP210200213.}

}



\maketitle

\begin{abstract}

Object detection is a cornerstone of emerging edge applications, from autonomous vehicles and intelligent surveillance to smart city infrastructure, where inference must run on resource-constrained devices under strict latency and energy budgets. However, practitioners currently lack systematic, empirical guidance on how modern detection models navigate the critical three-way trade-off between accuracy, latency, and energy consumption across heterogeneous edge hardware, and existing benchmarks rarely consider how scene complexity affects accuracy in deployment. To close this gap, we present a comprehensive benchmark of eight representative detectors, YOLOv8 (Nano, Small, Medium), EfficientDet Lite (Lite0–Lite2), and SSD variants (SSD MobileNet V1, SSDLite MobileDet),  across nine edge platforms: Raspberry Pi 3/4/5 with and without Coral TPU accelerators, Raspberry Pi 5 with AI HAT+, Jetson Nano, and Jetson Orin Nano. Our study makes two key contributions: (i) a systematic characterization of the accuracy–latency–energy trade-off space across model–device combinations, jointly measured under identical conditions rather than in isolation; and (ii) an object-count-based analysis that quantifies how detection accuracy degrades with scene complexity, a dimension largely overlooked in prior edge benchmarking studies. Our results uncover significant, and sometimes counterintuitive, trade-offs: SSD MobileNet V1 delivers the lowest latency and energy consumption at the cost of accuracy; YOLOv8 Medium attains the highest accuracy but with substantial computational overhead; and TPU acceleration boosts the efficiency of SSD and EfficientDet Lite while degrading YOLOv8 accuracy. The Jetson Orin Nano provides the most favorable overall balance across model families. Notably, model accuracy converges on simple scenes but diverges sharply as object density increases, implying that model selection matters most precisely in the complex scenes typical of real deployments. Together, these findings map the practical design space for edge object detection and offer actionable guidance for selecting model–hardware combinations under real-world accuracy, latency, and energy constraints.

\end{abstract}

\begin{IEEEkeywords}
Object Detection Models, Performance Evaluation, Inference Time, Energy Efficiency, Accuracy, Edge.
\end{IEEEkeywords}

\section{Introduction}
 \label{sec:intro}

\IEEEPARstart{O}{bject} detection is a fundamental computer vision task that underpins a wide range of applications, including autonomous driving, intelligent surveillance, and smart city systems. Many of these applications must operate under strict latency and energy constraints, for example, autonomous vehicles must detect pedestrians, vehicles, and obstacles in real time to support safe navigation. Driven by advances in deep learning and the emergence of \textit{edge computing}, it is now increasingly feasible to run object detection directly on resource-constrained devices such as embedded boards, smart cameras, and IoT gateways, offering lower latency, reduced bandwidth usage, improved privacy, and greater resilience to network disruptions~\cite{Balasubramaniam2022}. However, practical deployment remains challenging: edge platforms differ substantially in compute capability, accelerator support, memory capacity, and energy characteristics, so selecting an appropriate model–device combination requires carefully navigating the three-way trade-off between detection accuracy, inference latency, and energy consumption.

Although many light-weight object detection models are now available for edge deployment, there is still limited understanding of how they navigate this trade-off  across heterogeneous edge devices and under varying scene complexity. Existing benchmarking studies typically report aggregate results over an entire dataset, without examining how image content, particularly the number of objects present in a scene, influences the balance among accuracy, latency, and energy consumption. In real-world IoT and edge intelligence applications, however, workloads are rarely uniform: some images may contain no relevant objects, while others may contain many, imposing very different processing demands. A more fine-grained evaluation is therefore needed to understand how model-device combinations behave under realistic and diverse visual workloads. Beyond guiding deployment decisions, such complexity-aware performance profiles can serve as a foundation for runtime decision making in edge systems. For example, energy-conscious request routing~\cite{alqahtani2026ecore} and multi-objective load balancing across heterogeneous detection nodes~\cite{alqahtani2026multi} could leverage per-complexity accuracy, latency, and energy characteristics to dispatch each inference request to the most suitable model–device pair.

In this paper, we present  the first comprehensive benchmarking study of representative deep learning object detection models to jointly evaluate the accuracy, latency, and energy consumption of representative deep learning object detection models across a diverse set of edge devices, while accounting for scene complexity. Our testbed include a diverse set of edge devices, including Raspberry Pi platforms, TPU-enabled Raspberry Pi configurations, Raspberry Pi 5 with AI HAT+, Jetson Nano, and Jetson Orin Nano. We evaluate widely used object detection model families, including \textit{You Only Look Once (YOLO)}~\cite{Redmon2016}, \textit{Single Shot MultiBox Detector (SSD)}~\cite{Liu2016}, and \textit{EfficientDet Lite}~\cite{Tan2020}. For each family, we consider representative variants: YOLOv8 (Nano, Small, Medium)~\cite{ultralytics2024}, SSD (SSD MobileNet V1 and SSDLite MobileDet), and EfficientDet (Lite0, Lite1, Lite2). These models are evaluated using key performance metrics such as energy consumption, inference time, and mean Average Precision (mAP). Furthermore, to capture the effect of object count as a proxy for scene complexity, we partition the evaluation images into five groups according to the number of objects present: 0, 1, 2, 3, and 4 or more objects. This grouping enables a finer-grained analysis of model accuracy and system behavior, providing deeper insight into how different model-device combinations respond to varying levels of image content complexity.

Our \textbf{key contributions} are summarized as follows:
\begin{itemize}
    \item \textbf{The first joint characterization of the accuracy--latency--energy 
    trade-off space for object detection at the edge.} Unlike prior studies that 
    report these metrics in isolation or on a single platform, we measure all three 
    dimensions under identical conditions for eight representative detectors 
    (\textit{YOLOv8}, \textit{SSD}, and \textit{EfficientDet Lite} variants) across 
    nine heterogeneous edge platforms and four inference frameworks 
    (\textit{PyTorch}, \textit{TensorFlow Lite}, \textit{Hailo-8}, and 
    \textit{TensorRT}), yielding a directly comparable map of the practical 
    design space.
    
    \item \textbf{A novel scene-complexity-aware evaluation methodology.} We 
    introduce an object-count-based analysis that partitions evaluation images by 
    the number of objects present, a dimension largely overlooked in prior edge 
    benchmarking, and show that model accuracy converges on simple scenes but 
    diverges sharply as object density increases, implying that model selection 
    matters most in precisely the complex scenes typical of real deployments.
    
    \item \textbf{New empirical insights into hardware acceleration effects.} Our 
    results reveal counterintuitive interactions between models and accelerators; 
    most notably, Edge TPU acceleration substantially improves the efficiency of 
    SSD and EfficientDet Lite while \emph{degrading} YOLOv8 accuracy, and the 
    Jetson Orin Nano offers the most favorable overall balance across model 
    families.
    
    \item \textbf{Actionable deployment guidance and a reproducible benchmarking 
    pipeline.} We distill our findings into practical recommendations for selecting 
    model--device combinations under real-world accuracy, latency, and energy 
    constraints, and release an automated, service-based benchmarking environment 
    that enables consistent and extensible evaluation on future models and devices.
\end{itemize}

The remainder of this paper is organized as follows. 
Section~\ref{sec:background} provides background on the object detection model 
architectures and edge hardware platforms considered in this study. 
Section~\ref{sec:evaluation methodology} describes the evaluation methodology, 
including the evaluation metrics and experimental setup. 
Section~\ref{sec:evaluation results} reports the performance evaluation results. 
Section~\ref{sec:related work} reviews the related literature. 
Finally, Section~\ref{sec:conclusions} concludes the paper and outlines 
directions for future work.

\section{Background}
\label{sec:background}

This section provides background on the object detection models and edge hardware
platforms considered in this study. It first describes the main architectural
characteristics of SSD-based detectors, EfficientDet Lite, and YOLOv8, followed
by an overview of CPU-, TPU-, NPU-, and GPU-based edge platforms for on-device
inference.

\subsection{Object Detection Model Architectures}
\label{subsec:model_architectures}

The evaluated models represent three families with distinct architectural trade-offs between detection accuracy and computational efficiency. They differ mainly in two aspects: the detection scheme, i.e., whether predictions are made relative to \textit{anchors} (predefined reference boxes of various scales and aspect ratios) or directly in an anchor-free manner, and the mechanism for multi-scale feature fusion.

\textit{SSD-based detectors} originate from the Single Shot MultiBox Detector
framework, which predicts class scores and bounding-box offsets from multi-scale
feature maps using predefined default boxes~\cite{Liu2016}. SSD MobileNet V1
combines SSD with a MobileNet backbone based on depthwise separable
convolutions~\cite{howard2017mobilenets}. SSDLite MobileDet combines an
SSDLite lightweight detection head~\cite{sandler2018mobilenetv2} with the
MobileDet architecture for efficient mobile object detection~\cite{xiong2021mobiledets}.

\textit{EfficientDet Lite models} are mobile variants of EfficientDet, which 
combines an EfficientNet backbone with a bidirectional feature pyramid network 
(BiFPN) for weighted multi-scale feature fusion, and uses compound scaling to 
jointly scale resolution, backbone, feature network, and prediction 
heads~\cite{Tan2020}, yielding a spectrum of accuracy--efficiency operating 
points (Lite0--Lite2).

\textit{YOLOv8} also detects in a single forward pass but, unlike the two 
anchor-based families above, employs an anchor-free, decoupled head on top of 
its backbone--neck feature extractor~\cite{ultralytics2024}, offering the 
highest accuracy at correspondingly higher computational cost.

\subsection{Edge Hardware Platforms}

The evaluated edge platforms use different architectures for on-device inference.
\textit{CPU-only Raspberry Pi devices} rely on general-purpose ARM processors,
including Cortex-A53, Cortex-A72, and Cortex-A76 cores across different
generations\footnote{\url{https://www.raspberrypi.com/}}. These CPUs support
flexible model execution, but their limited parallel tensor-processing capability
can increase inference time and energy consumption for larger detectors.

In contrast, \textit{Coral USB accelerators} use Google's Edge TPU, a dedicated
ASIC optimized for low-power TensorFlow Lite inference and quantized execution%
\footnote{\url{https://gweb-coral-full.uc.r.appspot.com/docs/accelerator/datasheet/}}.
This makes them efficient for compatible models, but their performance depends
on whether the model operators, input format, and numerical precision are
supported by the Edge TPU compiler.

\textit{Raspberry Pi AI HAT+} uses a Hailo neural-network accelerator for local
AI inference on Raspberry Pi 5%
\footnote{\url{https://www.raspberrypi.com/documentation/accessories/ai-hat-plus.html}}.
Hailo accelerators use a structure-defined dataflow architecture and on-chip
memory to reduce data movement and support efficient edge AI inference.

\textit{NVIDIA Jetson devices} combine ARM CPUs with embedded GPUs: Jetson Nano
uses a Maxwell GPU with 128 CUDA cores, while Jetson Orin Nano uses a newer
Ampere GPU with CUDA and Tensor cores%
\footnote{\url{https://www.nvidia.com/en-au/autonomous-machines/embedded-systems/}}. With TensorRT optimizations such as
reduced precision, layer fusion, and kernel tuning, Jetson platforms provide
flexible acceleration for larger and more computationally demanding object
detection models.

\section{Evaluation Methodology}
\label{sec:evaluation methodology}

This section outlines the methodology used to evaluate the investigated object detection models on the selected edge devices. It introduces the evaluation framework and the criteria used to assess performance, including accuracy, inference time, energy consumption, and object-count-based analysis.

\subsection{Metrics}

\subsubsection{Inference Time}
Inference time measures the time required by a model service to process an input image and generate the detection response. In our setup, this time is measured on the target edge device from the moment the image is received by the object detection service to the moment the response is produced and returned by the service. Thus, the reported inference time captures the server-side processing time on the edge device, including model execution and the associated response generation within the service, but excluding network transmission delay between the client and the edge device. Inference time is reported in milliseconds for each model-device combination and is averaged over the processed requests to ensure consistent and reliable measurement.

\subsubsection{Energy Consumption}
Energy consumption is used to evaluate the energy efficiency of each model on the investigated edge devices. We first measure the base energy consumption, denoted by $BE$, while the device remains idle for a fixed duration of 5 minutes without executing any workload. We then measure the total energy consumption, denoted by $TE$, over the same duration while the object detection model is running.

The energy consumption per request, excluding base energy, denoted by $E_{\mathrm{excR}}$, is computed in milliwatt-hours (mWh) as
\begin{equation}
E_{\mathrm{excR}} = \frac{TE - BE}{NR},
\end{equation}
where $NR$ denotes the number of processed requests. Reporting energy on a per-request basis enables a fairer comparison across heterogeneous platforms, since total energy usage depends on the number of completed requests. This metric is particularly important for power-constrained and battery-powered edge devices.

\subsubsection{Model Evaluation Using the COCO Dataset}
To evaluate the detection capability of the investigated \textit{YOLOv8}, \textit{EfficientDet Lite}, and \textit{SSD} models, we use the COCO validation dataset containing 5,000 images~\cite{Lin2014}. The open source FiftyOne tool~\footnote{https://docs.voxel51.com/} is employed for dataset access, result visualization, and model evaluation using COCO annotations. It measures detection performance by comparing the predicted bounding boxes and class labels against the ground truth annotations. We report standard accuracy metric \textit{mAP}. Furthermore, we perform an object-count-based analysis by partitioning images into five groups according to the number of objects present in each image: 0, 1, 2, 3, and 4 or more objects. This grouping allows us to examine how accuracy varies with scene complexity and provides deeper insight into the behavior of different model-device combinations under diverse visual workloads.

\subsection{Experimental Setup}

\begin{table}[t]
\caption{Edge devices, models, input sizes, and deployment stack.}
\label{table:devices-models-frameworks}
\scriptsize
\centering
\setlength{\tabcolsep}{2pt}
\renewcommand{\arraystretch}{0.9}
\resizebox{\columnwidth}{!}{%
\begin{tabular}{|c|c|c|l|c|}
\hline
\textbf{Device (RAM)} & \textbf{Model} & \textbf{Size} & \textbf{Input} & \textbf{Stack} \\
\hline
\multirow{8}{*}{\makecell{Pi3 B+ (1GB) \\ Pi4 B (4GB) \\ Pi5 (4GB)}}
& Det-Lite0 & 5.7 MB & $320^2{\times}3$ & TFLite \\
& Det-Lite1 & 7.6 MB & $384^2{\times}3$ & TFLite \\
& Det-Lite2 & 10.2 MB & $448^2{\times}3$ & TFLite \\
& SSD-V1 & 7.0 MB & $300^2{\times}3$ & TFLite \\
& SSDLite & 5.1 MB & $320^2{\times}3$ & TFLite \\
& YOLOv8n & 6.5 MB & $640^2{\times}3$ & PyTorch \\
& YOLOv8s & 22.6 MB & $640^2{\times}3$ & PyTorch \\
& YOLOv8m & 52.1 MB & $640^2{\times}3$ & PyTorch \\
\hline
\multirow{8}{*}{\makecell{Pi3 B+ + TPU (1GB) \\ Pi4 B + TPU (4GB) \\ Pi5 + TPU (4GB)}}
& Det-Lite0 & 6.0 MB & $320^2{\times}3$ & TFLite \\
& Det-Lite1 & 8.0 MB & $384^2{\times}3$ & TFLite \\
& Det-Lite2 & 10.7 MB & $448^2{\times}3$ & TFLite \\
& SSD-V1 & 7.4 MB & $300^2{\times}3$ & TFLite \\
& SSDLite & 5.4 MB & $320^2{\times}3$ & TFLite \\
& YOLOv8n & 3.9 MB & $320^2{\times}3$ & TFLite \\
& YOLOv8s & 11.6 MB & $320^2{\times}3$ & TFLite \\
& YOLOv8m & 27.6 MB & $320^2{\times}3$ & TFLite \\
\hline
\multirow{8}{*}{Pi5 + AI HAT+ (4GB)}
& Det-Lite0 & 9.4 MB & $320^2{\times}3$ & HailoRT \\
& Det-Lite1 & 11.4 MB & $384^2{\times}3$ & HailoRT \\
& Det-Lite2 & 14.0 MB & $448^2{\times}3$ & HailoRT \\
& SSD-V1 & 7.0 MB & $300^2{\times}3$ & HailoRT \\
& SSDLite & 6.8 MB & $320^2{\times}3$ & HailoRT \\
& YOLOv8n & 5.2 MB & $640^2{\times}3$ & HailoRT \\
& YOLOv8s & 10.4 MB & $640^2{\times}3$ & HailoRT \\
& YOLOv8m & 27.6 MB & $640^2{\times}3$ & HailoRT \\
\hline
\multirow{8}{*}{\makecell{Jetson Nano (4GB) \\ Jetson Orin Nano (4GB)}}
& Det-Lite0 & 15.4 MB & $320^2{\times}3$ & TensorRT \\
& Det-Lite1 & 19.8 MB & $384^2{\times}3$ & TensorRT \\
& Det-Lite2 & 24.9 MB & $448^2{\times}3$ & TensorRT \\
& SSD-V1 & 14.7 MB & $300^2{\times}3$ & TensorRT \\
& SSDLite & 10.8 MB & $320^2{\times}3$ & TensorRT \\
& YOLOv8n & 14.2 MB & $640^2{\times}3$ & TensorRT \\
& YOLOv8s & 46.1 MB & $640^2{\times}3$ & TensorRT \\
& YOLOv8m & 105.3 MB & $640^2{\times}3$ & TensorRT \\
\hline
\end{tabular}}
\end{table}

\subsubsection{Hardware and Device Setup}
Our experimental setup includes a diverse set of edge devices for evaluating object detection performance, as summarized in Table~\ref{table:devices-models-frameworks}. The evaluated platforms include Raspberry Pi 3 Model B+, Raspberry Pi 4 Model B, Raspberry Pi 5, Raspberry Pi 3 with TPU, Raspberry Pi 4 with TPU, Raspberry Pi 5 with TPU, Raspberry Pi 5 with AI HAT+, Jetson Nano, and Jetson Orin Nano. These devices were selected to represent a broad range of edge computing configurations with different levels of computational capability and hardware acceleration. In particular, the setup includes CPU-only platforms, TPU-enabled Raspberry Pi platforms, Raspberry Pi 5 with a dedicated NPU-based accelerator, and NVIDIA GPU-based edge devices. This diversity enables a comprehensive comparison of object detection performance across heterogeneous model-device combinations. In addition, a UM25C\footnote{https://joy-it.net/en/products/JT-UM25C} power meter with Bluetooth connectivity is used to measure the energy consumption of each device.

\subsubsection{Software and Deployment Stack}

In our experimental setup, we employ different deployment stacks to execute object detection models on the investigated edge devices, as summarized in Table~\ref{table:devices-models-frameworks}. The choice of framework or runtime is determined by the capabilities of each target platform and the need to obtain efficient inference performance across CPU-based, TPU-enabled, AI accelerator-based, and GPU-based devices. 
For Raspberry Pi 3, Raspberry Pi 4, and Raspberry Pi 5, \textit{PyTorch} is used to run YOLOv8 models, while \textit{TensorFlow Lite} is used to deploy EfficientDet Lite and SSD models. To exploit Edge TPU acceleration, the YOLOv8 models are converted from PyTorch to TensorFlow Lite format and then compiled for execution on Raspberry Pi devices equipped with Coral USB Accelerators. In this configuration, the YOLOv8 models use an input size of 320 $\times$ 320 rather than 640 $\times$ 640 in order to fit the constraints of TPU-based deployment. EfficientDet Lite and SSD models are also deployed in TensorFlow Lite format and compiled for execution on Edge TPU. For Raspberry Pi 5 with AI HAT+, the evaluated models are deployed using the Hailo software stack to execute inference on the Hailo-8 accelerator. For both Jetson Nano and Jetson Orin Nano, the models are converted and deployed using \textit{TensorRT} to enable optimized inference on NVIDIA GPUs.

The operating systems used in the experiments are Raspberry Pi OS Bookworm 64-bit for the Raspberry Pi-based devices and Jetson Linux, which is Ubuntu-based, for Jetson Nano and Jetson Orin Nano. Additional software tools, including OpenCV for image processing and FiftyOne for model evaluation, are used to facilitate the experiments. To perform object detection, we implement each model as a Python-based service using the Flask RESTful library and expose it through an API endpoint. This design enables consistent integration, automation, and testing across different edge platforms.

\subsubsection{Experimental Procedure}

\begin{figure}[t]
    \centering
    \includegraphics[width=1\linewidth]{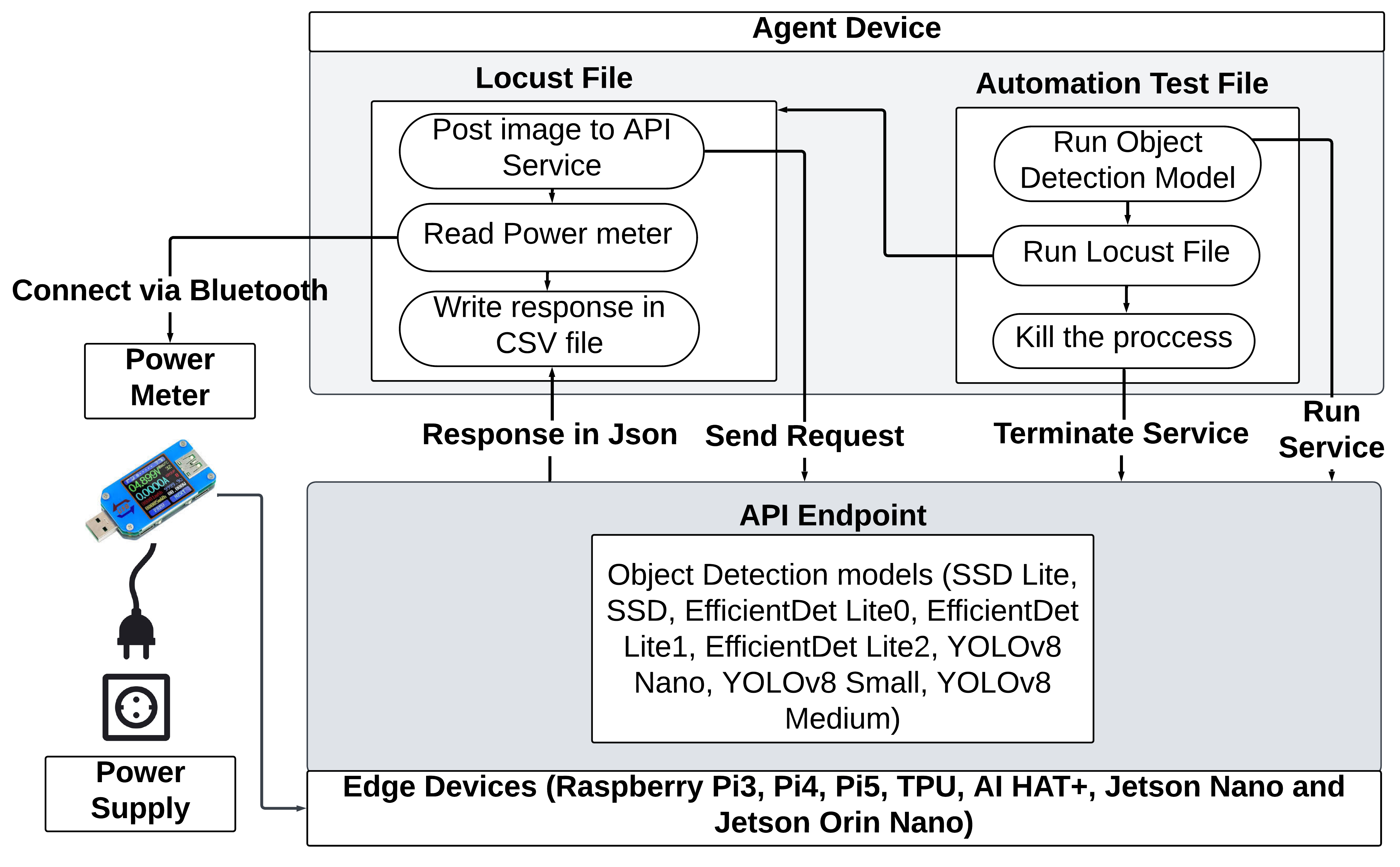}
    \caption{Experimental software and hardware setup.}
    \label{fig:expermintal proc}
\end{figure}

The evaluation procedure consists of several stages, as illustrated in Fig.~\ref{fig:expermintal proc}. First, the base energy consumption of each device is measured over a five-minute period while the device remains idle and no inference workload is executed. Energy readings are collected using a Python-based energy reader running on a separate device connected to the UM25C power meter.

Next, the object detection service is executed on the target edge device, and a Locust~\footnote{https://locust.io/} client is used to generate requests for five minutes. In particular, the workload is generated using a single Locust user, meaning that only one request is active at a time and each new request is issued after the previous response is received. During this period, the total energy consumption, number of processed requests, and inference time values returned by the service are recorded automatically and stored in CSV files.

To automate the experiments, a bash script running on an agent device launches the object detection service on the target edge platform, executes the Locust workload, terminates the service after completion, and then proceeds to the next model. This process ensures consistent execution across all evaluated model-device combinations. Each experiment is repeated three times, and the average values are used in the final analysis.

Finally, accuracy is evaluated using FiftyOne and the COCO validation dataset by comparing model predictions with ground-truth annotations to compute \textit{mAP}. In addition to the full validation dataset evaluation, we also perform an object-content-based accuracy analysis to examine how detection accuracy varies with image content.

\section{Performance Evaluation Results}
\label{sec:evaluation results}
\subsection{Energy Consumption}
This section presents the results for base energy consumption, total energy consumption per request, and energy consumption per request excluding base energy across the evaluated edge devices.

\begin{figure}[t]
    \centering
    \subfigure[Base energy]{
        \includegraphics[width=0.22\textwidth]{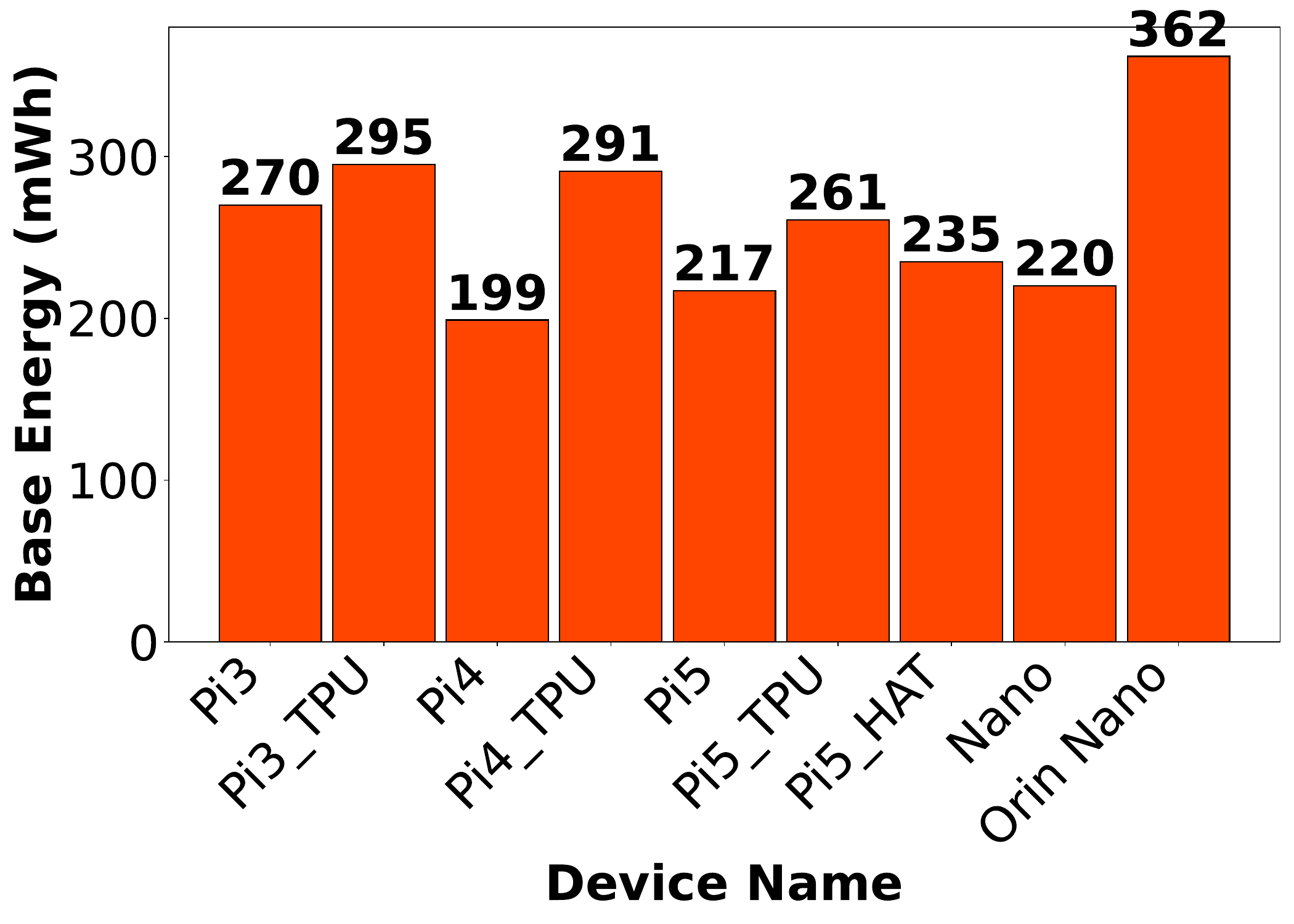}
    }
    \subfigure[Raspberry Pi3]{
        \includegraphics[width=0.22\textwidth]{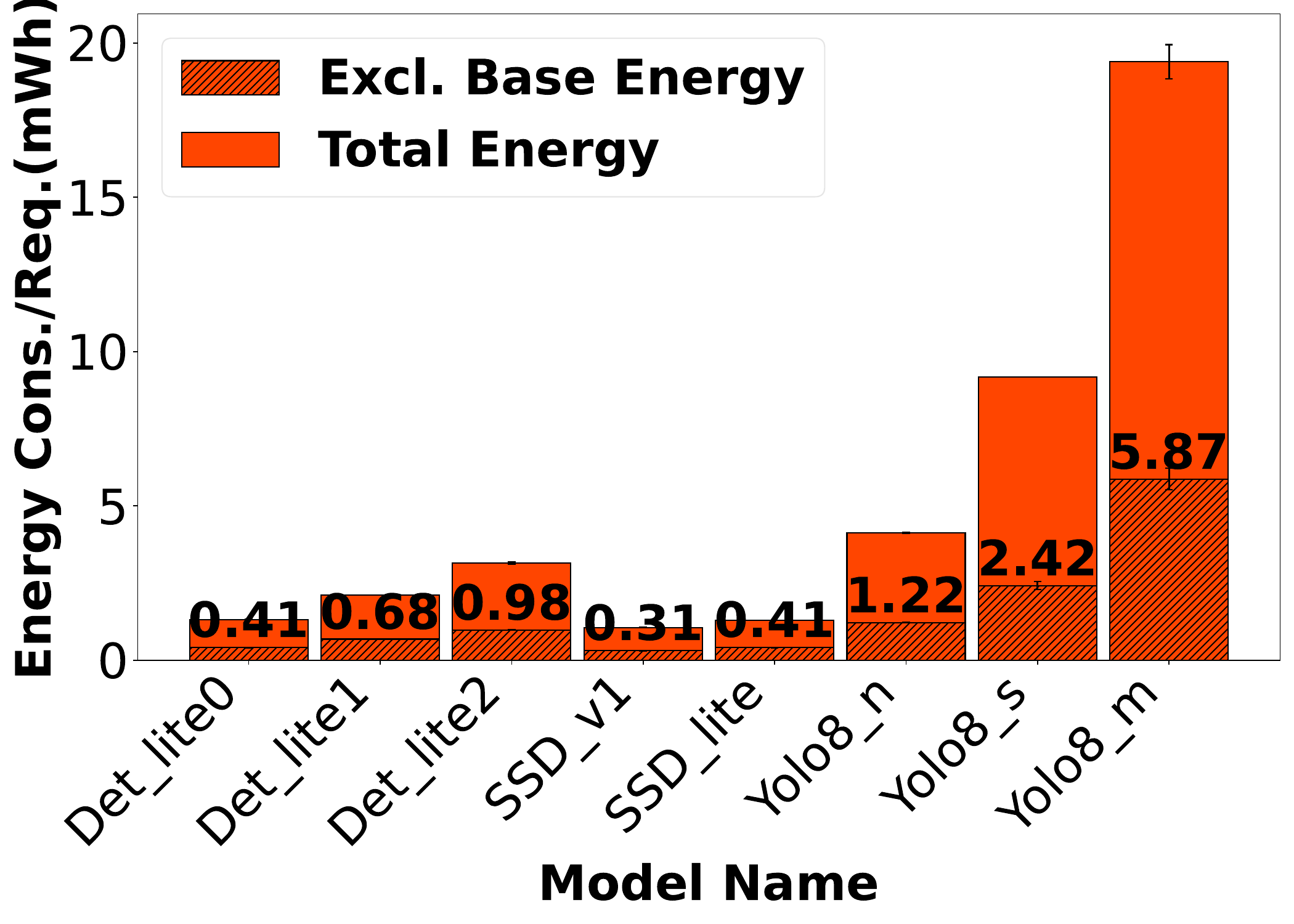}
    }
    \subfigure[Pi3 + TPU]{
        \includegraphics[width=0.22\textwidth]{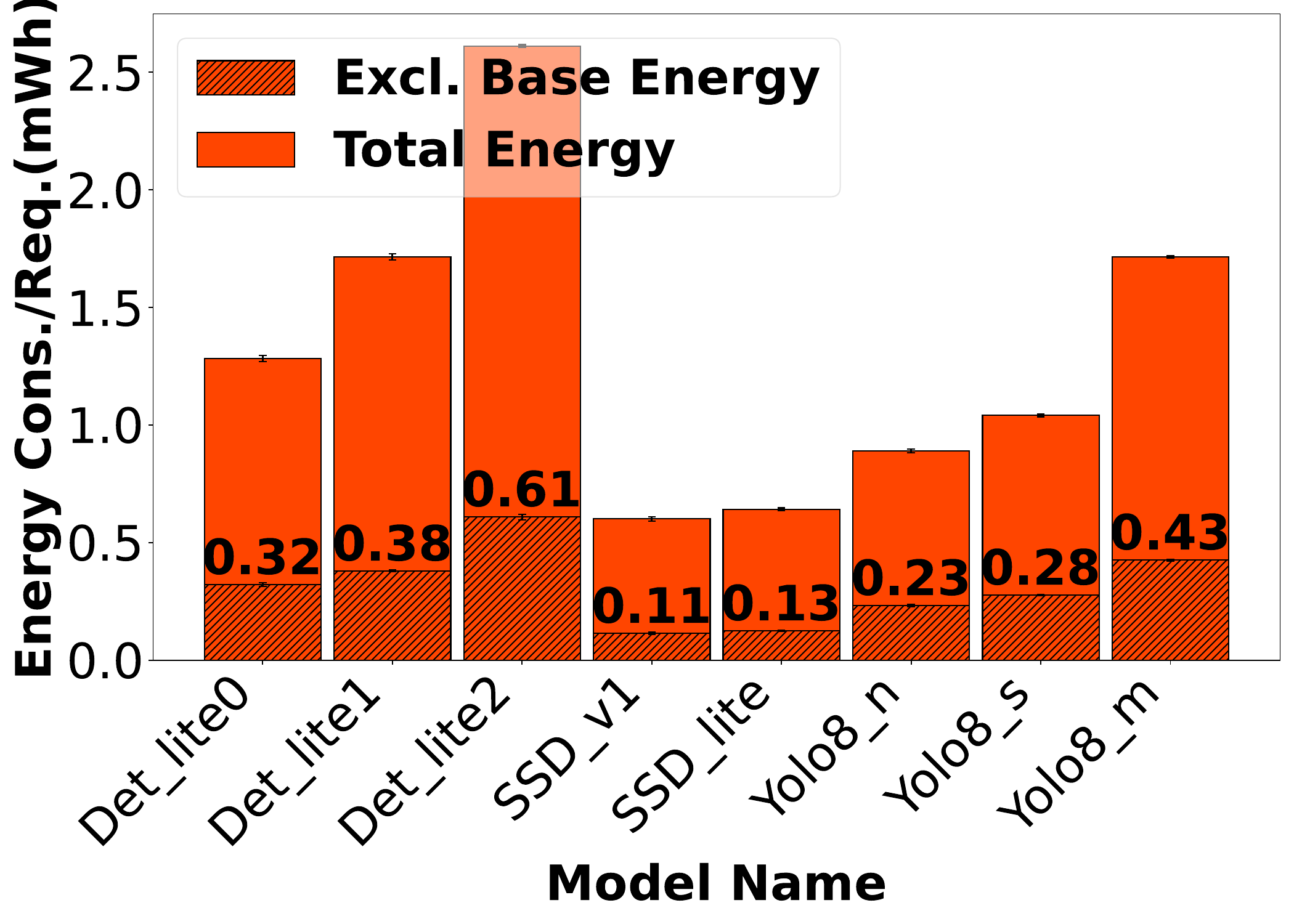}
    }
    \subfigure[Raspberry Pi4]{
        \includegraphics[width=0.22\textwidth]{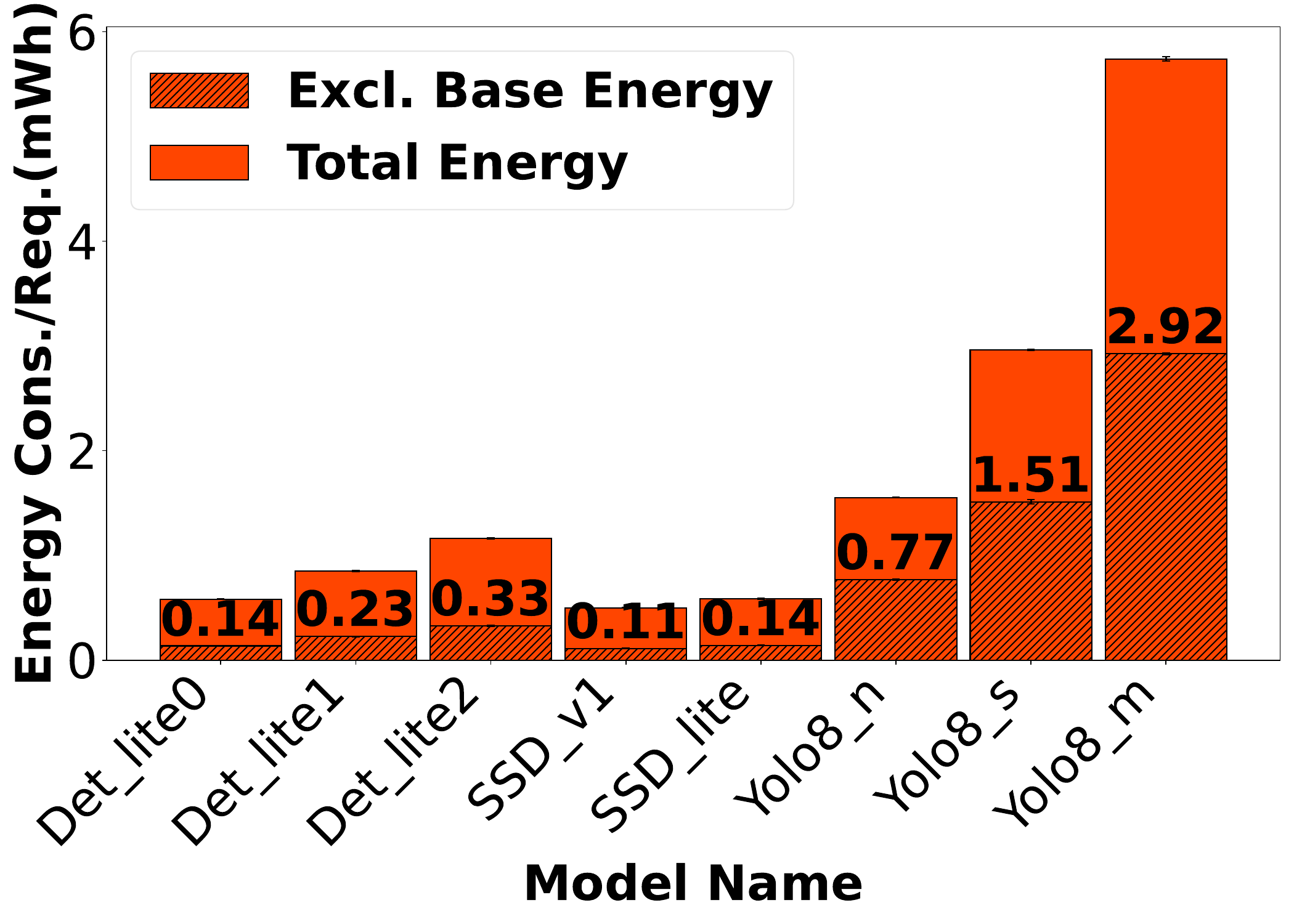}
    }
    \subfigure[Pi4 + TPU]{
        \includegraphics[width=0.22\textwidth]{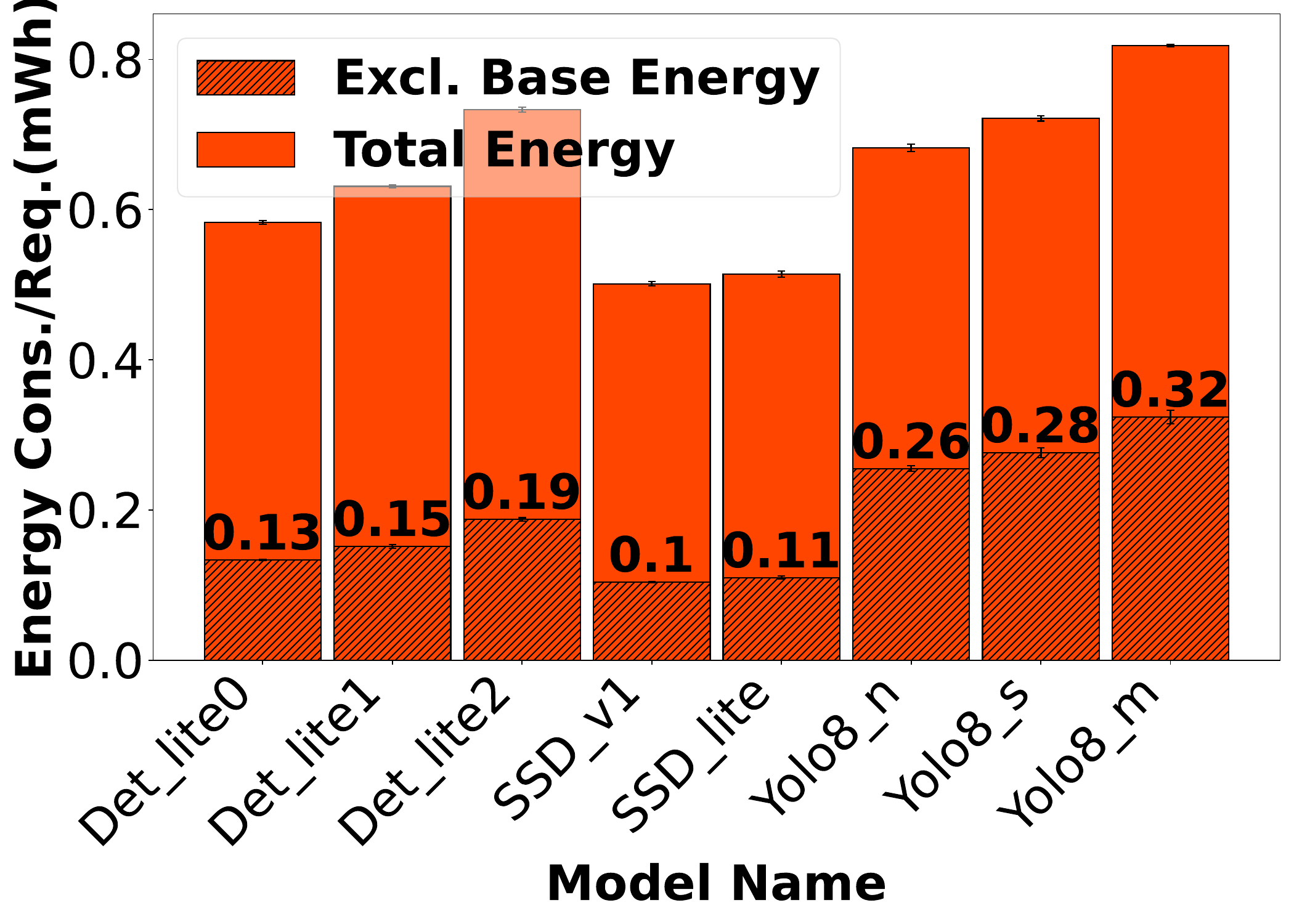}
    }
    \subfigure[Raspberry Pi5]{
        \includegraphics[width=0.22\textwidth]{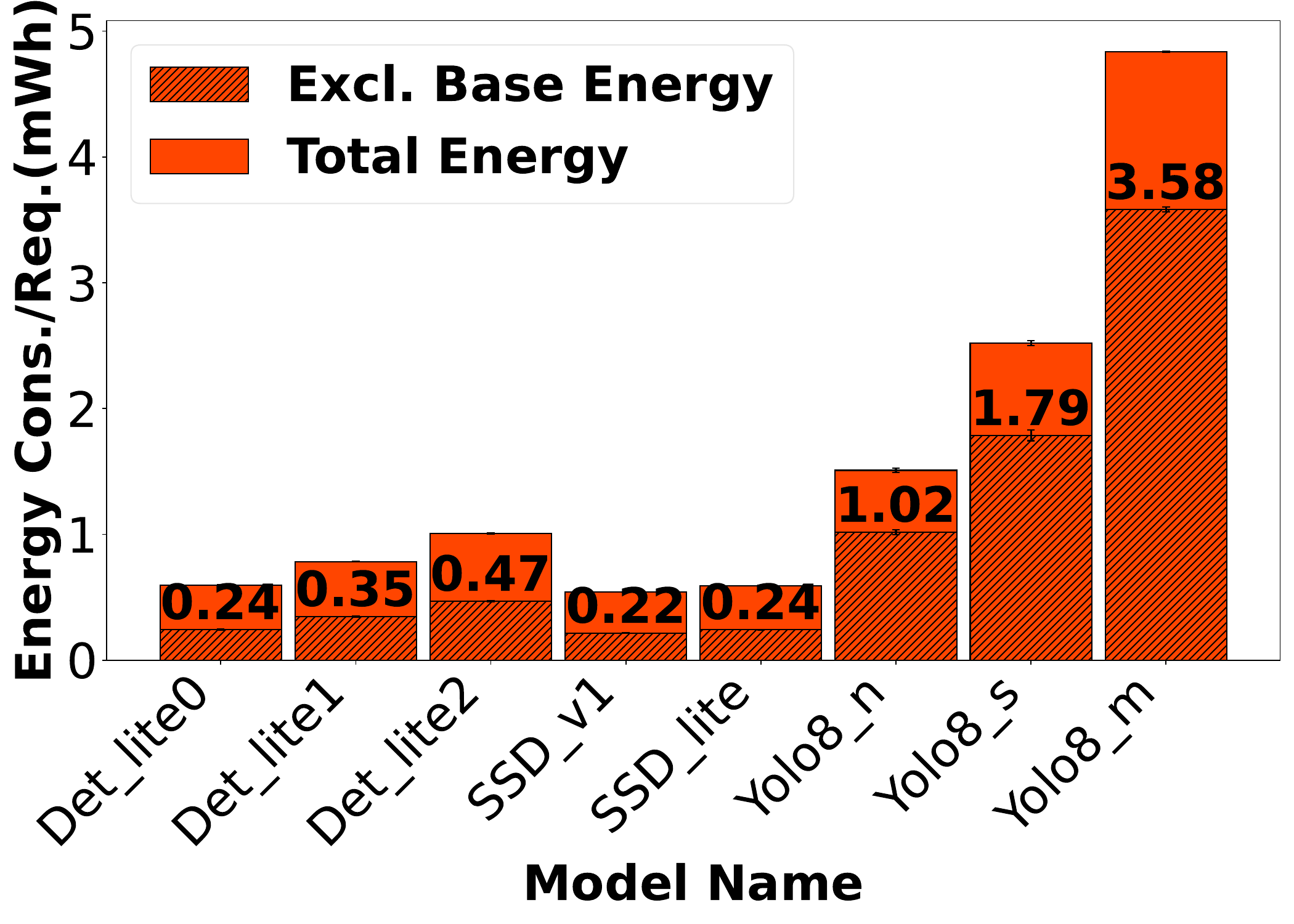}
    }
    \subfigure[Pi5 + TPU]{
        \includegraphics[width=0.22\textwidth]{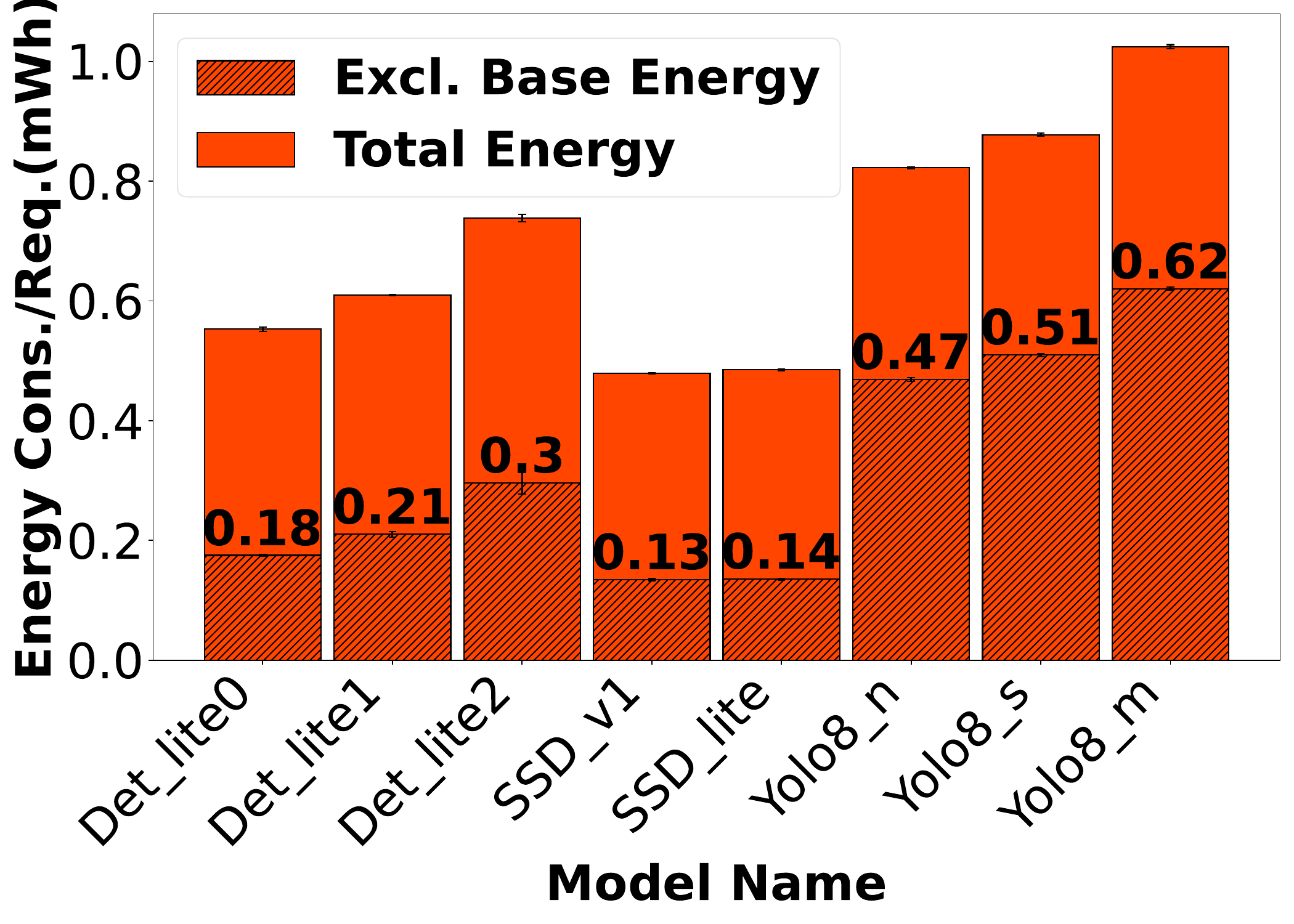}
    }
    \subfigure[Pi5 + AI HAT]{
        \includegraphics[width=0.22\textwidth]{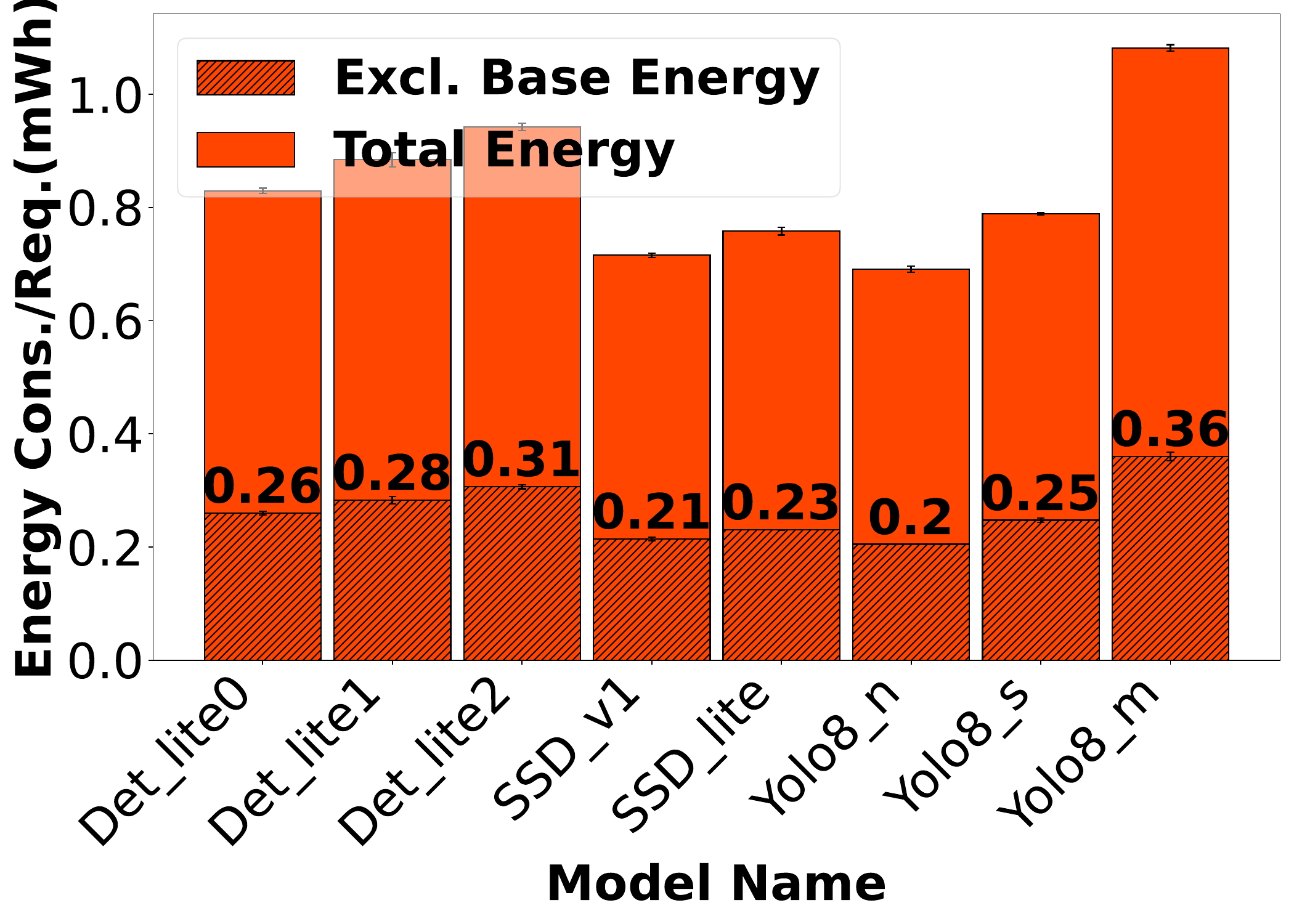}
    }
     \subfigure[Jetson Nano]{
        \includegraphics[width=0.22\textwidth]{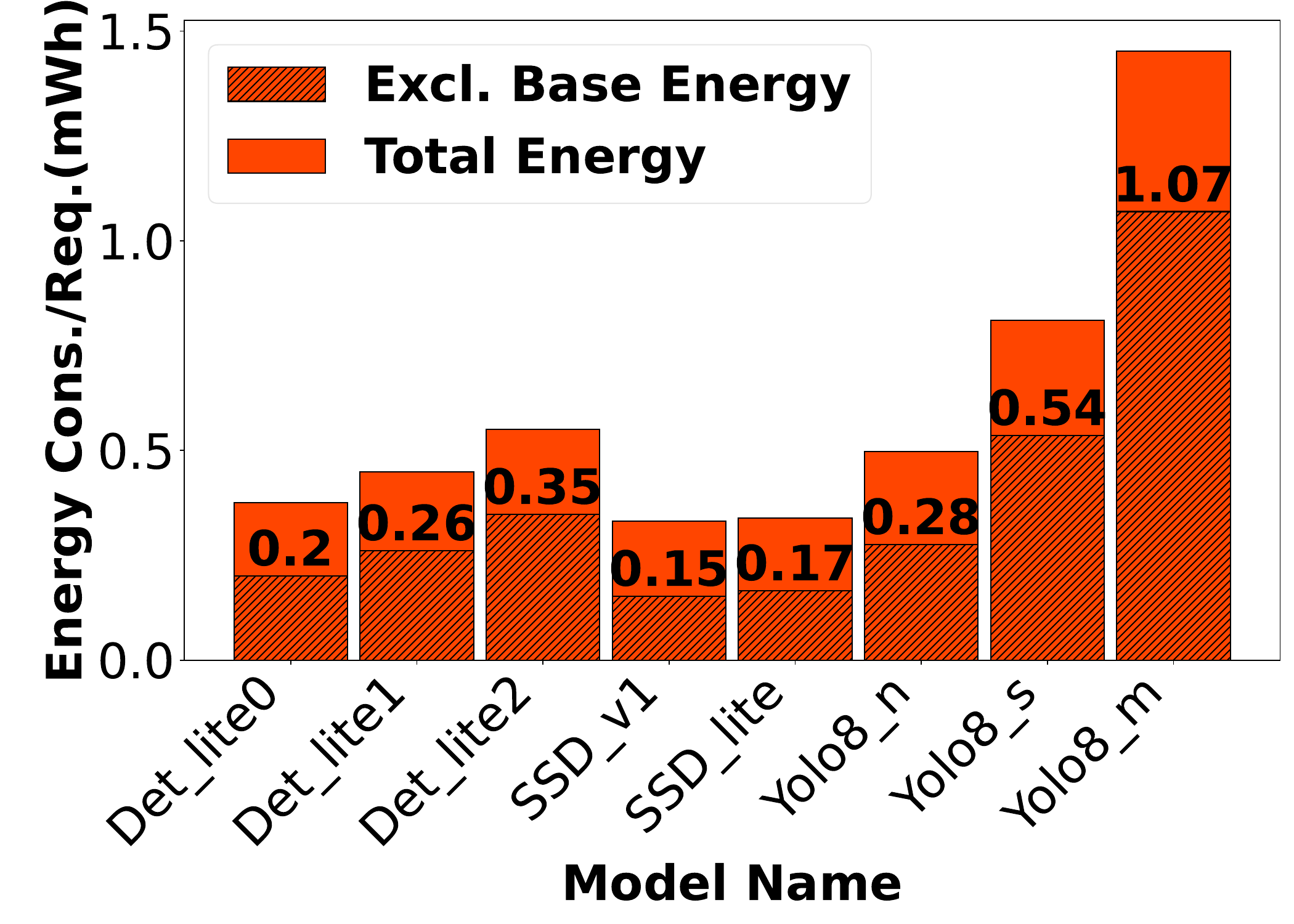}
    }
    \subfigure[Jetson Orin Nano]{
        \includegraphics[width=0.22\textwidth]{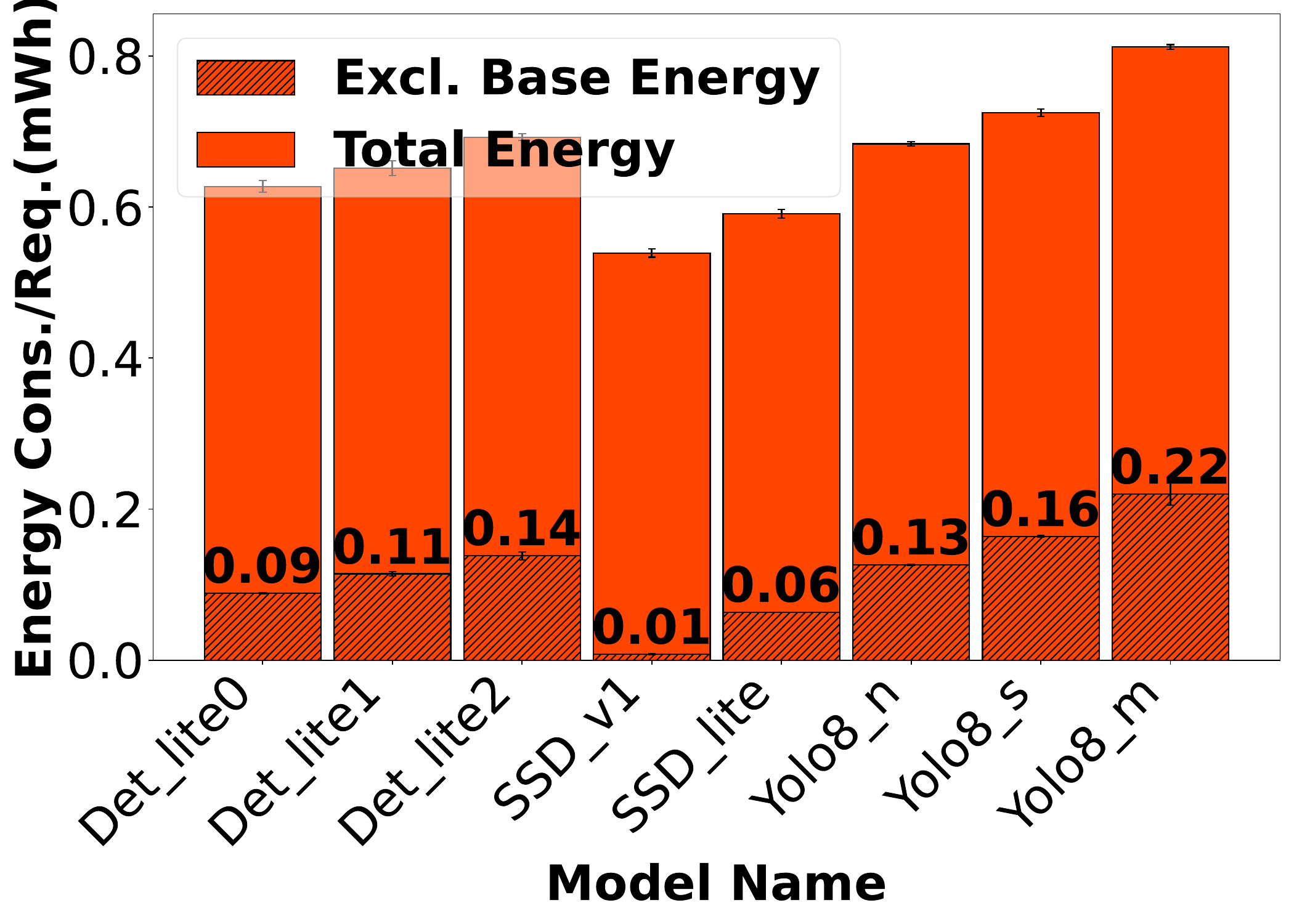}
    }
    \caption{Base energy for different edge devices, energy consumption per request excluding base energy and total energy consumption per request across different edge devices.}
    \label{fig:energy consumption per request for devices}
\end{figure}

We begin with the base energy consumption results, which provide insight into the idle energy characteristics of the different hardware platforms before analyzing request-level energy behavior. Fig.~\ref{fig:energy consumption per request for devices}(a) first reports the base energy consumption of the devices in mWh. Among the CPU-only Raspberry Pi platforms, Raspberry Pi 3 exhibits the highest base energy consumption at 270 mWh, followed by Raspberry Pi 5 at 217 mWh and Raspberry Pi 4 at 199 mWh. This suggests that the newer Raspberry Pi generations are generally more energy efficient in the idle state. When TPU-based configurations are considered, Raspberry Pi 3 with TPU and Raspberry Pi 4 with TPU show similar base energy consumption values of 295 mWh and 291 mWh, respectively, while Raspberry Pi 5 with TPU records a lower value of 261 mWh. Raspberry Pi 5 with AI HAT+ has a base energy consumption of 235 mWh, which is lower than Raspberry Pi 5 with TPU but slightly higher than Raspberry Pi 5 without acceleration. Among the NVIDIA platforms, Jetson Nano records a base energy consumption of 220 mWh, whereas Jetson Orin Nano exhibits the highest base energy consumption overall at 362 mWh.

In addition to the base energy consumption, the energy consumption per request excluding base energy reveals clear differences across the evaluated model-device combinations. On Raspberry Pi 3, as shown in Fig.~\ref{fig:energy consumption per request for devices}(b), the EfficientDet Lite models consume between 0.41 and 0.98 mWh per request, while SSD\_v1 and SSD\_lite consume 0.31 mWh and 0.41 mWh, respectively. The YOLOv8 models exhibit substantially higher energy consumption, ranging from 1.22 to 5.87 mWh per request. When TPU acceleration is added to Raspberry Pi 3, as shown in Fig.~\ref{fig:energy consumption per request for devices}(c), the energy consumption decreases across all model families. In this setting, EfficientDet Lite models consume between 0.32 and 0.61 mWh per request, SSD\_v1 and SSD\_lite consume 0.11 mWh and 0.13 mWh, respectively, and YOLOv8 models range from 0.23 to 0.43 mWh per request. 

A similar trend is observed for Raspberry Pi 4 and its TPU-based configuration. On Raspberry Pi 4, as shown in Fig.~\ref{fig:energy consumption per request for devices}(d), EfficientDet Lite models consume between 0.14 and 0.33 mWh per request, while SSD\_v1 and SSD\_lite consume 0.11 mWh and 0.14 mWh, respectively. The YOLOv8 models show noticeably higher energy consumption, ranging from 0.77 to 2.92 mWh per request. When TPU acceleration is used, as shown in Fig.~\ref{fig:energy consumption per request for devices}(e), the energy consumption decreases across all model families. In this setting, EfficientDet Lite models consume between 0.13 and 0.19 mWh per request, SSD\_v1 and SSD\_lite consume 0.10 mWh and 0.11 mWh, respectively, and YOLOv8 models range from 0.26 to 0.32 mWh per request.

For Raspberry Pi 5 and its accelerator-based configurations, the same overall pattern is maintained. On Raspberry Pi 5, shown in Fig.~\ref{fig:energy consumption per request for devices}(f), EfficientDet Lite models consume between 0.24 and 0.47 mWh per request, while SSD\_v1 and SSD\_lite consume 0.22 mWh and 0.24 mWh, respectively. The YOLOv8 models again record higher values, ranging from 1.02 to 3.58 mWh per request. With TPU acceleration on Raspberry Pi 5, as shown in Fig.~\ref{fig:energy consumption per request for devices}(g), EfficientDet Lite models consume between 0.18 and 0.30 mWh per request, SSD\_v1 and SSD\_lite consume 0.13 mWh and 0.14 mWh, respectively, and YOLOv8 models range from 0.47 to 0.62 mWh per request. For Raspberry Pi 5 with AI HAT+, shown in Fig.~\ref{fig:energy consumption per request for devices}(h), EfficientDet Lite models consume between 0.26 and 0.31 mWh per request, SSD\_v1 and SSD\_lite consume 0.21 mWh and 0.23 mWh, respectively, and YOLOv8 models range from 0.20 to 0.36 mWh per request.

For the NVIDIA platforms, as shown in Fig.~\ref{fig:energy consumption per request for devices}(i) and Fig.~\ref{fig:energy consumption per request for devices}(j), Jetson Nano and Jetson Orin Nano both show strong energy efficiency. On Jetson Nano, EfficientDet Lite models consume between 0.20 and 0.35 mWh per request, while SSD\_v1 and SSD\_lite consume 0.15 mWh and 0.17 mWh, respectively. The YOLOv8 models range from 0.28 to 1.07 mWh per request. On Jetson Orin Nano, the corresponding values decrease further, with EfficientDet Lite models consuming between 0.09 and 0.14 mWh per request, SSD\_v1 and SSD\_lite consuming 0.01 mWh and 0.06 mWh, respectively, and YOLOv8 models ranging from 0.13 to 0.22 mWh per request.

\begin{keyinsight}
Per-request energy decreases significantly when the selected model can exploit the target accelerator efficiently. SSD and EfficientDet Lite benefit from TPU acceleration because their TFLite deployments are well matched to Edge TPU execution, while YOLOv8 consumes substantially more energy on CPU-only Raspberry Pi devices because larger YOLO variants require heavier computation and use larger input resolution. However, idle power also matters: devices such as Jetson Orin Nano have high base energy but low per-request energy because they complete inference quickly. Therefore, for battery-powered IoT deployments, both idle energy and request-level energy should be considered.  
\end{keyinsight}

\subsection{Inference Time}

\begin{figure}[t]
    \centering
    \subfigure[Raspberry Pi3]{
        \includegraphics[width=0.22\textwidth]{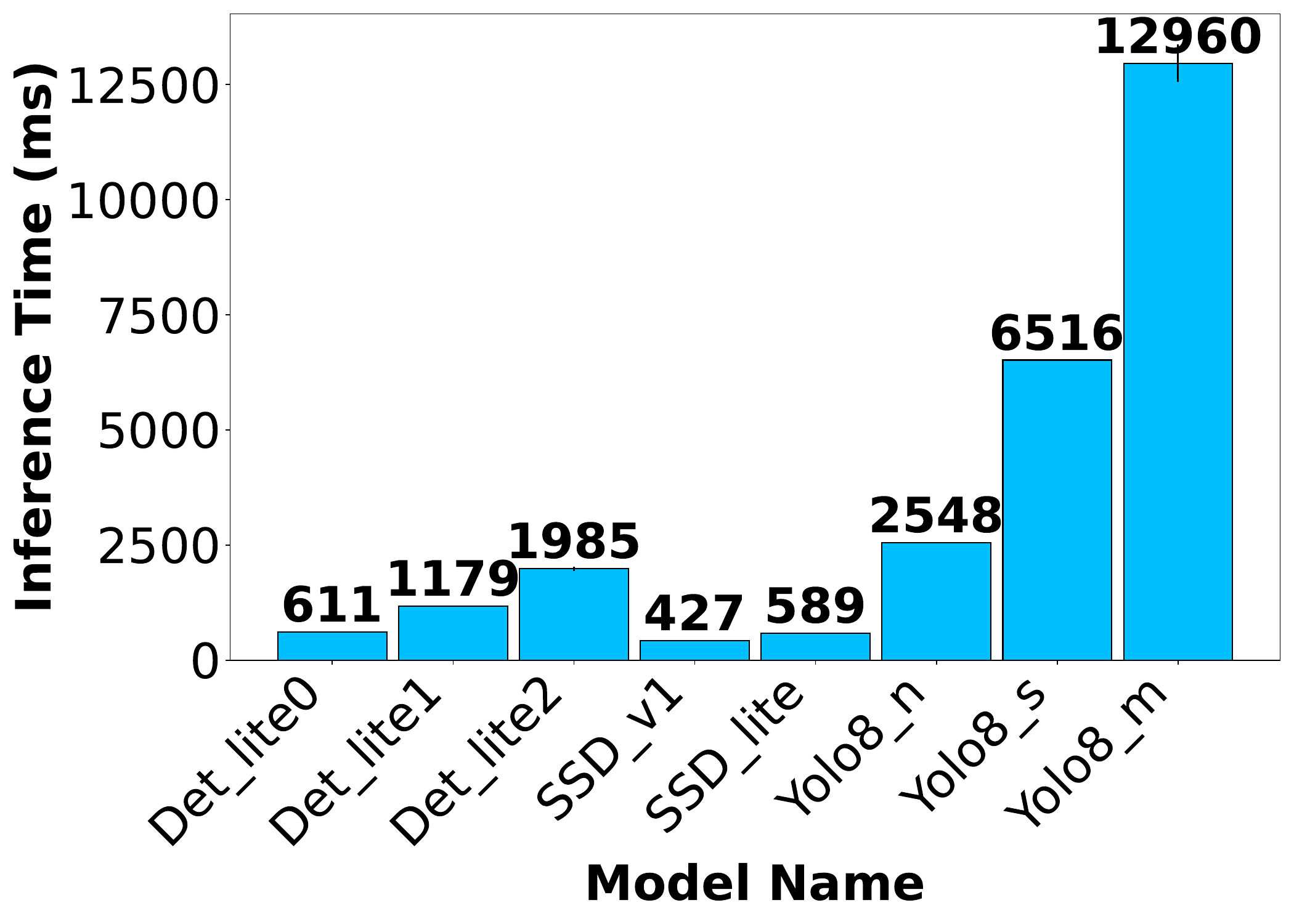}
    }
    \subfigure[Pi3 + TPU]{
        \includegraphics[width=0.22\textwidth]{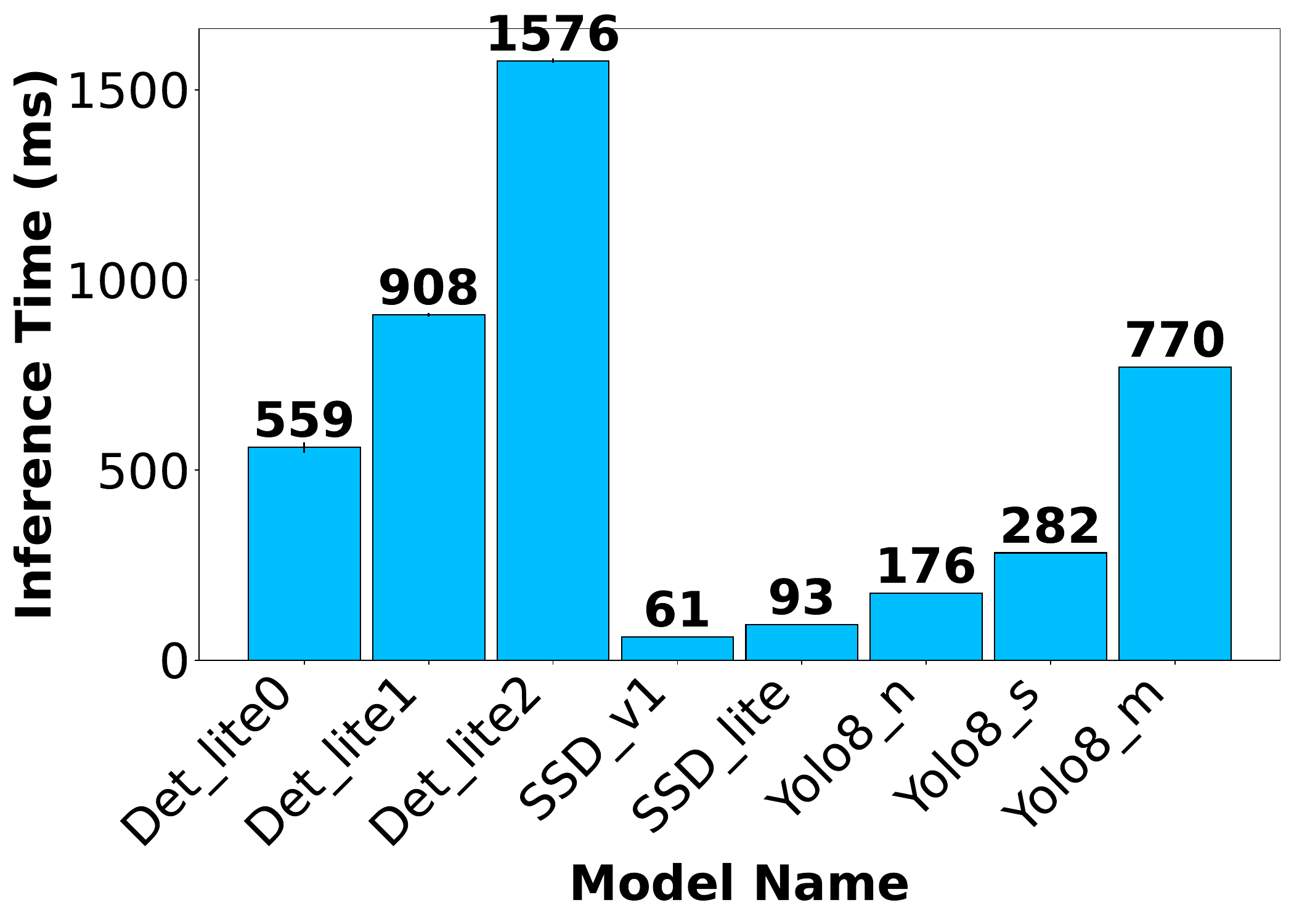}
    }
    \subfigure[Raspberry Pi4]{
        \includegraphics[width=0.22\textwidth]{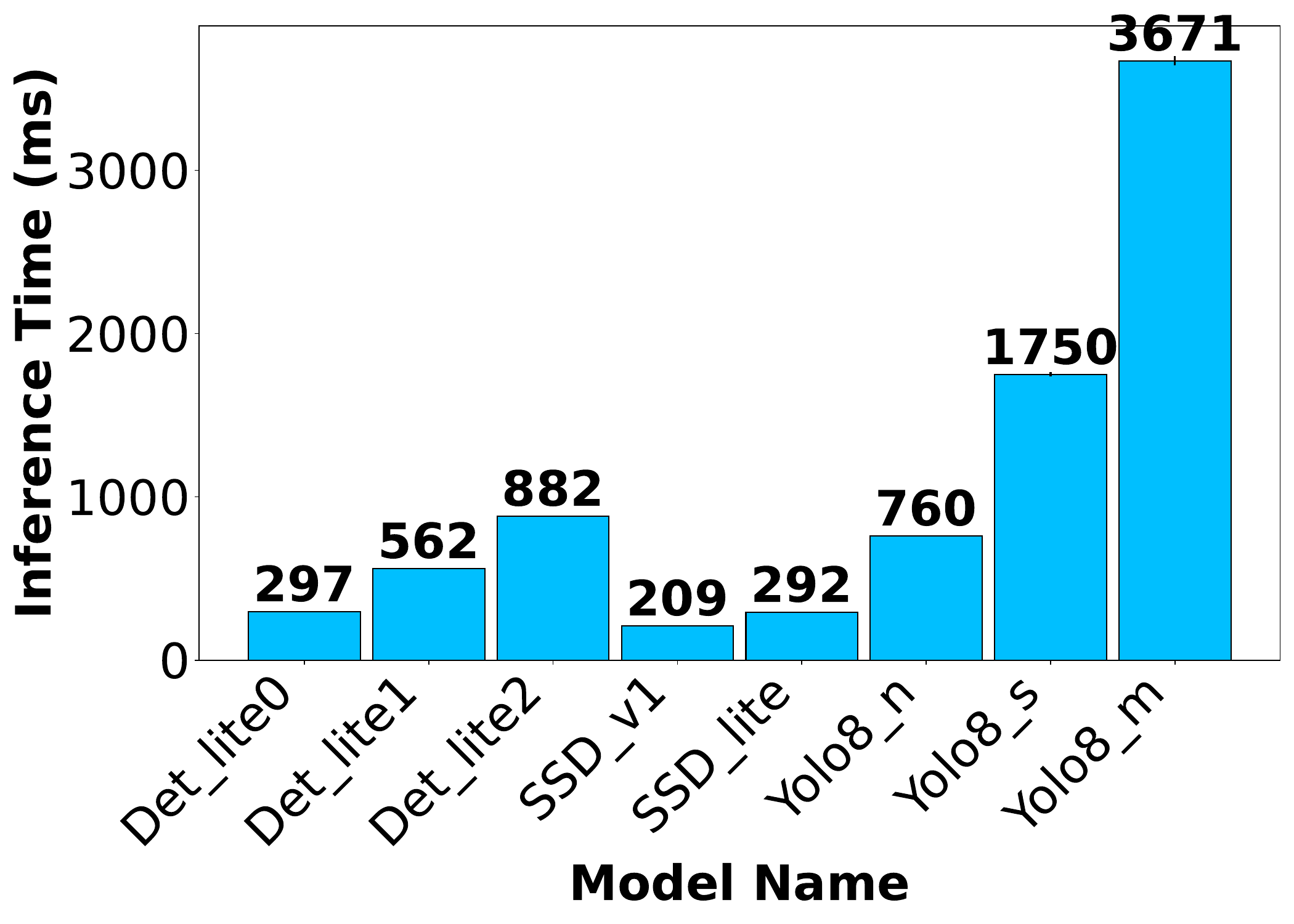}
    }
    \subfigure[Pi4 + TPU]{
        \includegraphics[width=0.22\textwidth]{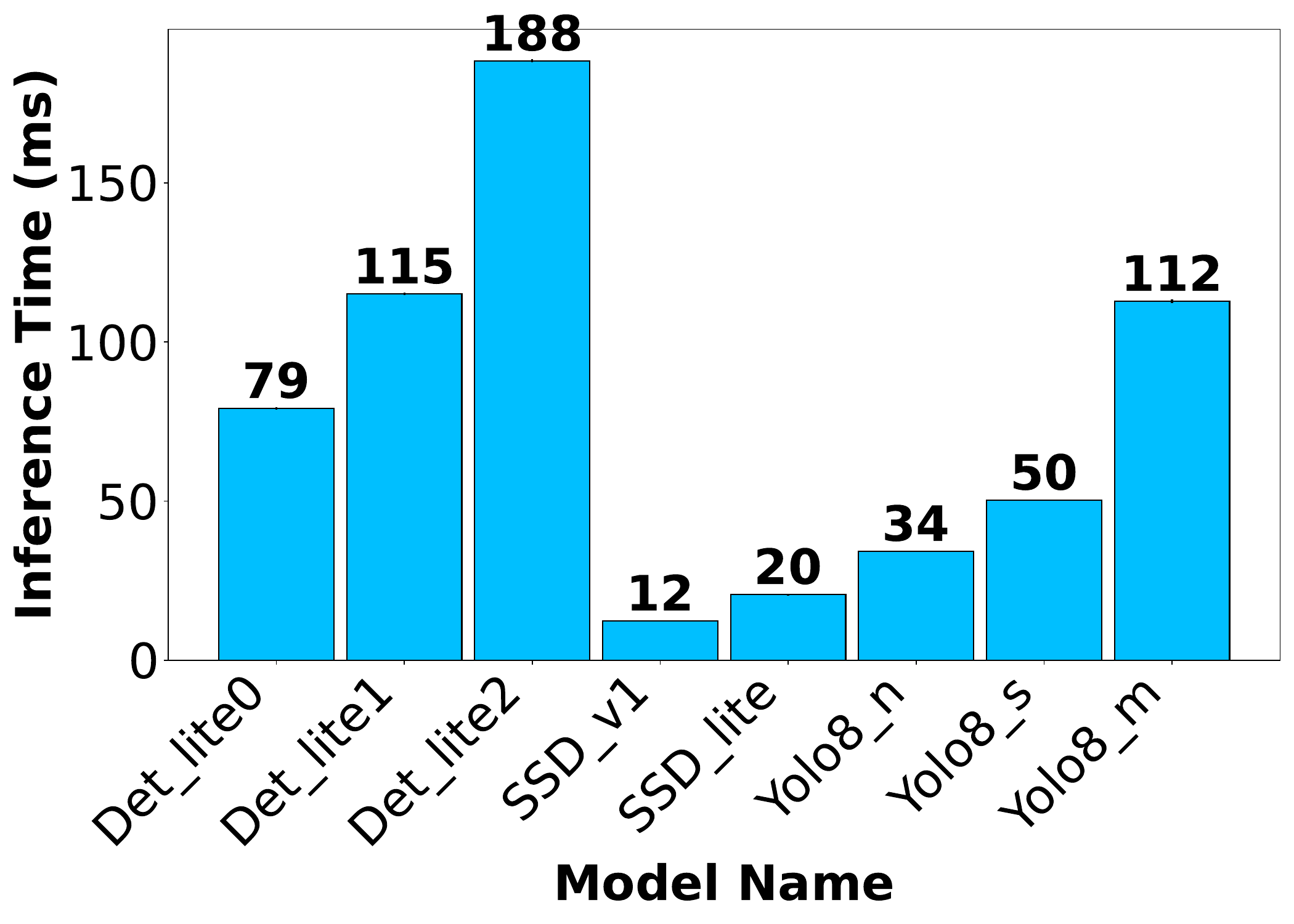}
    }
    \subfigure[Raspberry Pi5]{
        \includegraphics[width=0.22\textwidth]{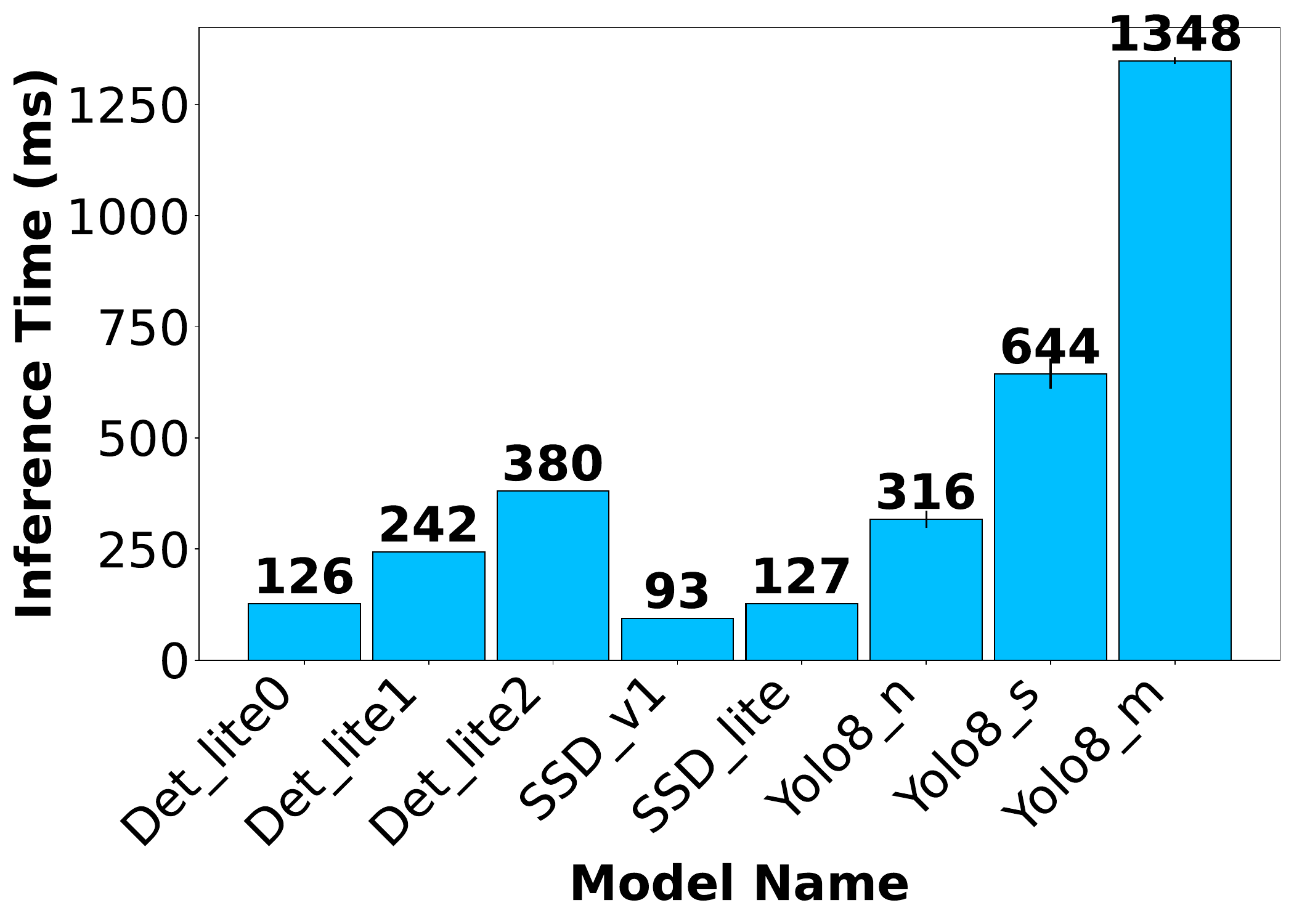}
    }
    \subfigure[Pi5 + TPU]{
        \includegraphics[width=0.22\textwidth]{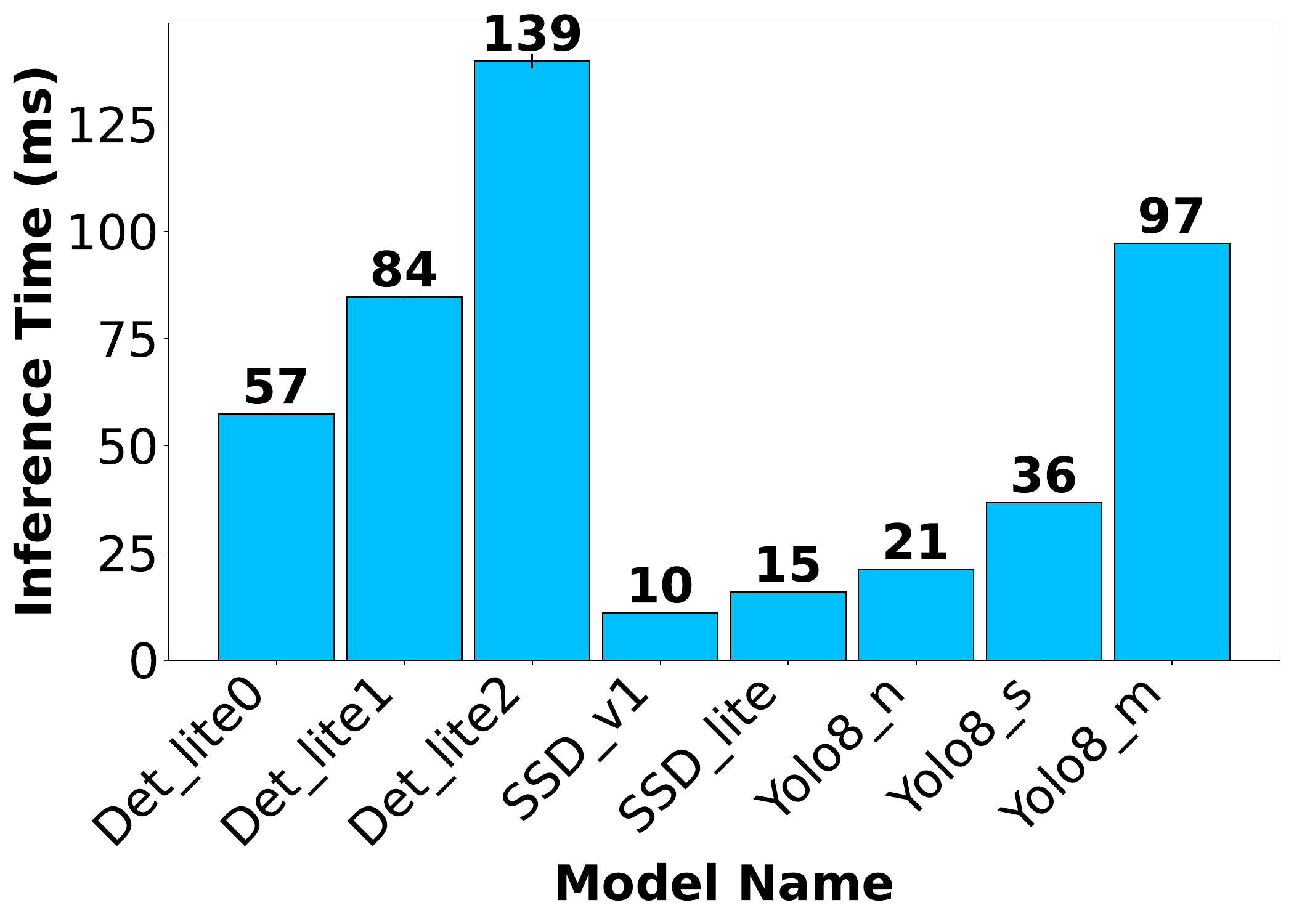}
    }
    \subfigure[Pi5 + AI HAT]{
        \includegraphics[width=0.22\textwidth]{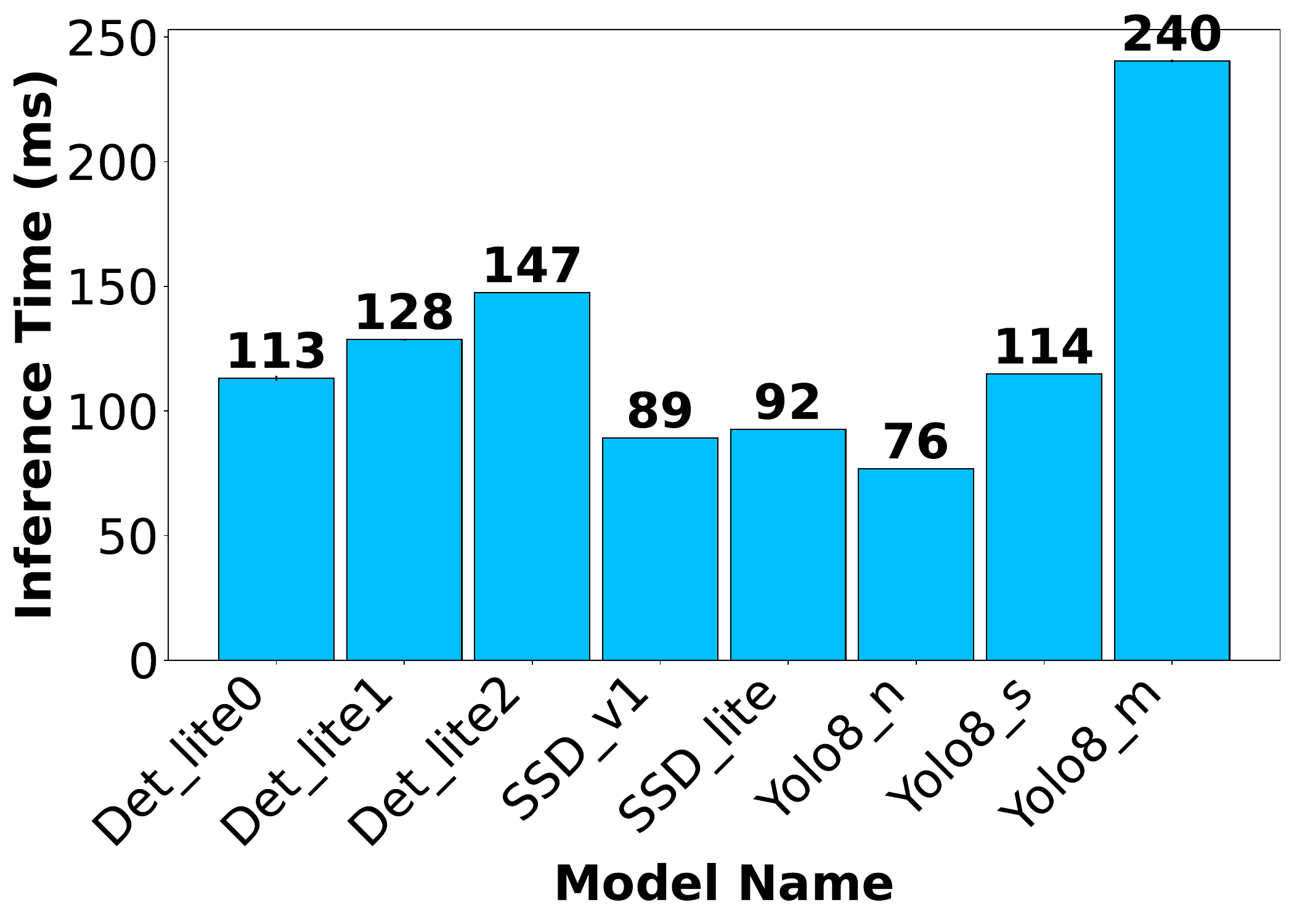}
    }
    \subfigure[Jetson Nano]{
        \includegraphics[width=0.22\textwidth]{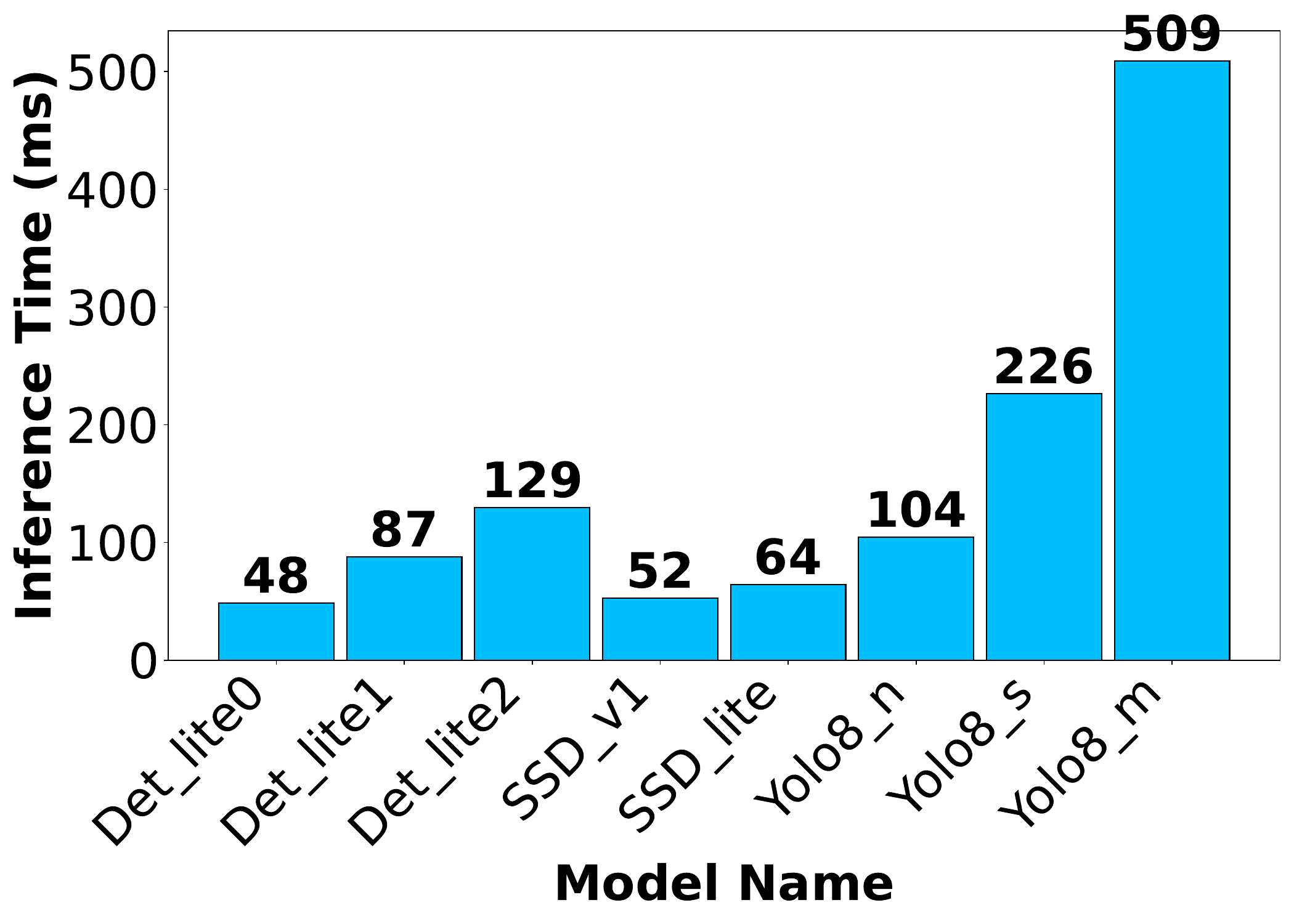}
    }
     \subfigure[Jetson Orin Nano]{
        \includegraphics[width=0.22\textwidth]{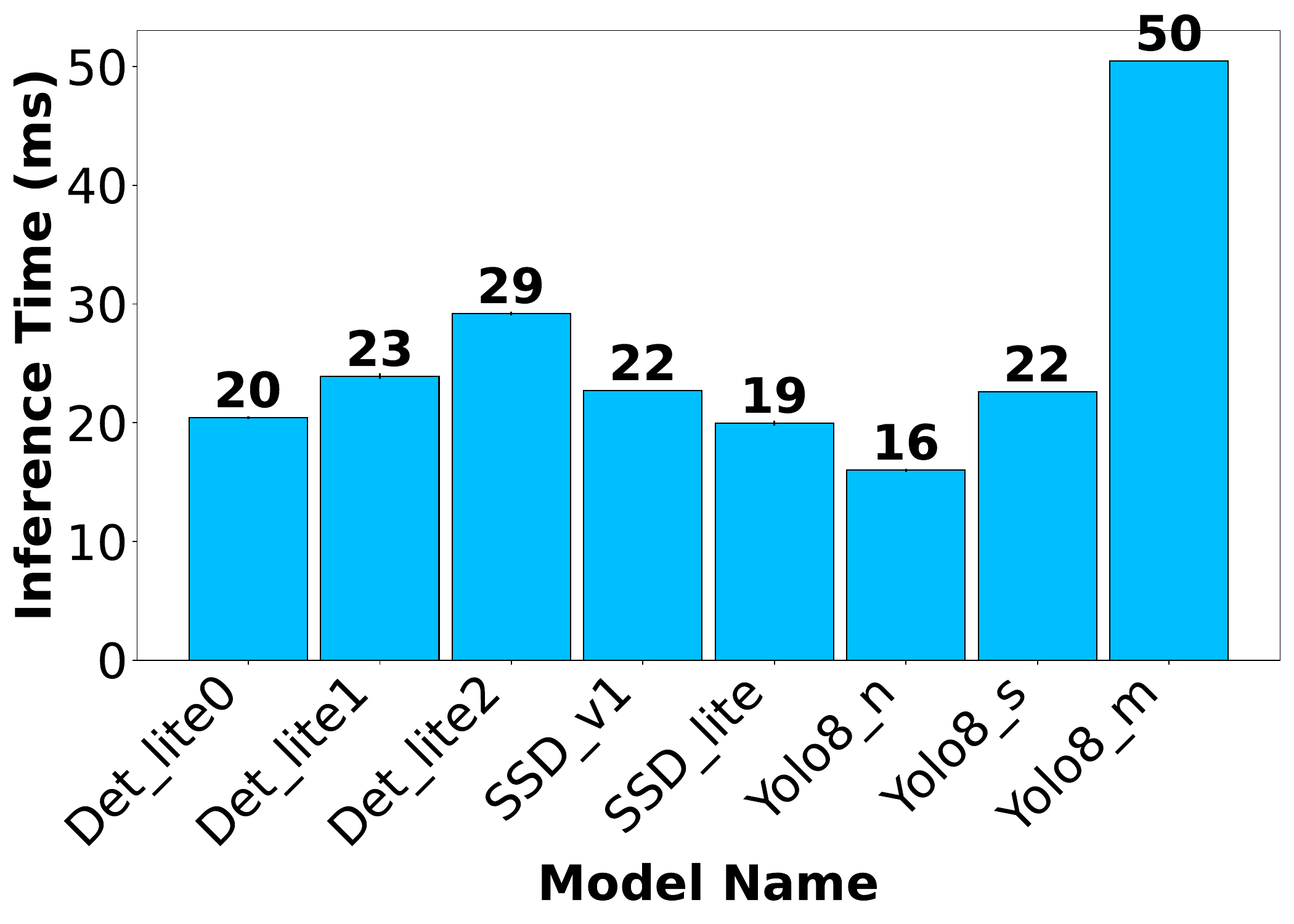}
    }
    \caption{Inference Time per request for different edge devices.}
    \label{fig:inference time per request for devices}
\end{figure}

This section analyzes the inference time of the evaluated object detection models across the investigated edge devices. The results, reported in milliseconds, reveal clear differences in latency across model-device combinations.

On Raspberry Pi 3, as shown in Fig.~\ref{fig:inference time per request for devices}(a), SSD\_v1 achieves the lowest inference time at 427 ms, followed by SSD\_lite at 589 ms. The EfficientDet Lite models require between 611 and 1,985 ms, while the YOLOv8 models exhibit substantially higher latency, ranging from 2,548 to 12960 ms. When TPU acceleration is added to Raspberry Pi 3, as shown in Fig.~\ref{fig:inference time per request for devices}(b), the inference times of SSD and YOLOv8 models decrease markedly. In this configuration, SSD\_v1 remains the fastest at 61 ms, whereas EfficientDet Lite2 records the highest inference time at 1,576 ms.

A similar trend is observed for Raspberry Pi 4 and its TPU-based configuration. On Raspberry Pi 4, as shown in Fig.~\ref{fig:inference time per request for devices}(c), SSD\_v1 and SSD\_lite achieve the lowest inference times at 209 ms and 292 ms, respectively. EfficientDet Lite models range from 297 to 882 ms, while YOLOv8 models range from 760 to 3,671 ms, with YOLOv8\_m exhibiting the highest latency. When TPU acceleration is used, as shown in Fig.~\ref{fig:inference time per request for devices}(d), latency decreases substantially across all model families. In this setting, SSD\_v1 records 12 ms, SSD\_lite records 20 ms, EfficientDet Lite models range from 79 to 188 ms, and YOLOv8 models range from 34 to 112 ms.

For Raspberry Pi 5 and its accelerator-based configurations, the same overall pattern is maintained. On Raspberry Pi 5, shown in Fig.~\ref{fig:inference time per request for devices}(e), SSD\_v1 and SSD\_lite achieve the lowest inference times at 93 ms and 127 ms, respectively, while EfficientDet Lite models range from 126 to 380 ms and YOLOv8 models range from 316 to 1,348 ms. For Raspberry Pi 5 with TPU, shown in Fig.~\ref{fig:inference time per request for devices}(f), SSD\_v1 records the lowest inference time at 10 ms, SSD\_lite records 15 ms, EfficientDet Lite models range from 57 to 139 ms, and YOLOv8 models range from 21 to 97 ms. For Raspberry Pi 5 with AI HAT+, shown in Fig.~\ref{fig:inference time per request for devices}(g), SSD\_v1 and SSD\_lite record 89 ms and 92 ms, respectively, EfficientDet Lite models range from 113 to 147 ms, and YOLOv8 models range from 76 to 240 ms. Notably, YOLOv8\_n achieves the lowest inference time on this platform at 76 ms.

For the NVIDIA platforms, as shown in Fig.~\ref{fig:inference time per request for devices}(h) and Fig.~\ref{fig:inference time per request for devices}(i), Jetson Nano and Jetson Orin Nano both exhibit strong latency performance. On Jetson Nano, EfficientDet Lite models range from 48 to 129 ms, while SSD\_v1 and SSD\_lite record 52 ms and 64 ms, respectively. The YOLOv8 models range from 104 to 509 ms. On Jetson Orin Nano, the corresponding values decrease further, with EfficientDet Lite models ranging from 20 to 29 ms, SSD\_v1 and SSD\_lite recording 22 ms and 19 ms, respectively, and YOLOv8 models ranging from 16 to 50 ms. On this platform, YOLOv8\_n achieves the minimum inference time at 16 ms, while YOLOv8\_m remains the slowest at 50 ms.

\begin{keyinsight}
The large inference-time gap between CPU-only Raspberry Pi devices and accelerator-enabled platforms is mainly caused by the limited compute capability of CPU-only devices and the ability of TPUs, NPUs, and GPUs to execute tensor operations more efficiently. This improvement is especially visible for computationally intensive models such as YOLOv8. However, the TPU results should be interpreted carefully because YOLOv8 is deployed with a reduced 320 × 320 input size on TPU, while CPU, Hailo, and TensorRT deployments use 640 × 640. Therefore, the lower inference time on TPU is influenced by both hardware acceleration and reduced input resolution.
\end{keyinsight}

\subsection{Accuracy}

\begin{figure}
    \centering
    \subfigure[Raspberry Pi]{
        \includegraphics[width=0.22\textwidth]{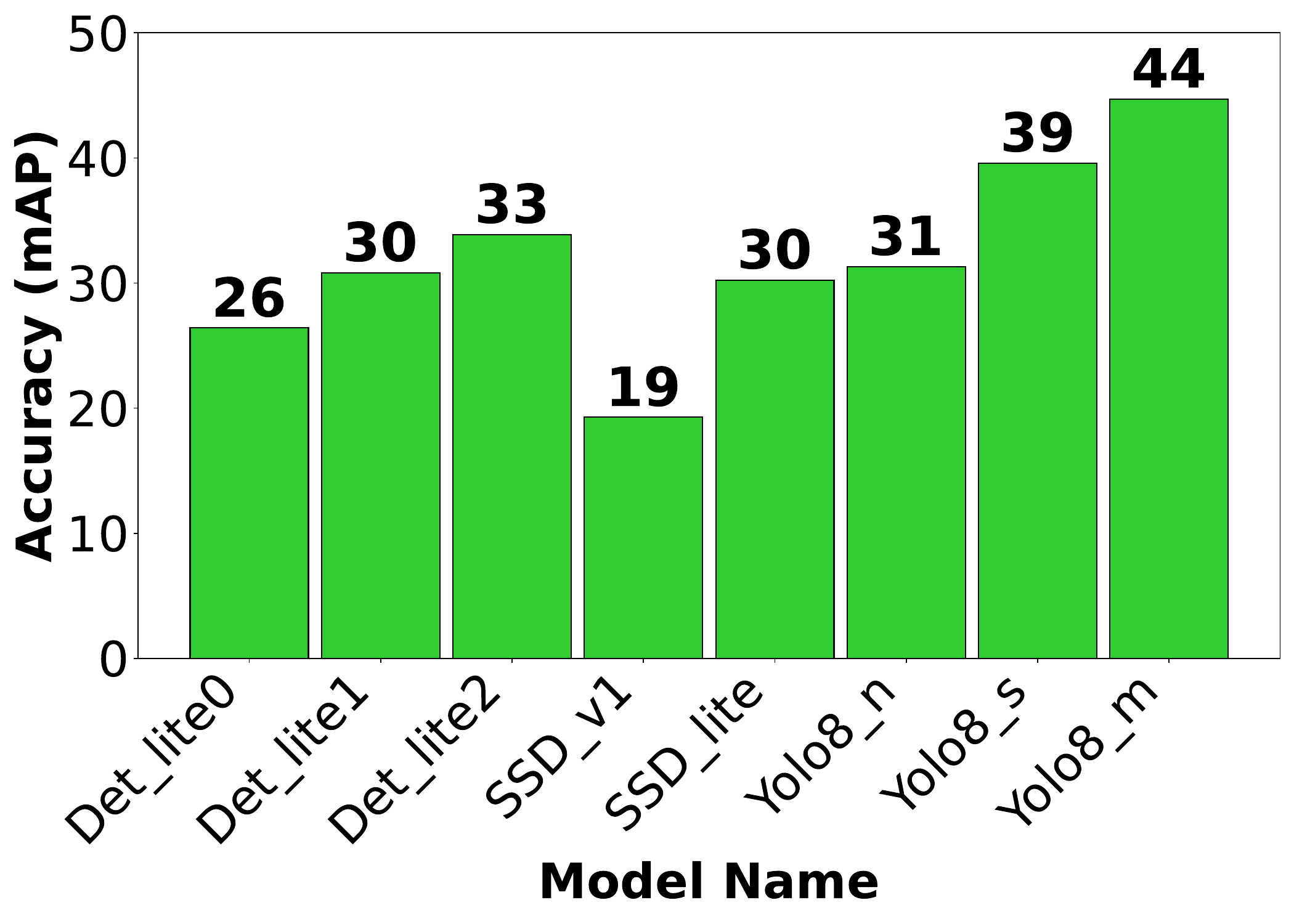}
    }
    \subfigure[Pi + TPU]{
        \includegraphics[width=0.22\textwidth]{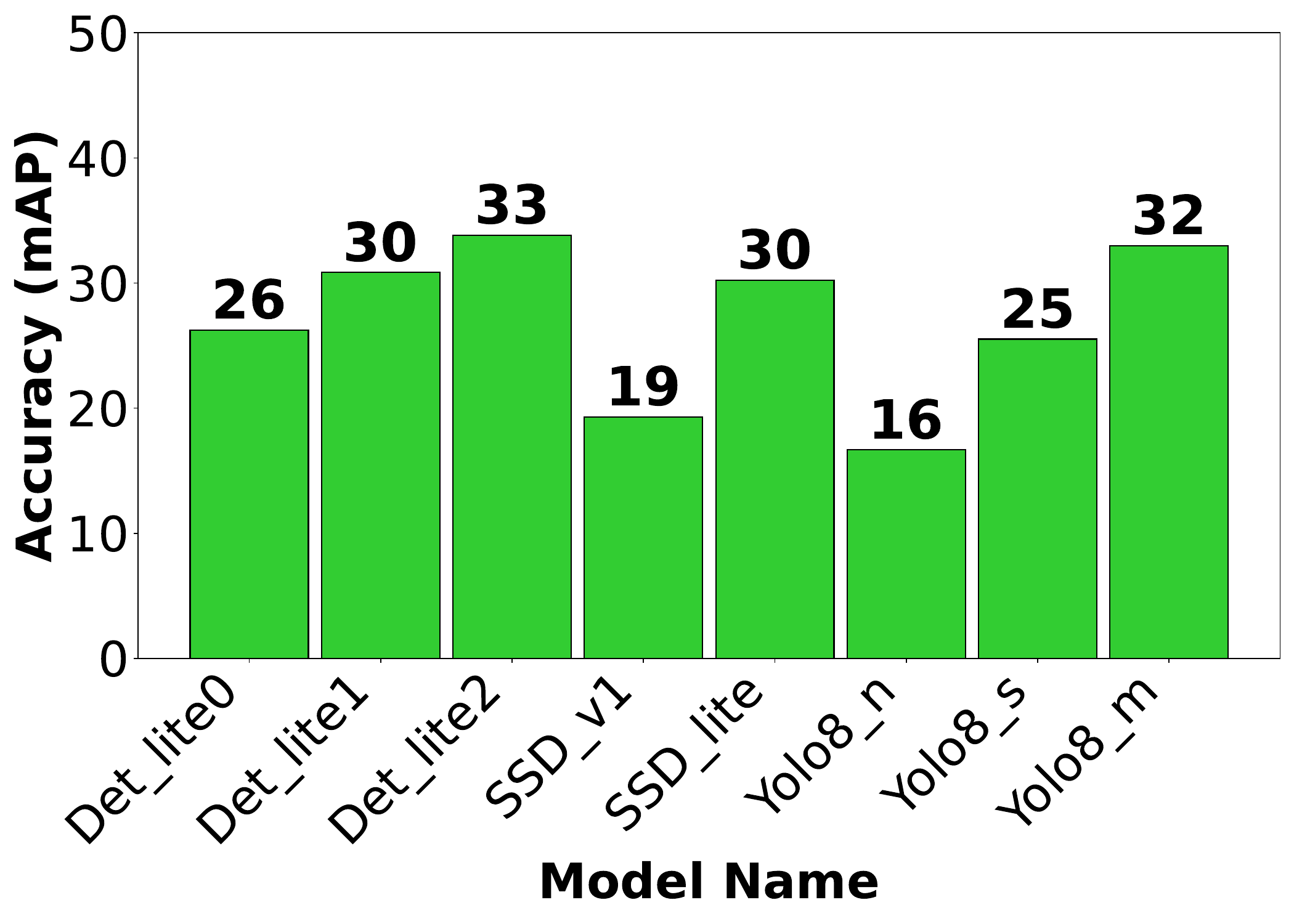}
    }
    \subfigure[Pi + AI HAT]{
        \includegraphics[width=0.22\textwidth]{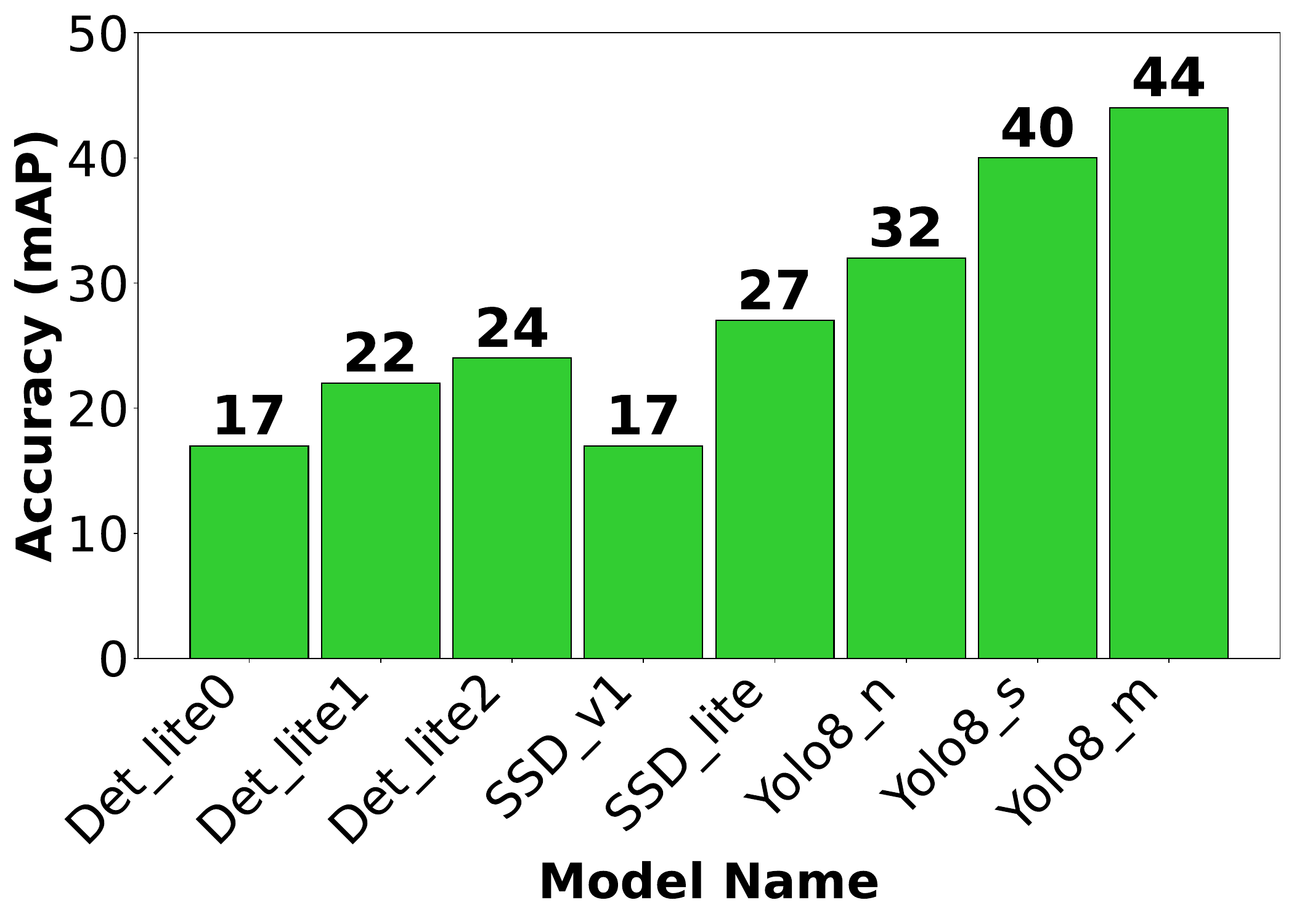}
    }
    \subfigure[NVIDIA]{
        \includegraphics[width=0.22\textwidth]{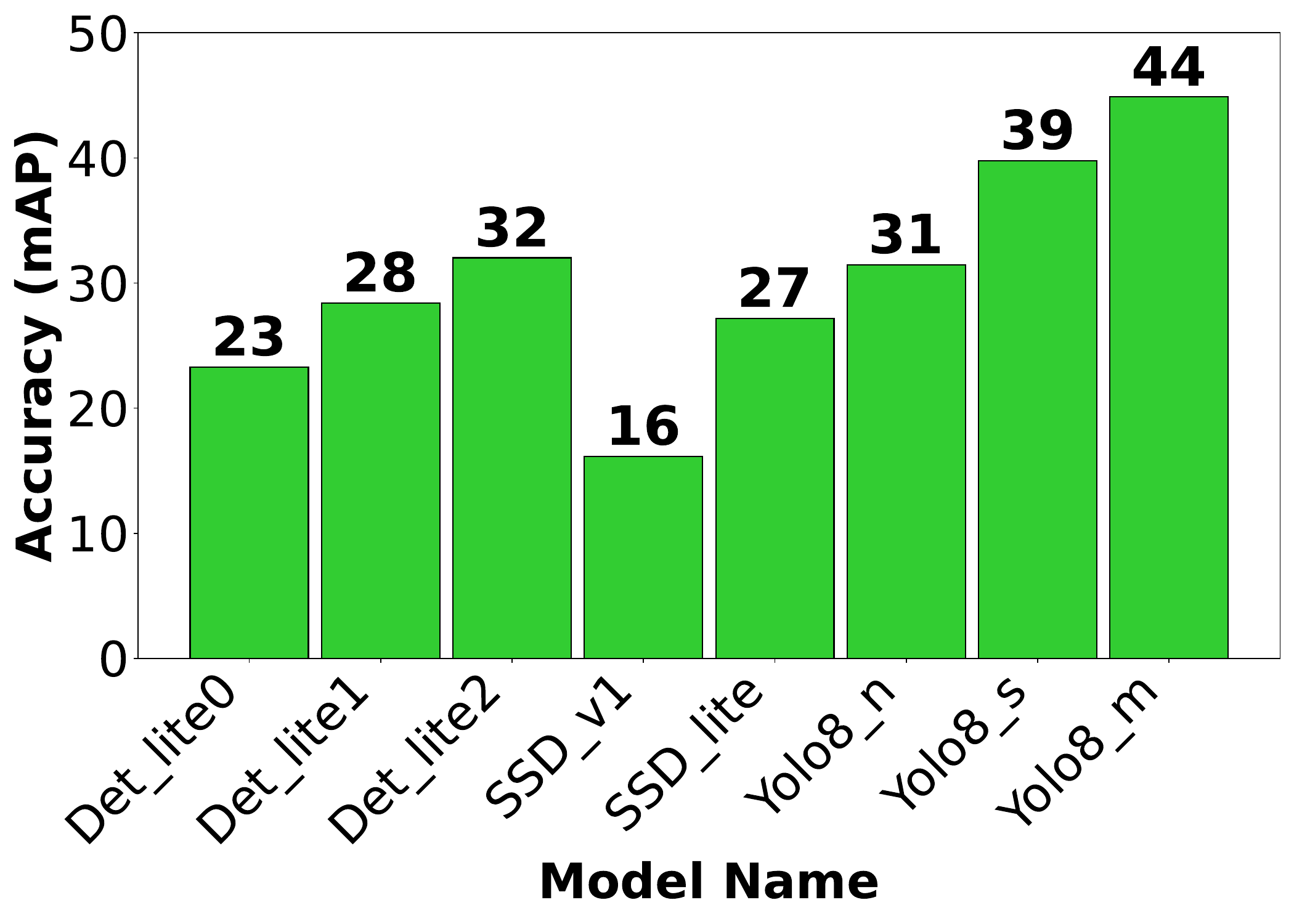}
    }

    \caption{Accuracy (mAP) for different edge devices.}
    \label{fig:mAP for devices}
\end{figure}

This subsection presents the accuracy results of the evaluated object detection models in terms of mAP. On the Raspberry Pi platforms, as shown in Fig.~\ref{fig:mAP for devices}(a), the accuracy varies noticeably across model families and model sizes. Among the evaluated models, SSD\_v1 records the lowest mAP at 19, whereas YOLOv8\_m achieves the highest mAP at 44. The EfficientDet Lite models exhibit intermediate accuracy levels, with mAP values ranging from 26 to 33. SSD\_lite and YOLOv8\_n achieve similar mAP values around 30, while YOLOv8\_s attains a higher mAP of 39.

For the Raspberry Pi devices equipped with TPU accelerators, as shown in Fig.~\ref{fig:mAP for devices}(b), the EfficientDet Lite and SSD model families maintain accuracy levels comparable to those observed on the CPU-based Raspberry Pi platforms. In contrast, the YOLOv8 models experience a noticeable reduction in accuracy under TPU-based deployment. Specifically, YOLOv8\_n decreases to an mAP of 16, YOLOv8\_s decreases to 25, and YOLOv8\_m decreases to 32. This reduction is consistent with the TPU deployment setting, in which the YOLOv8 models are executed with a reduced input size.

On Raspberry Pi 5 with AI HAT+, as shown in Fig.~\ref{fig:mAP for devices}(c), YOLOv8\_m again achieves the highest mAP at 44, while YOLOv8\_s and YOLOv8\_n record 40 and 32, respectively. SSD\_v1 records 17, SSD\_lite records 27, and the EfficientDet Lite models range from 17 to 24. Compared with the CPU-based Raspberry Pi results, the YOLOv8 models retain competitive accuracy on this platform, whereas EfficientDet Lite and SSD\_v1 show lower mAP values.

For the NVIDIA platforms, Jetson Nano and Jetson Orin Nano exhibit the same accuracy results in our setup, as shown in Fig.~\ref{fig:mAP for devices}(d). The YOLOv8 models follow a similar pattern to the Raspberry Pi platforms, with mAP values ranging from 31 to 44. In particular, YOLOv8\_m achieves the highest mAP at 44, while YOLOv8\_n and YOLOv8\_s record 31 and 39, respectively. In contrast, the EfficientDet Lite and SSD models exhibit slightly lower mAP values than those observed on the Raspberry Pi platforms. Specifically, SSD\_v1 records the lowest mAP at 16, SSD\_lite records 27, and EfficientDet Lite0, Lite1, and Lite2 record 23, 28, and 32, respectively.

\begin{keyinsight}
For the same detector family, accuracy remains broadly consistent when the deployment preserves similar input resolution and numerical precision. This explains why YOLOv8 maintains strong accuracy on CPU Raspberry Pi, Hailo AI HAT+, and Jetson deployments. In contrast, YOLOv8 accuracy drops on TPU-enabled Raspberry Pis because the TPU deployment requires conversion to TFLite and uses a reduced 320×320 input size. This shows that accuracy should not be compared only by model name; the deployed model format, input resolution, and quantization path must also be considered.
\end{keyinsight}

\subsection{Energy Consumption vs Inference Time}
\begin{figure}[t]
    \centering
    \subfigure[SSD\_v1]{
        \includegraphics[width=0.22\textwidth]{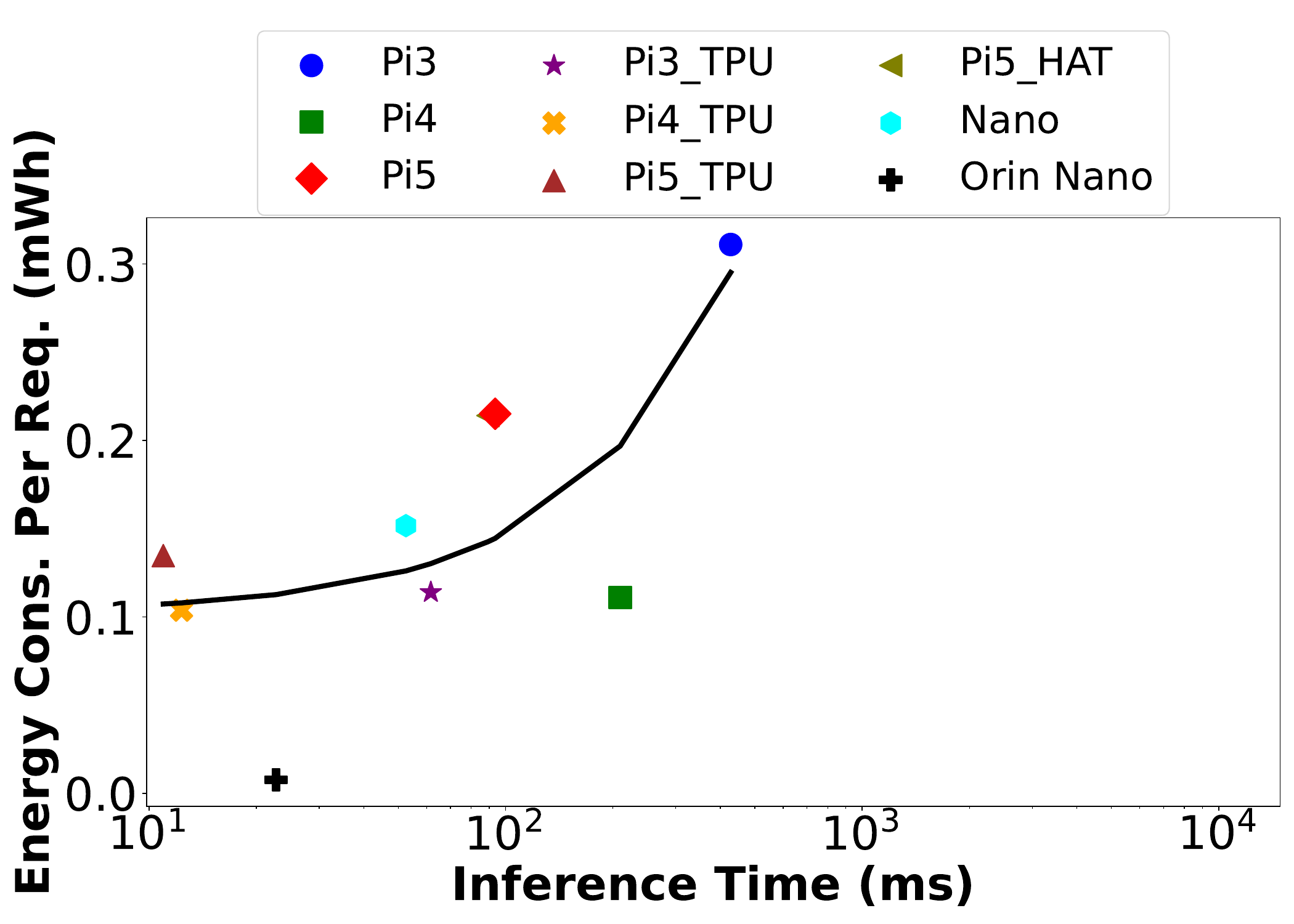}
    }
    \subfigure[SSD\_lite]{
        \includegraphics[width=0.22\textwidth]{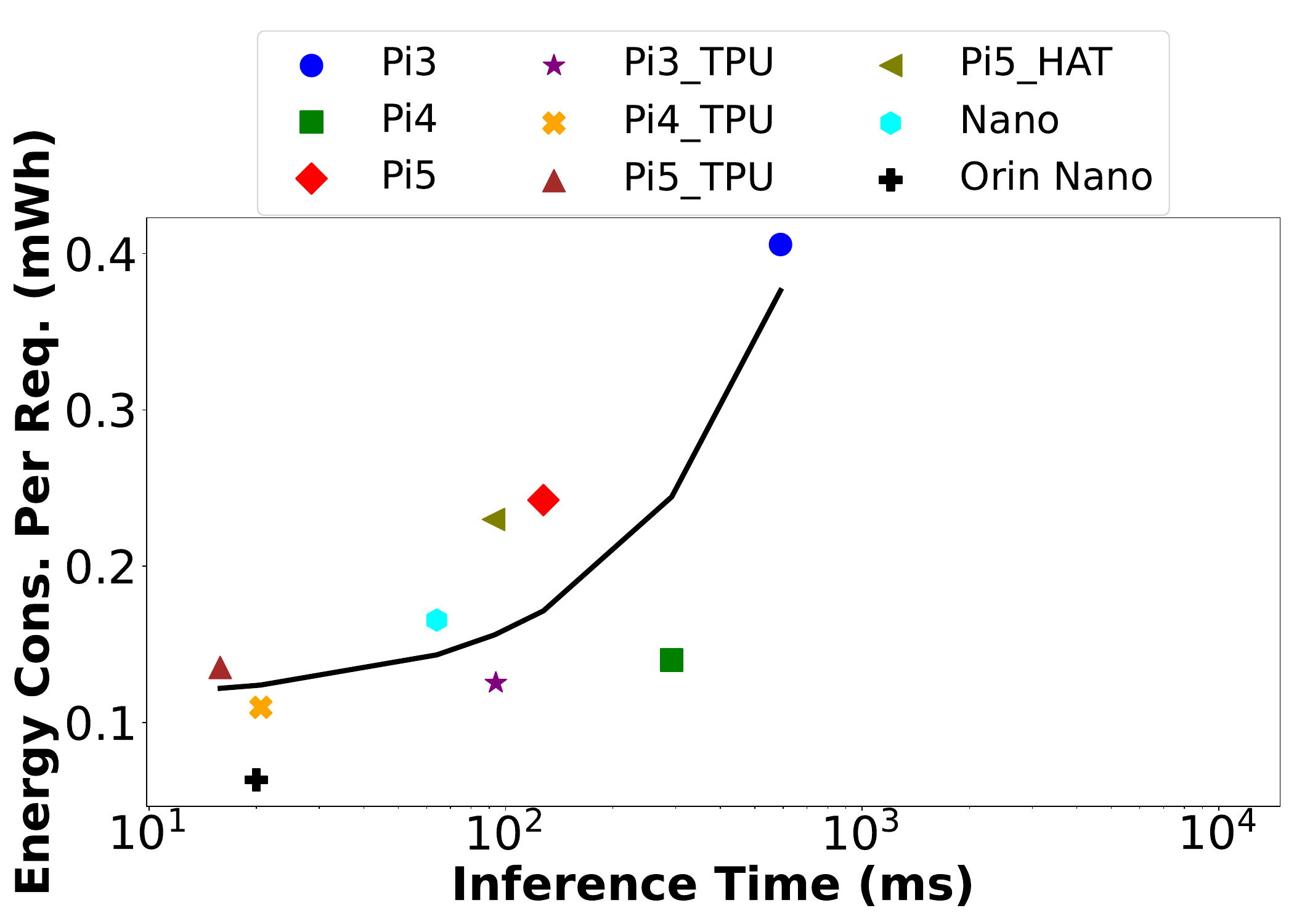}
    }
    \subfigure[Det\_lit0]{
        \includegraphics[width=0.22\textwidth]{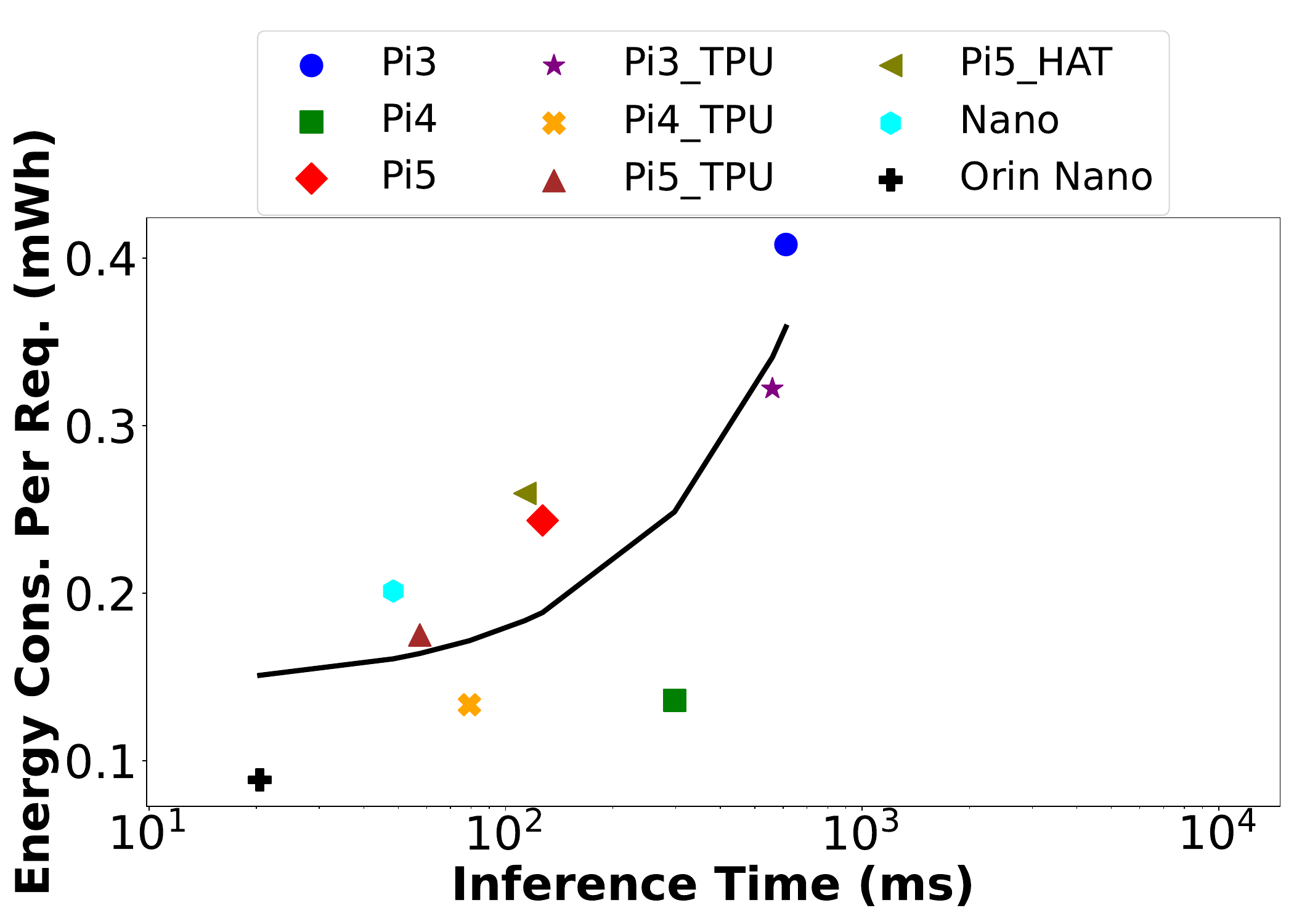}
    }
    \subfigure[Det\_lite1]{
        \includegraphics[width=0.22\textwidth]{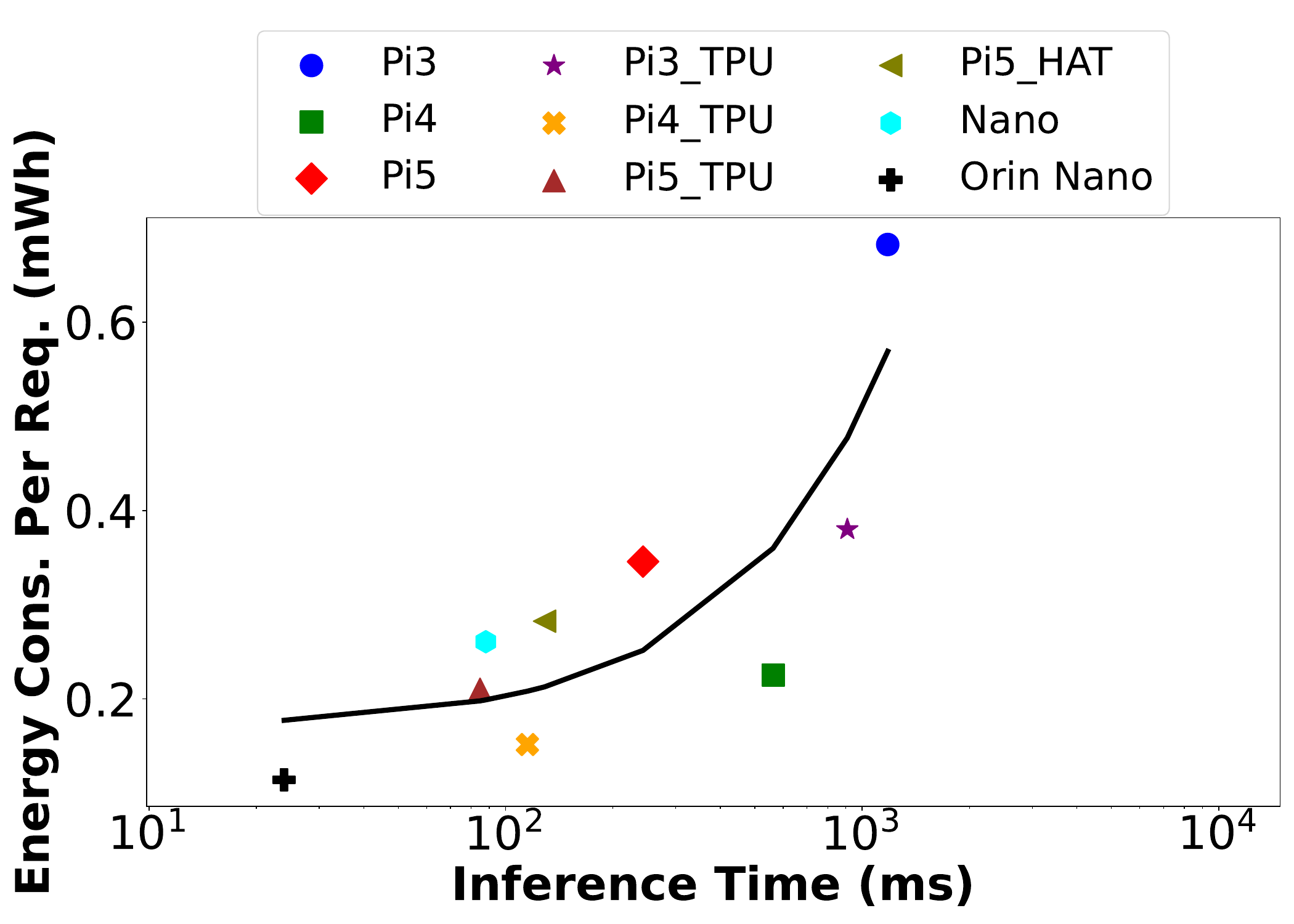}
    }
    \subfigure[Det\_lite2]{
        \includegraphics[width=0.22\textwidth]{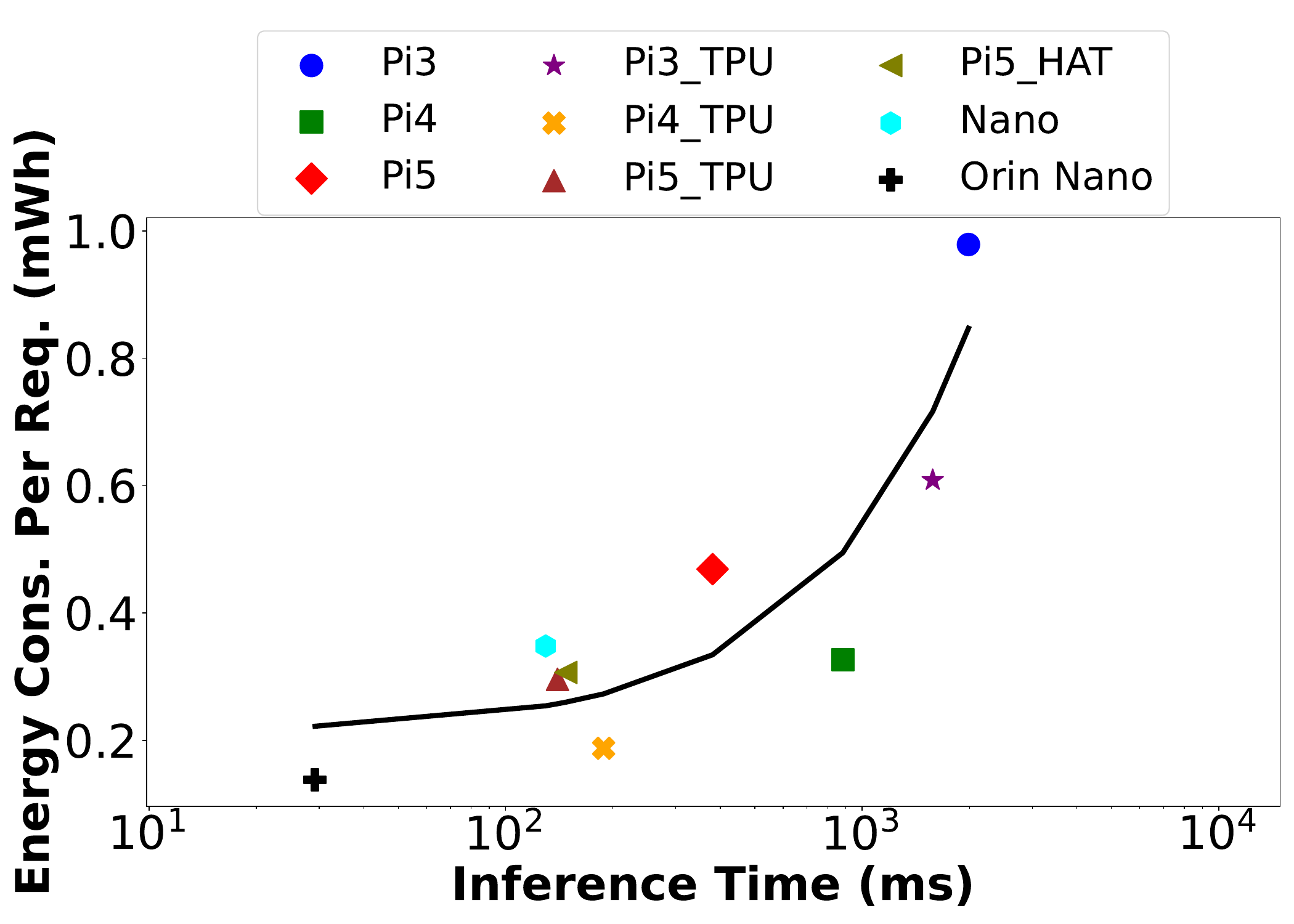}
    }
    \subfigure[Yolo8\_n]{
        \includegraphics[width=0.22\textwidth]{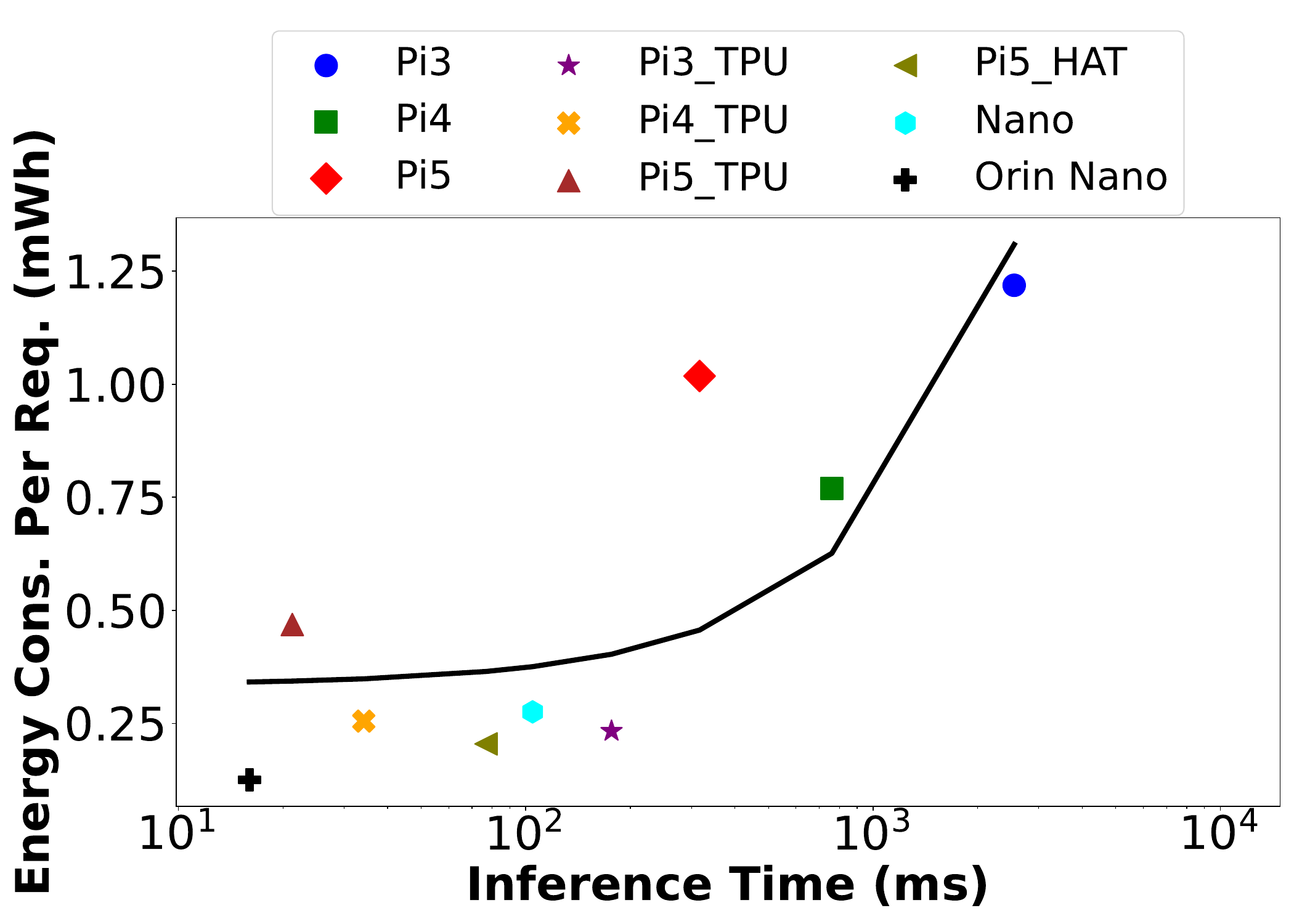}
    }
    \subfigure[Yolo8\_s]{
        \includegraphics[width=0.22\textwidth]{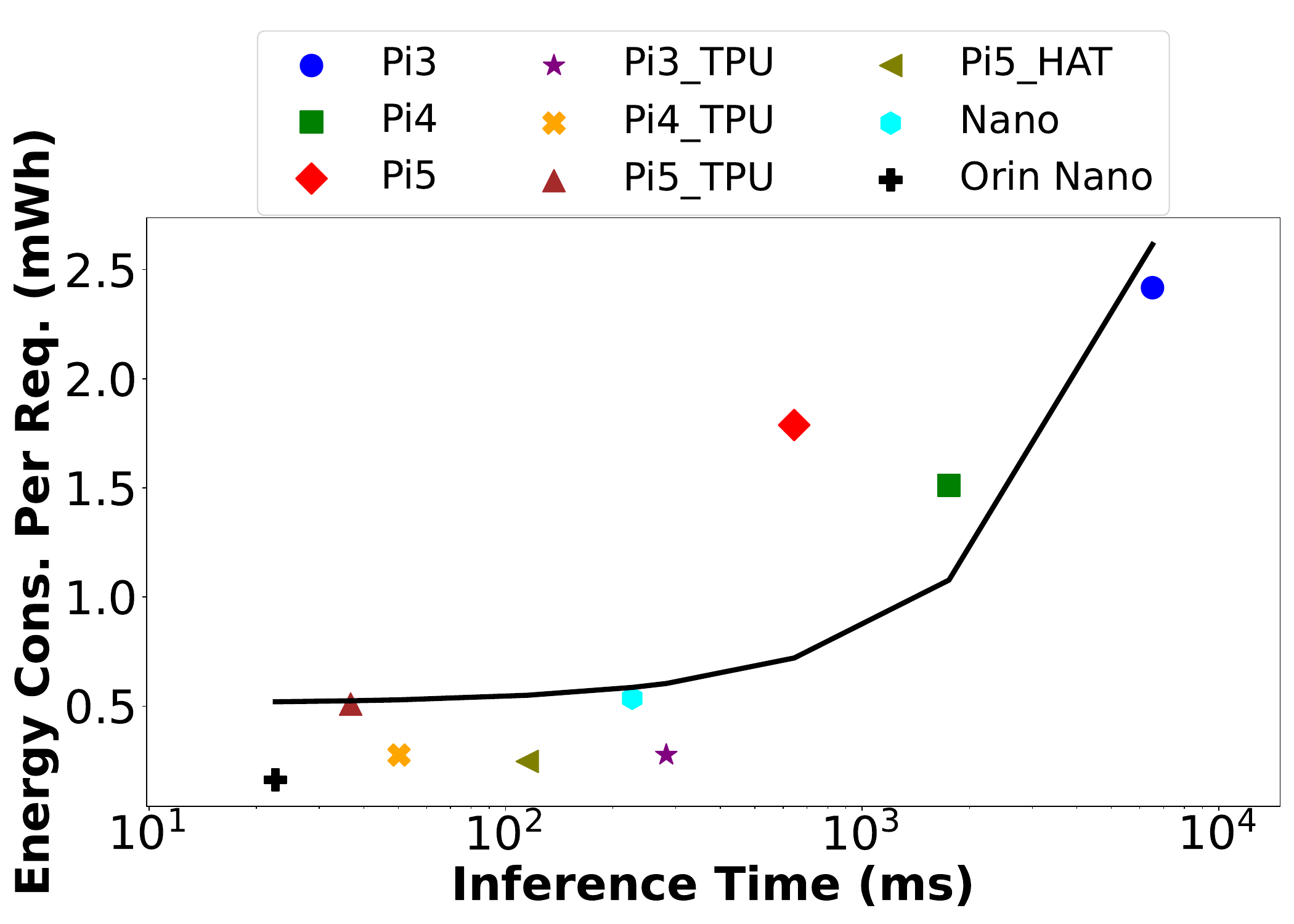}
    }
     \subfigure[Yolo8\_m]{
        \includegraphics[width=0.22\textwidth]{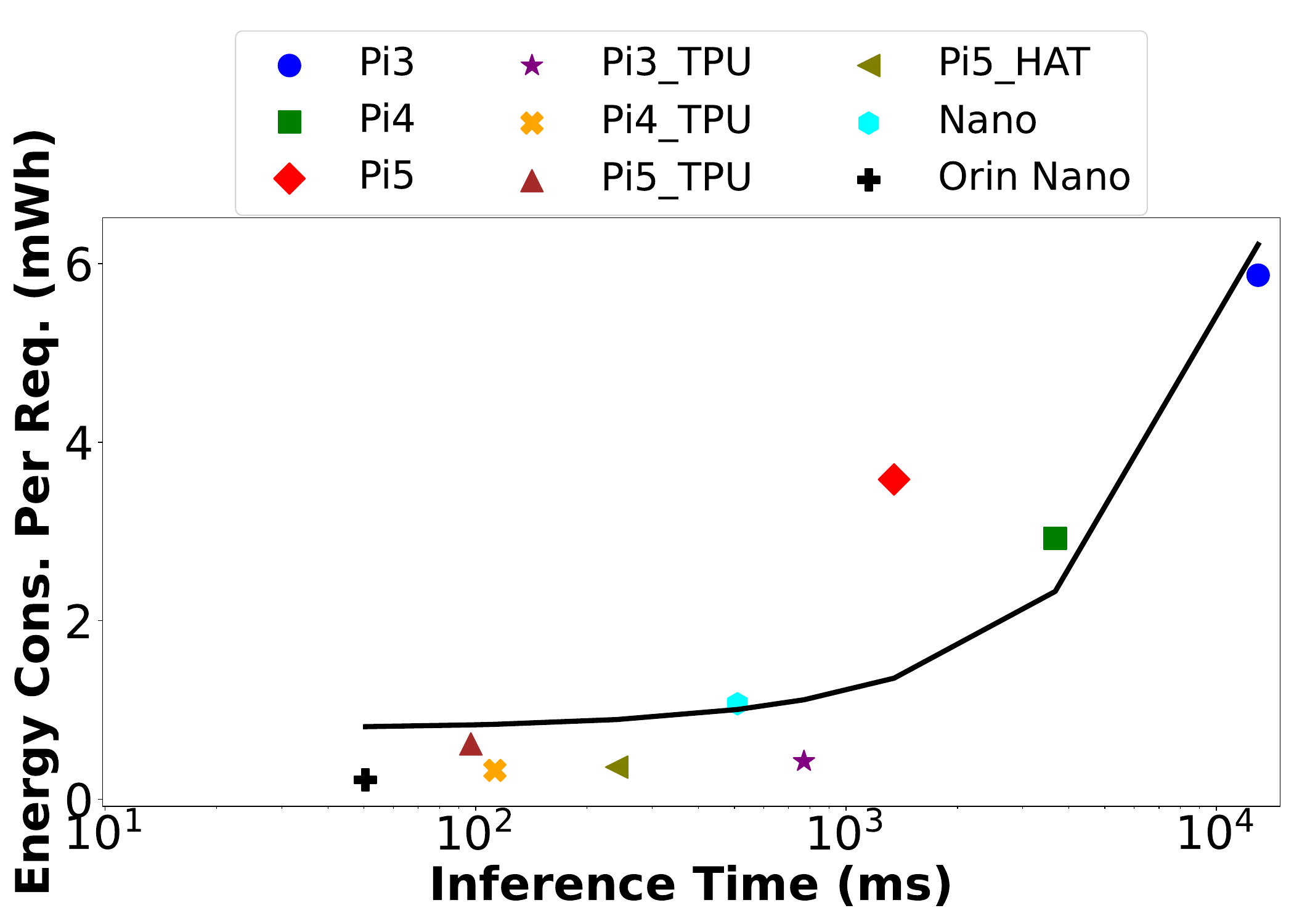}
    }
    \caption{Energy consumption per request (excluding base energy) versus inference time for various object detection models (A fitted linear regression line is shown, which appears curved due to the logarithmic scale of the inference time).}
    
    \label{fig:energy consumption vs inference time} 
\end{figure}

This subsection shows the trade-off between energy consumption per request and inference time across edge devices, which is important for energy-efficient, latency-sensitive deployment.

For the SSD models, as shown in Fig.~\ref{fig:energy consumption vs inference time}(a) and Fig.~\ref{fig:energy consumption vs inference time}(b), energy consumption per request and inference time exhibit a clear positive correlation across the evaluated devices. Raspberry Pi 3 lies at the least favorable end of the trade-off space, with both high latency and high energy consumption. In contrast, Raspberry Pi 4 with TPU, Raspberry Pi 5 with TPU, Jetson Nano, and Jetson Orin Nano occupy the most favorable region of the plots, indicating substantially improved efficiency. Among these platforms, Raspberry Pi 5 with TPU generally provides one of the lowest inference times, whereas Jetson Orin Nano achieves the lowest energy consumption, with Jetson Nano also showing competitive efficiency. Raspberry Pi 3 with TPU provides a noticeable improvement over the CPU-only Raspberry Pi platforms, while Raspberry Pi 5 with AI HAT+, Raspberry Pi 4 and Raspberry Pi 5 without accelerators remain in an intermediate region.

A similar overall relationship is observed for the EfficientDet Lite and YOLOv8 models, as shown in Fig.~\ref{fig:energy consumption vs inference time}(c)--(h). In general, devices with lower inference time also tend to exhibit lower energy consumption per request, indicating a positive correlation between the two metrics across model families. Jetson Orin Nano consistently occupies the most favorable region of the trade-off space, while Jetson Nano and the accelerator-enabled Raspberry Pi platforms also achieve competitive results. Raspberry Pi 3 again shows the least favorable behavior, with both high latency and high energy consumption. Raspberry Pi 5 with AI HAT+ achieves competitive results, particularly for the YOLOv8 models, where it clearly outperforms the CPU-only Raspberry Pi platforms and provides a favorable compromise between inference time and energy consumption. At the same time, Raspberry Pi 5 occasionally deviates slightly from the overall trend for the YOLOv8 models relative to the fitted regression line.

\begin{keyinsight}
    Across the evaluated models, energy consumption per request and inference time are strongly correlated, indicating that faster execution often leads to lower request-level energy usage.
\end{keyinsight}

\subsection{Energy Consumption vs Accuracy}
\begin{figure}[t]
     \centering
     \subfigure[SSD\_v1]{
         \includegraphics[width=0.22\textwidth]{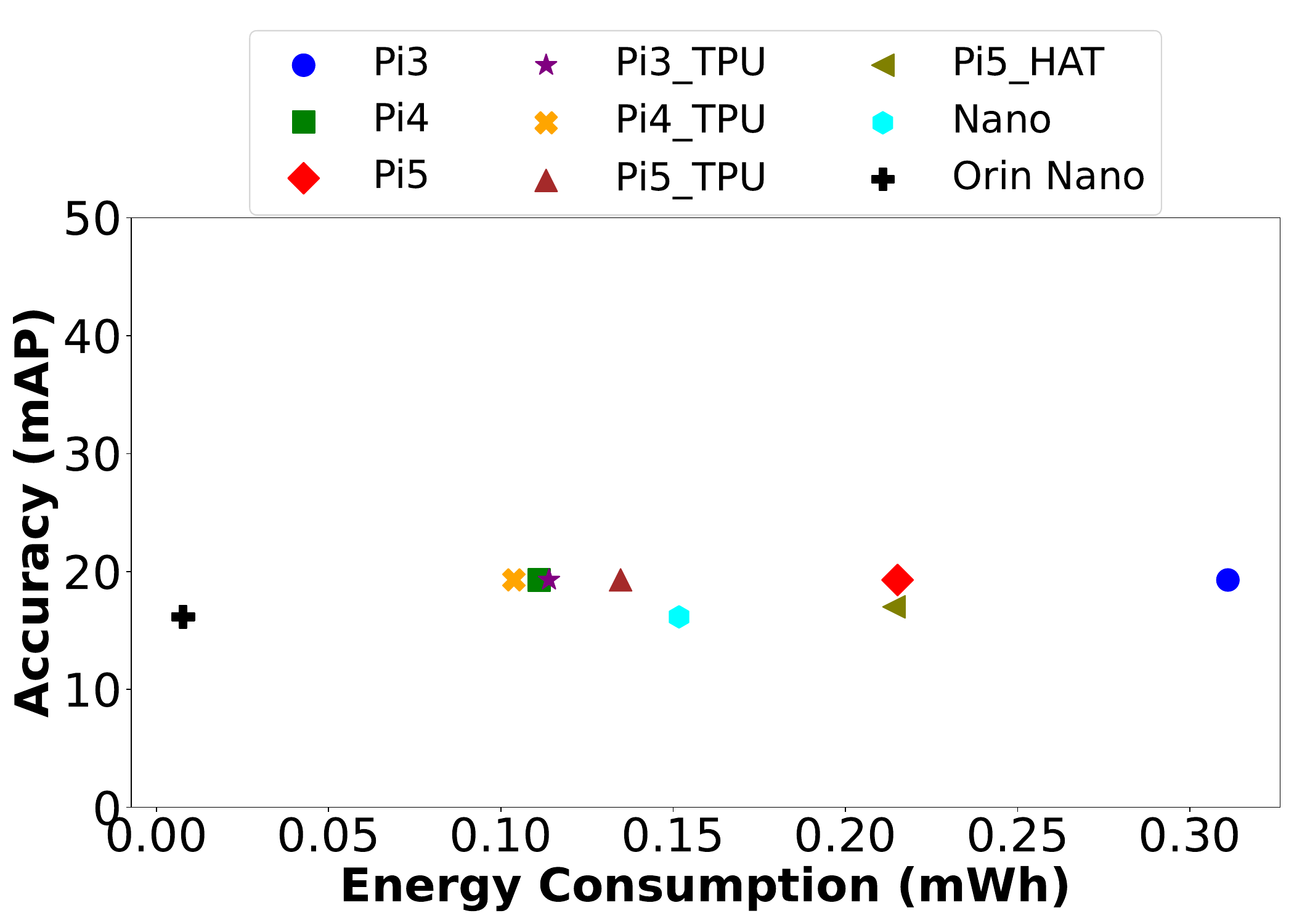}
     }
     \subfigure[SSD\_lite]{
         \includegraphics[width=0.22\textwidth]{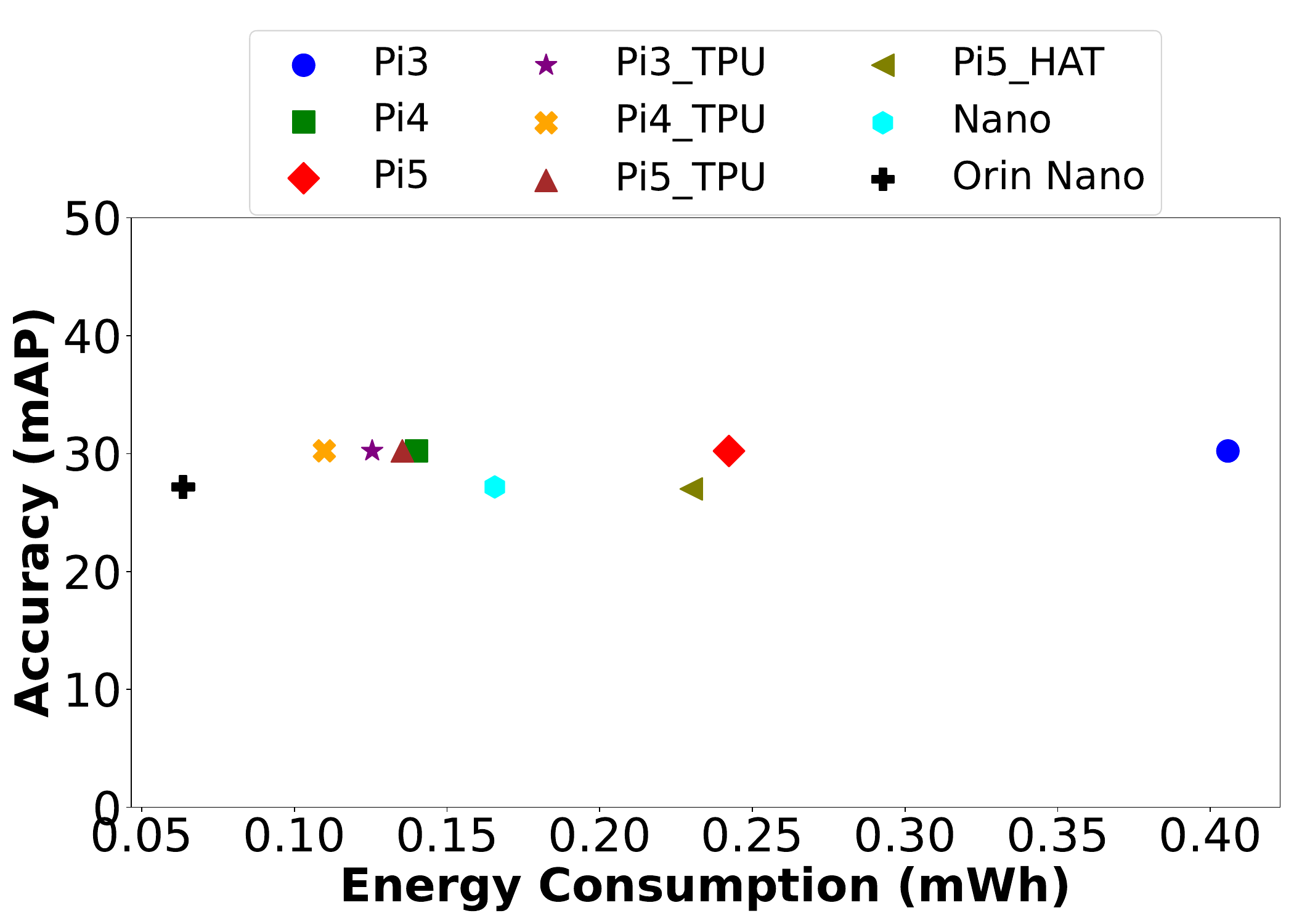}
     }
     \subfigure[Det\_lit0]{
         \includegraphics[width=0.22\textwidth]{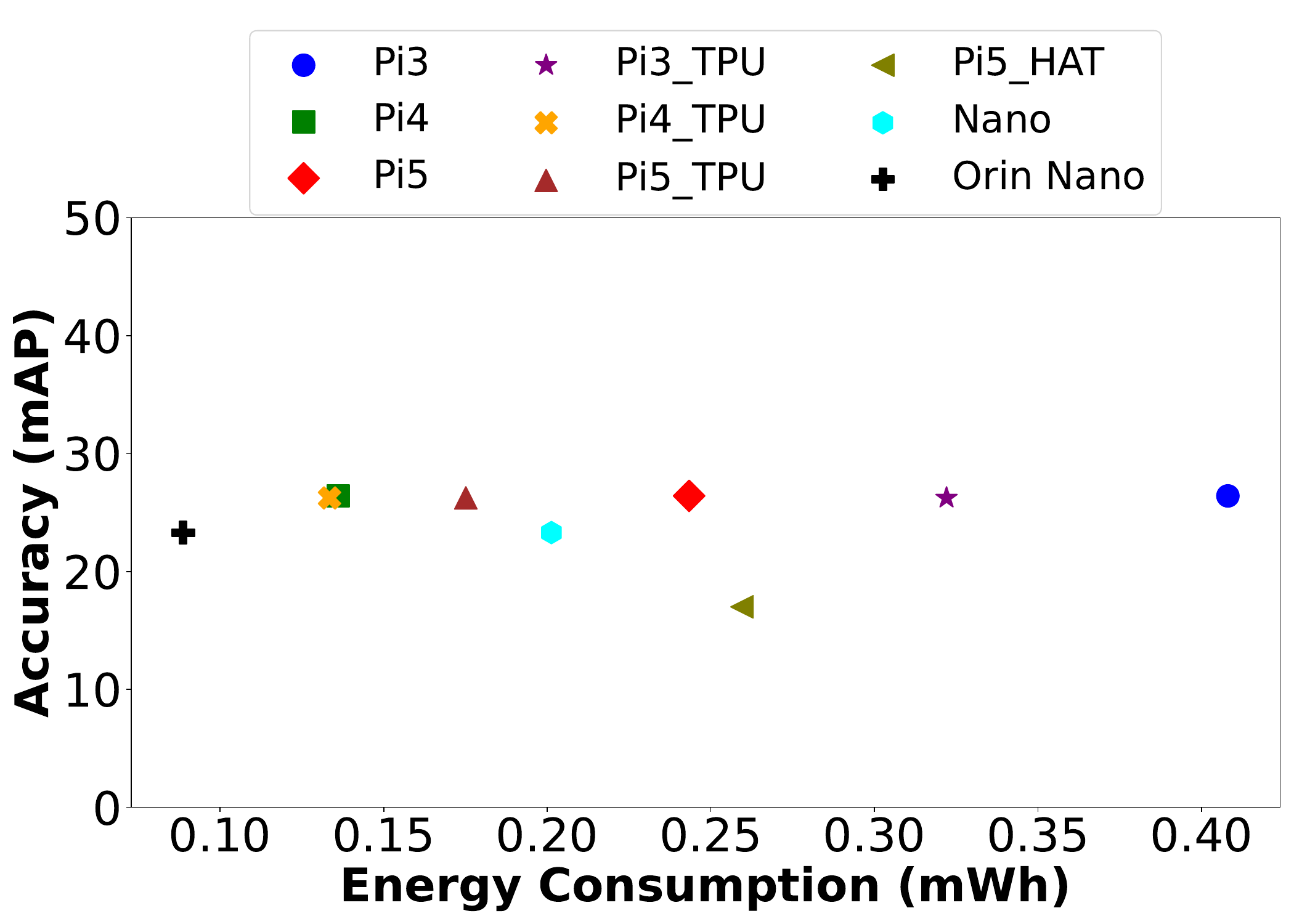}
     }
     \subfigure[Det\_lite1]{
         \includegraphics[width=0.22\textwidth]{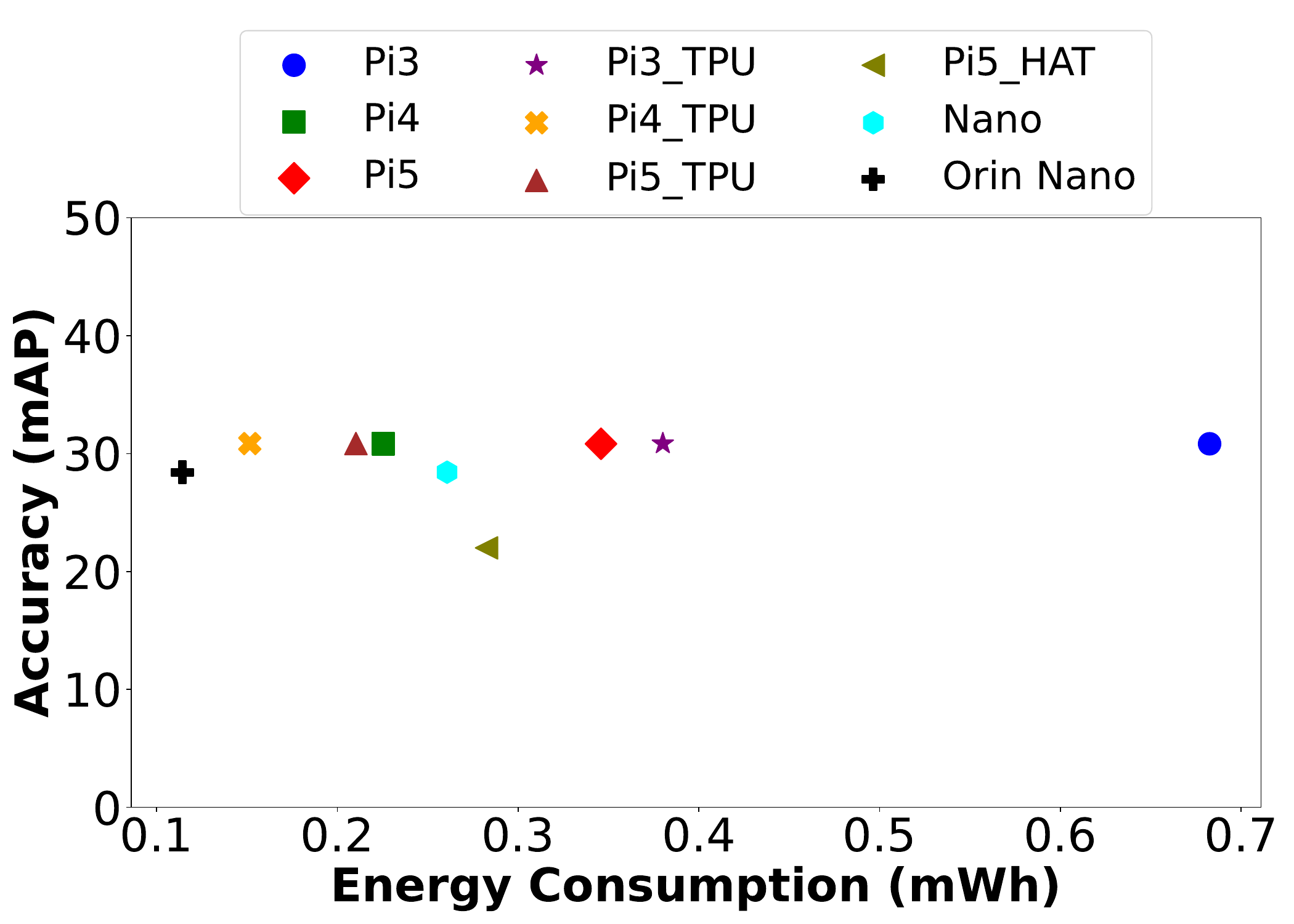}
     }
     \subfigure[Det\_lite2]{
         \includegraphics[width=0.22\textwidth]{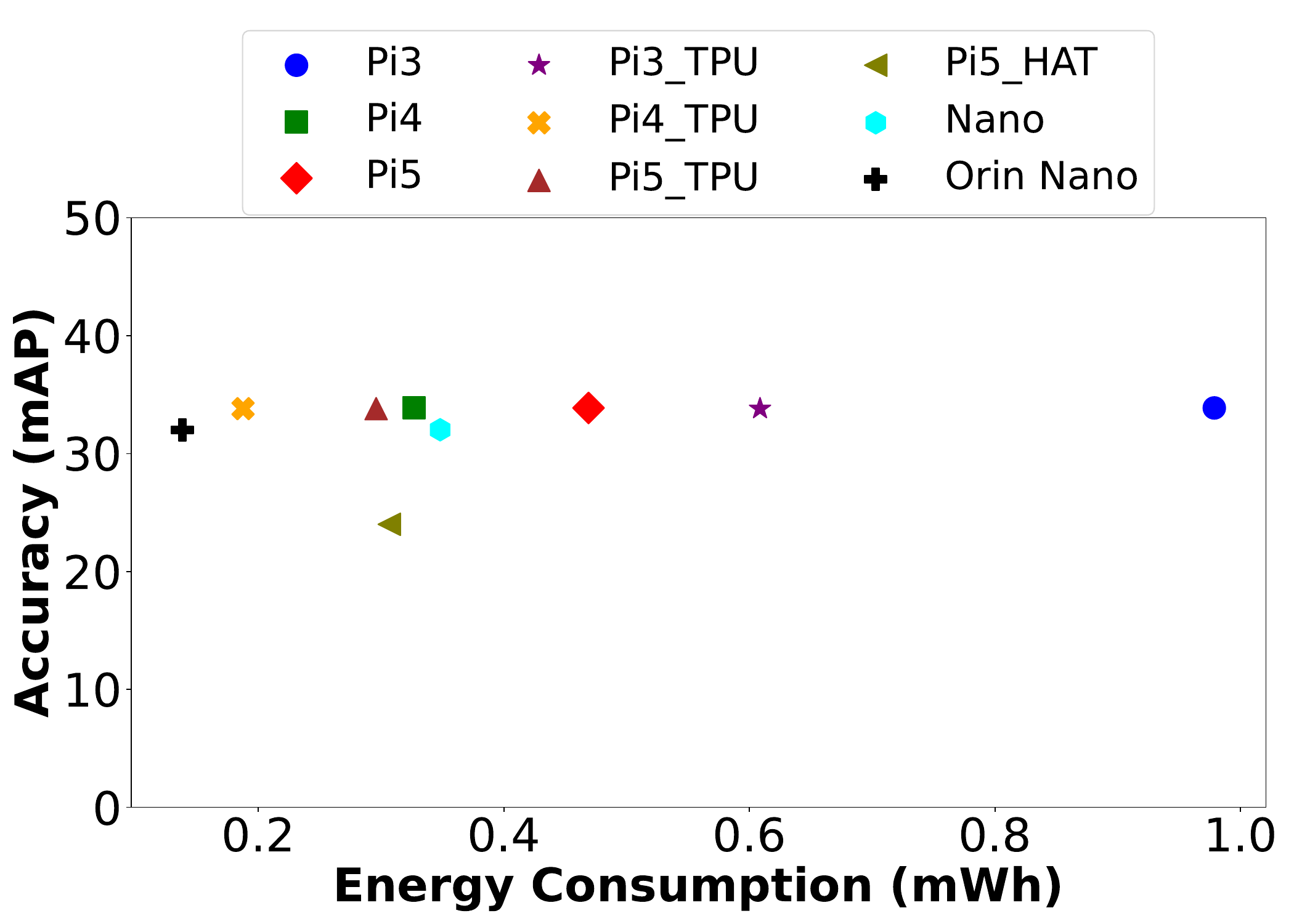}
     }
     \subfigure[Yolo8\_n]{
         \includegraphics[width=0.22\textwidth]{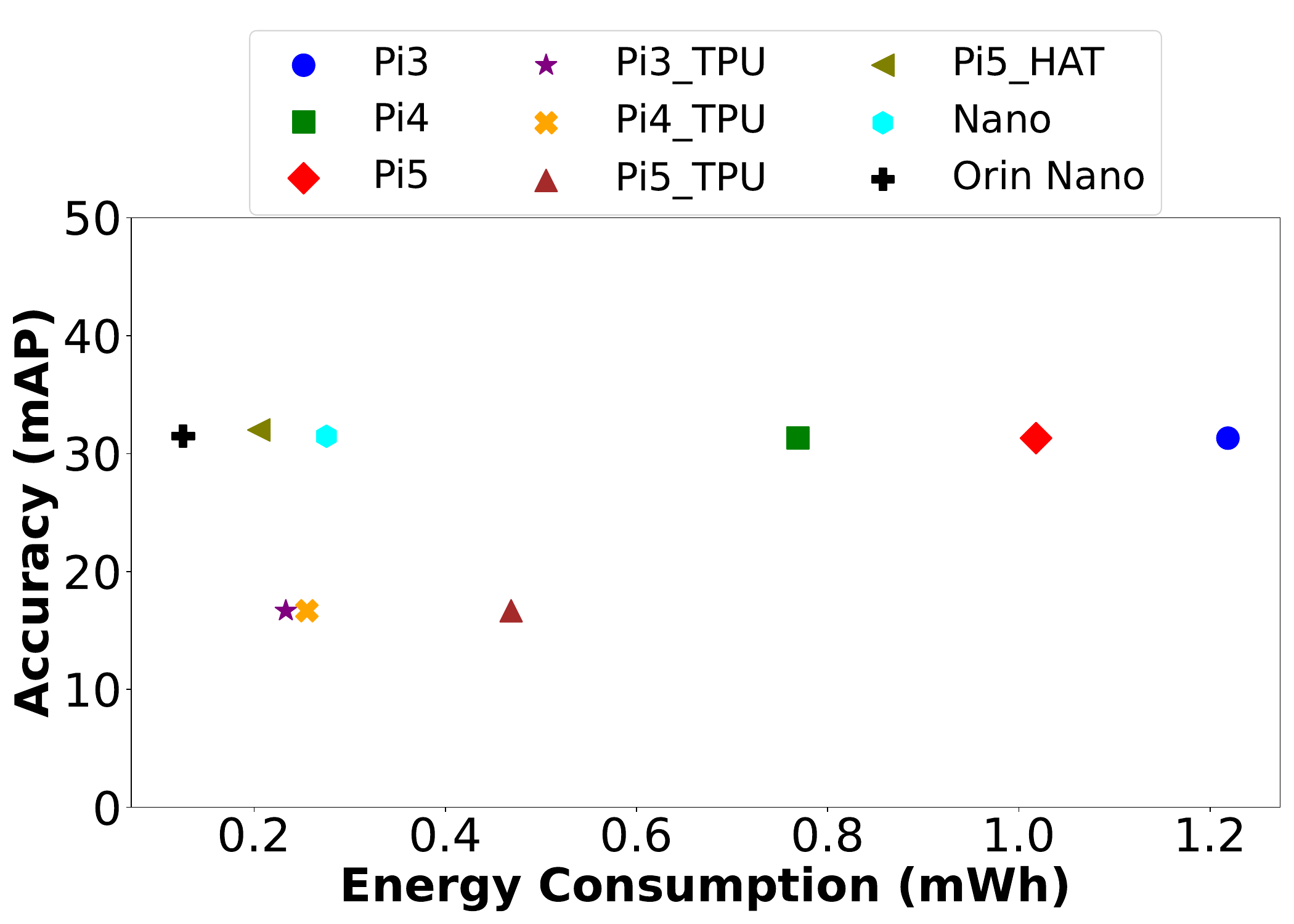}
     }
     \subfigure[Yolo8\_s]{
         \includegraphics[width=0.22\textwidth]{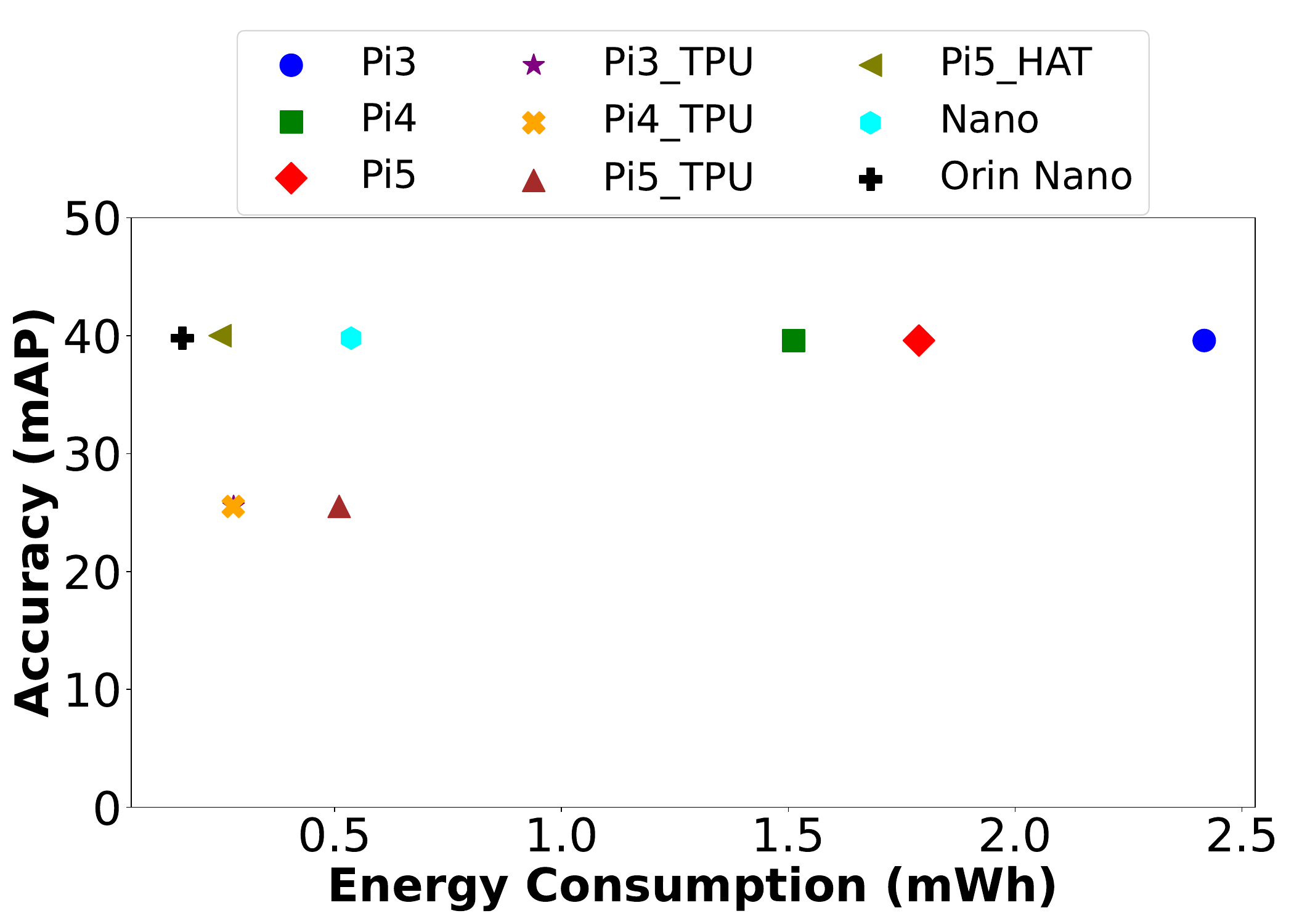}
     }
      \subfigure[Yolo8\_m]{
         \includegraphics[width=0.22\textwidth]{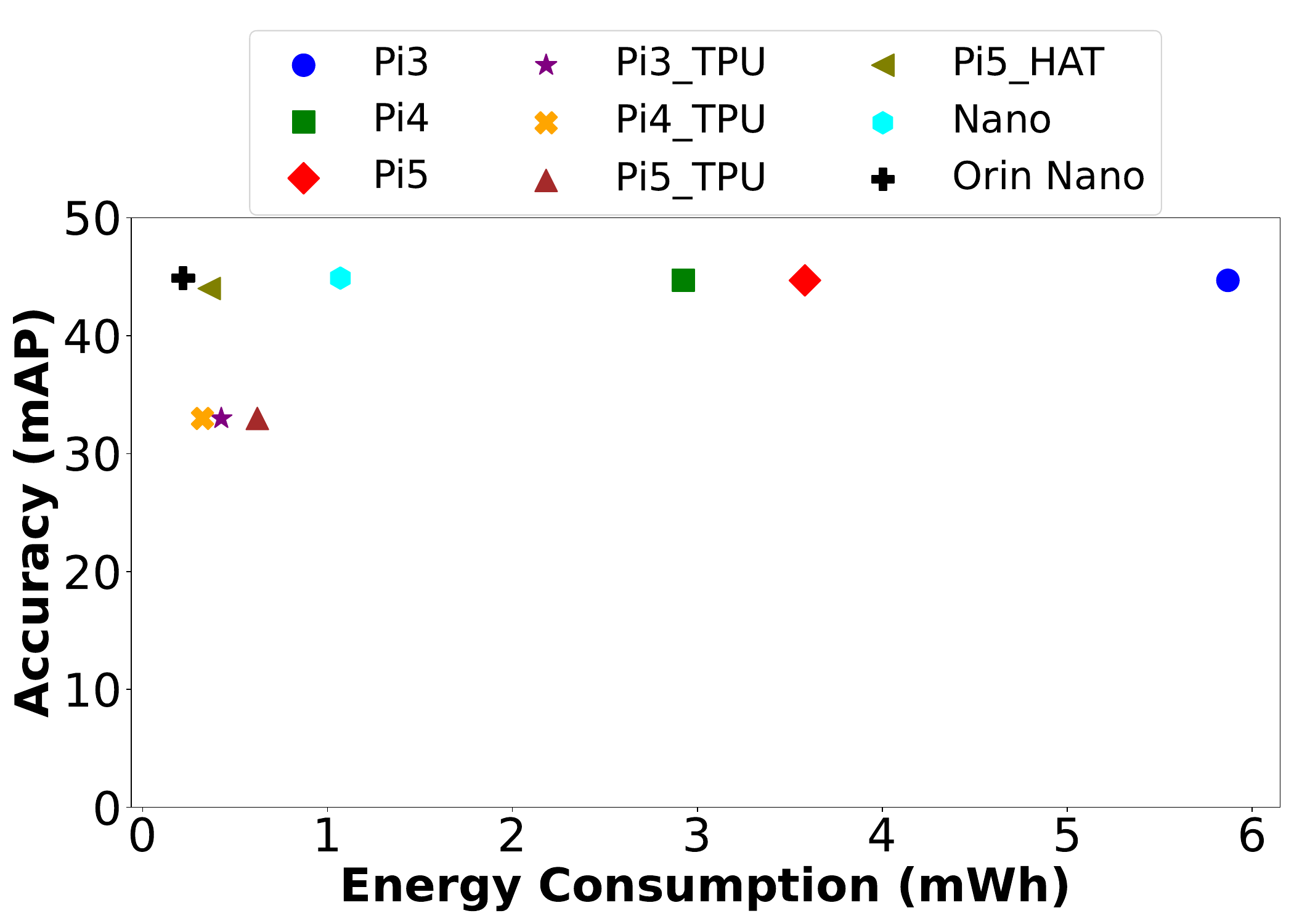}
     }
     \caption{Energy consumption per request (excluding base energy) versus accuracy for various object detection models}
     \label{fig:energy consumption vs accuracy}
 \end{figure}

This subsection highlights the trade-off between energy consumption per request and accuracy across the evaluated edge devices, which is important for improving energy efficiency while maintaining detection accuracy.

For the SSD and EfficientDet Lite models, as shown in Fig.~\ref{fig:energy consumption vs accuracy}(a)--(e), accuracy remains relatively stable across many of the evaluated devices, while the energy consumption per request varies substantially. In these model families, Jetson Orin Nano occupies one of the most favorable regions of the trade-off space because it achieves low energy consumption while maintaining competitive mAP values. In contrast, several CPU-only Raspberry Pi platforms, particularly Raspberry Pi 3 and Raspberry Pi 5, often require higher energy consumption without a clear accuracy advantage. The TPU-based Raspberry Pi platforms generally maintain comparable accuracy while reducing energy consumption, although some accuracy reduction is visible for certain EfficientDet Lite configurations. Raspberry Pi 5 with AI HAT+ typically achieves lower energy consumption than CPU-only Raspberry Pi 5, but for several non-YOLO models this is accompanied by a modest reduction in accuracy.

For the YOLOv8 models, as shown in Fig.~\ref{fig:energy consumption vs accuracy}(f)--(h), accuracy remains stable across most non-TPU devices, while the TPU-based Raspberry Pi configurations show a noticeable reduction in mAP. This behavior is consistent with the reduced input size used for TPU-based deployment. In contrast, Raspberry Pi 5 with AI HAT+ preserves strong YOLOv8 accuracy while achieving substantially lower energy consumption than CPU-only Raspberry Pi 5. Jetson Orin Nano again occupies a highly favorable region of the plots, combining strong accuracy with low energy consumption. Overall, the CPU-only Raspberry Pi platforms tend to be less favorable because they often consume more energy without delivering a corresponding accuracy benefit.

\begin{keyinsight}
There is a clear trade-off between accuracy and energy consumption per request, with higher accuracy generally requiring greater energy. However, some models achieve comparable accuracy with lower energy costs—for example, SSD Lite delivers similar performance to YOLOv8 Nano while consuming less energy per request.
\end{keyinsight} 

\subsection{Inference Time vs Accuracy}
\begin{figure}[t]
     \centering
     \subfigure[SSD\_v1]{\includegraphics[width=0.22\textwidth]{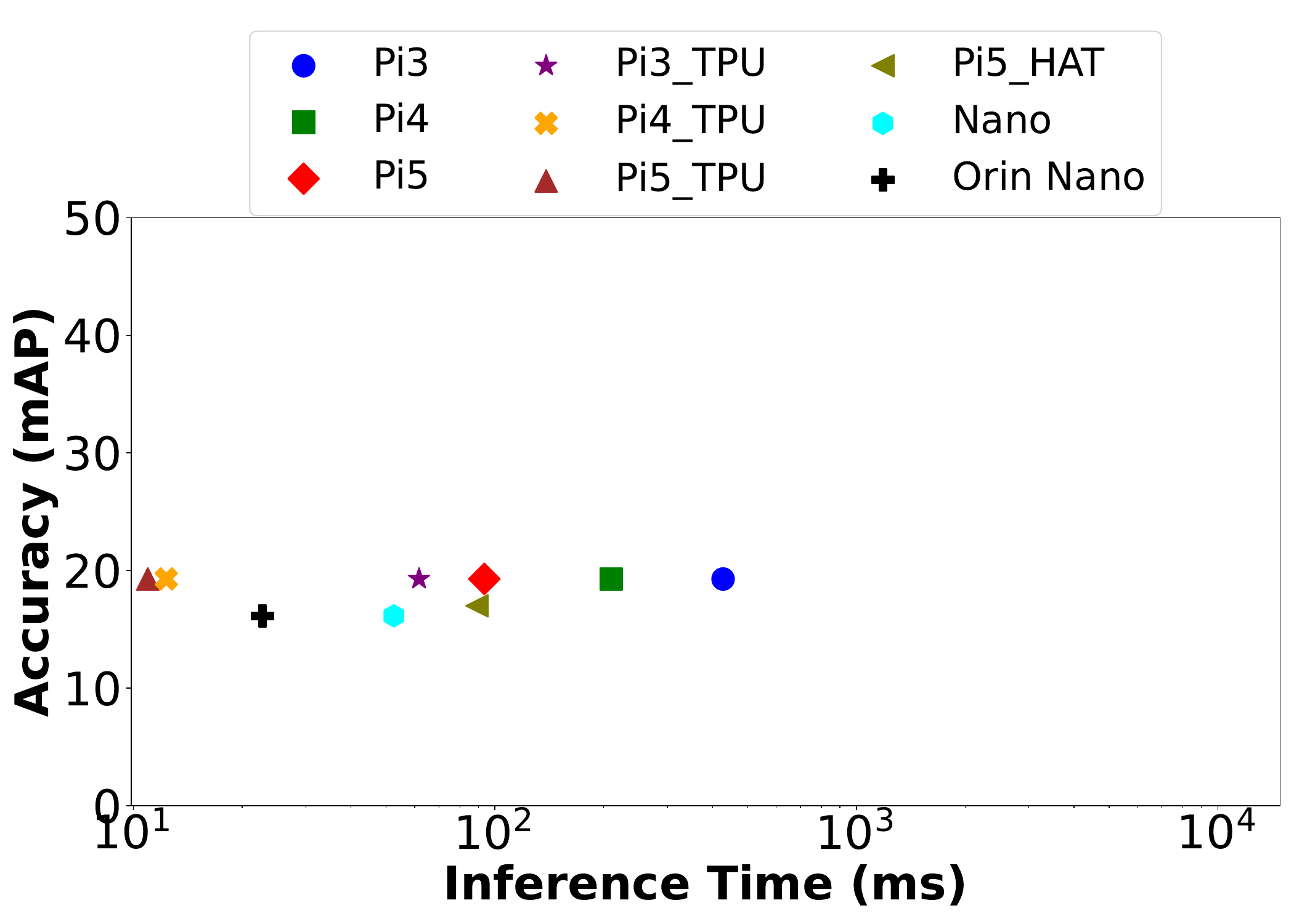}
     }
     \subfigure[SSD\_lite]{
         \includegraphics[width=0.22\textwidth]{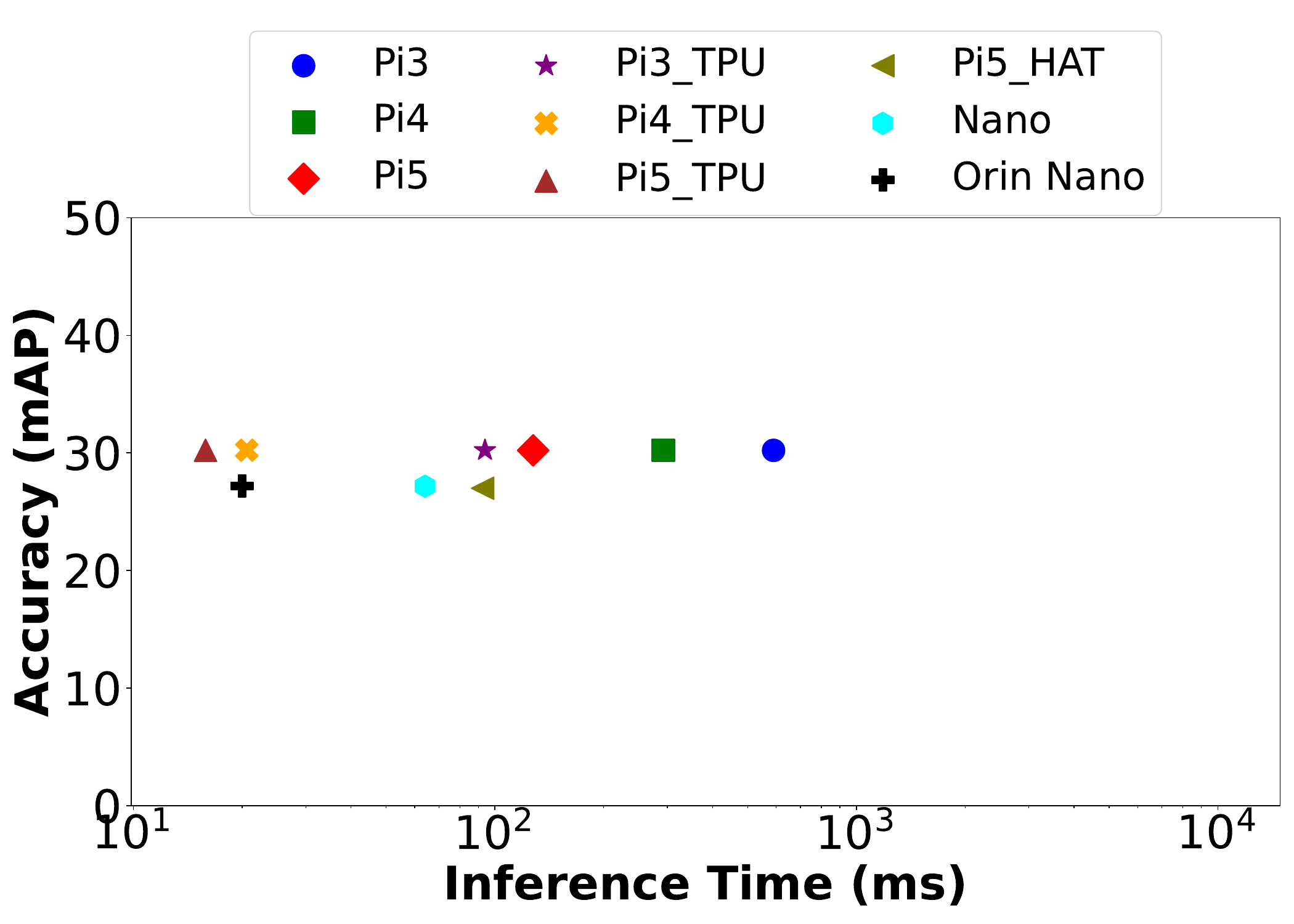}
     }
     \subfigure[Det\_lit0]{
         \includegraphics[width=0.22\textwidth]{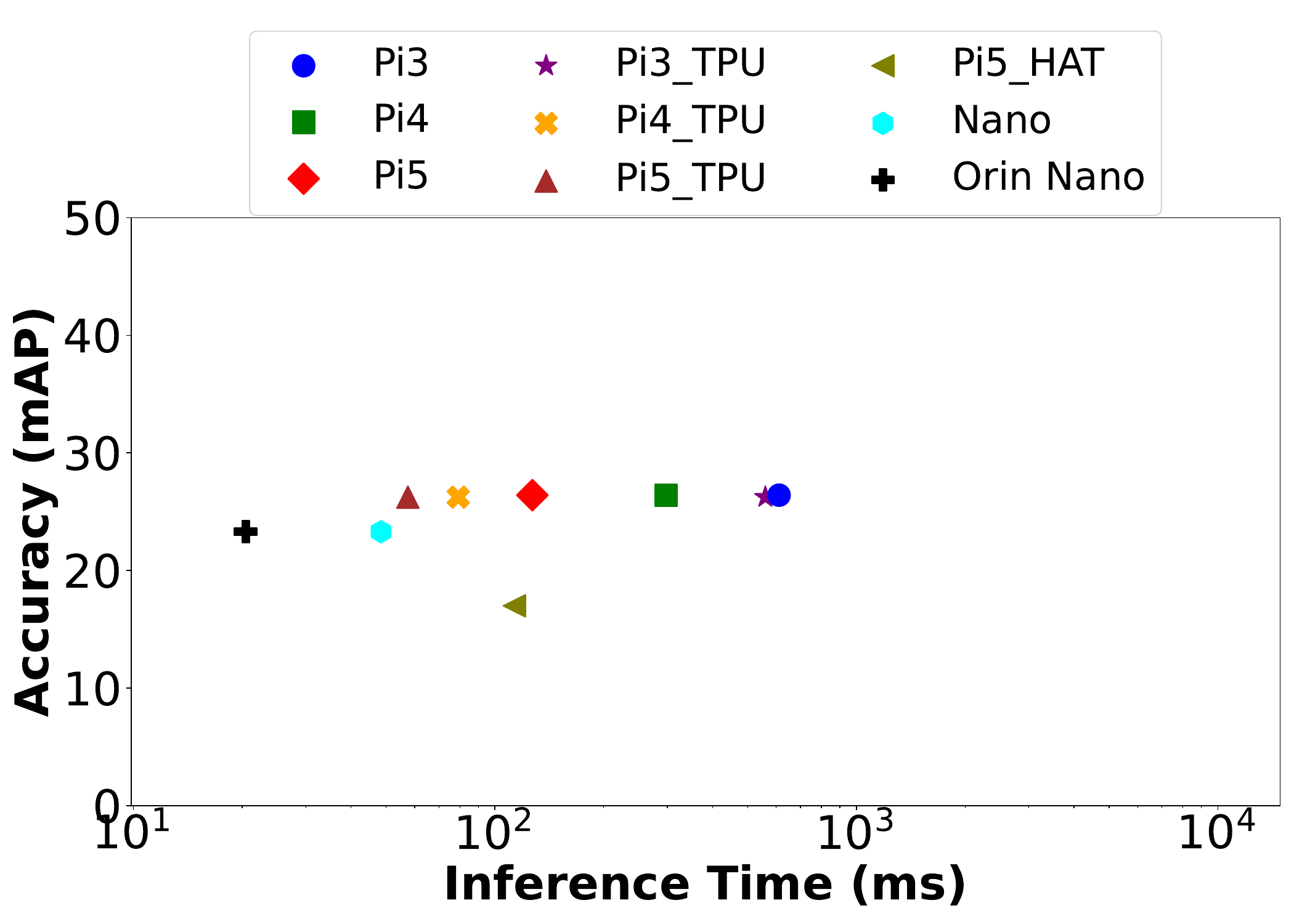}
     }
     \subfigure[Det\_lite1]{
         \includegraphics[width=0.22\textwidth]{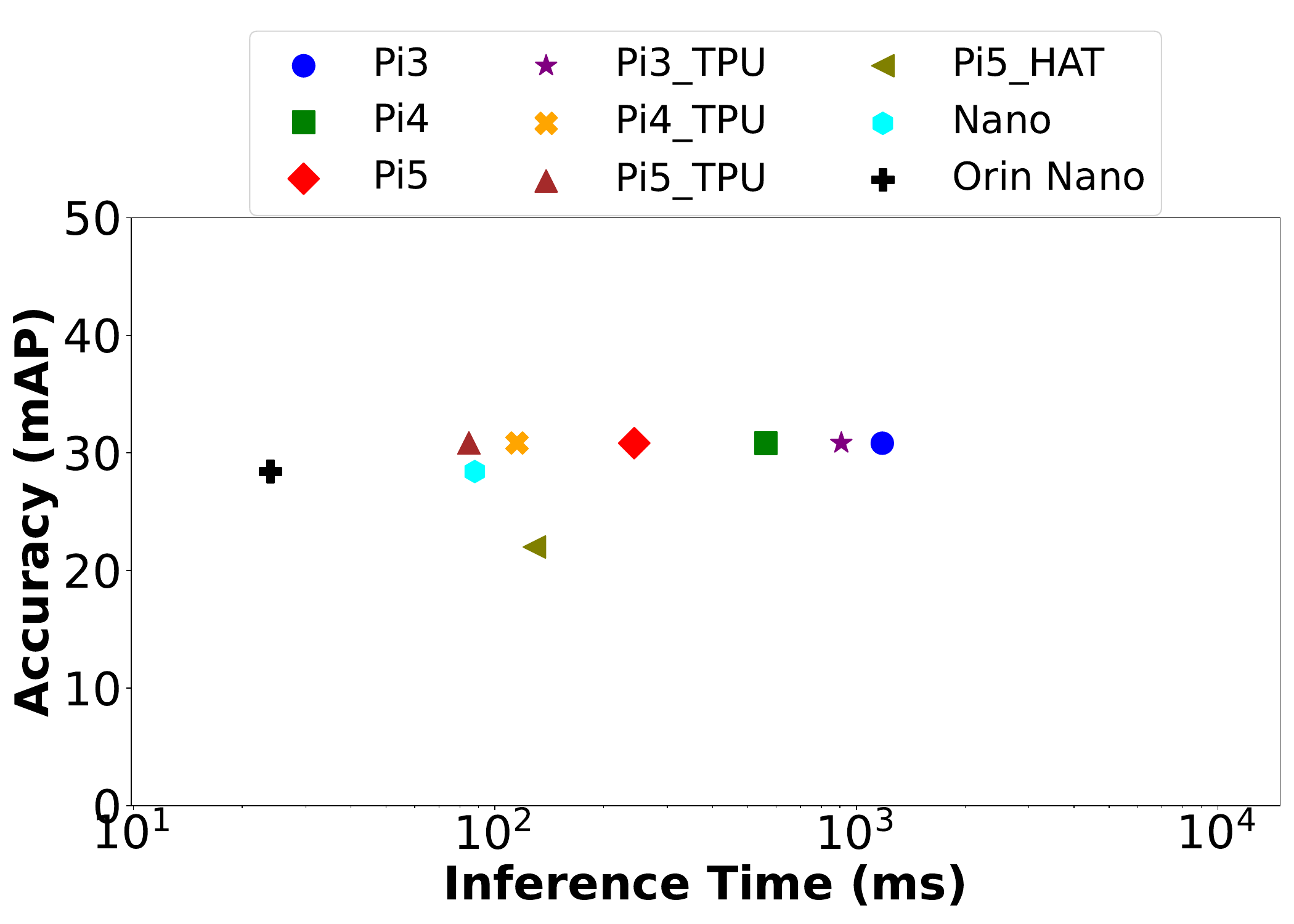}
     }
     \subfigure[Det\_lite2]{
         \includegraphics[width=0.22\textwidth]{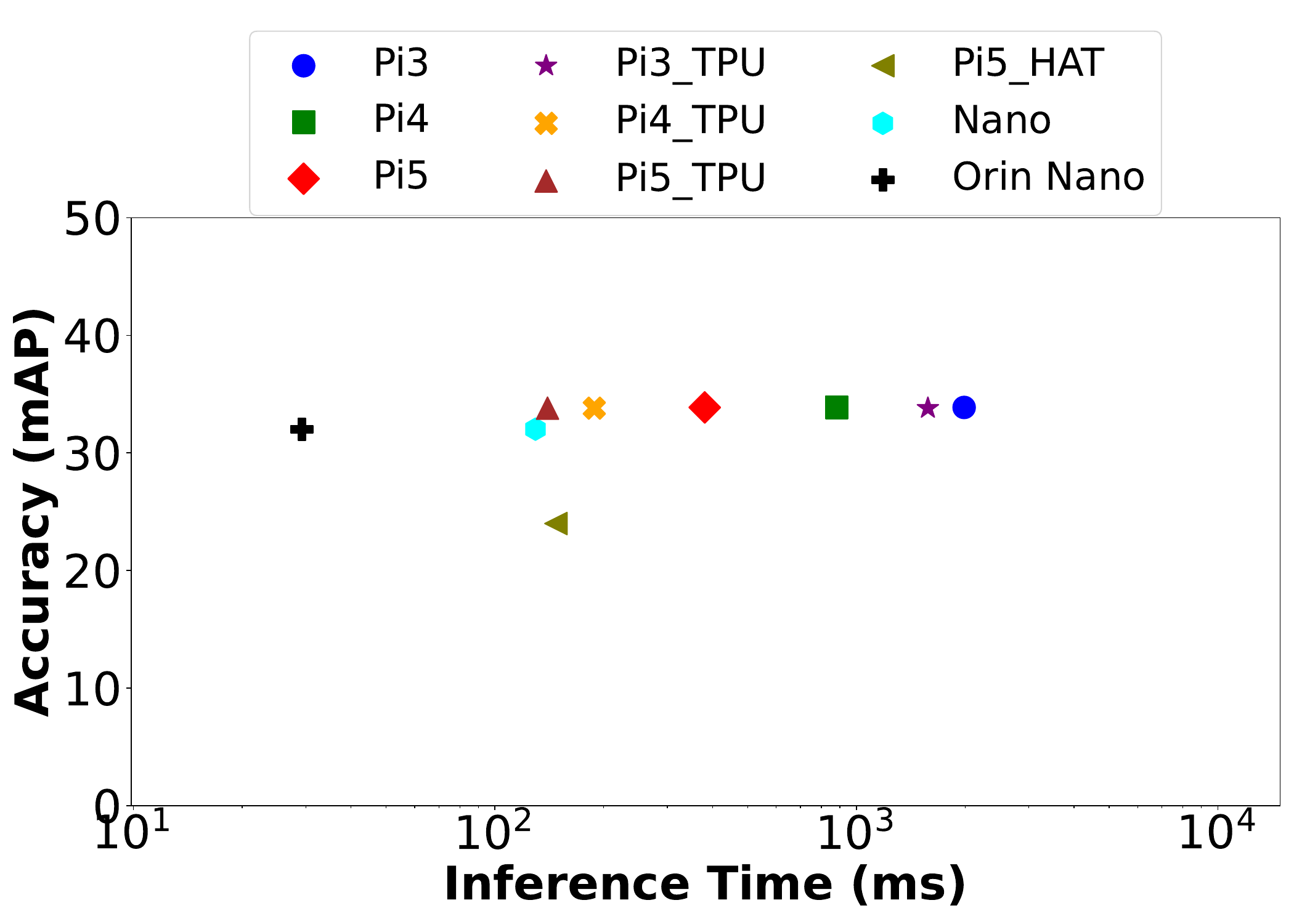}
     }
     \subfigure[Yolo8\_n]{
         \includegraphics[width=0.22\textwidth]{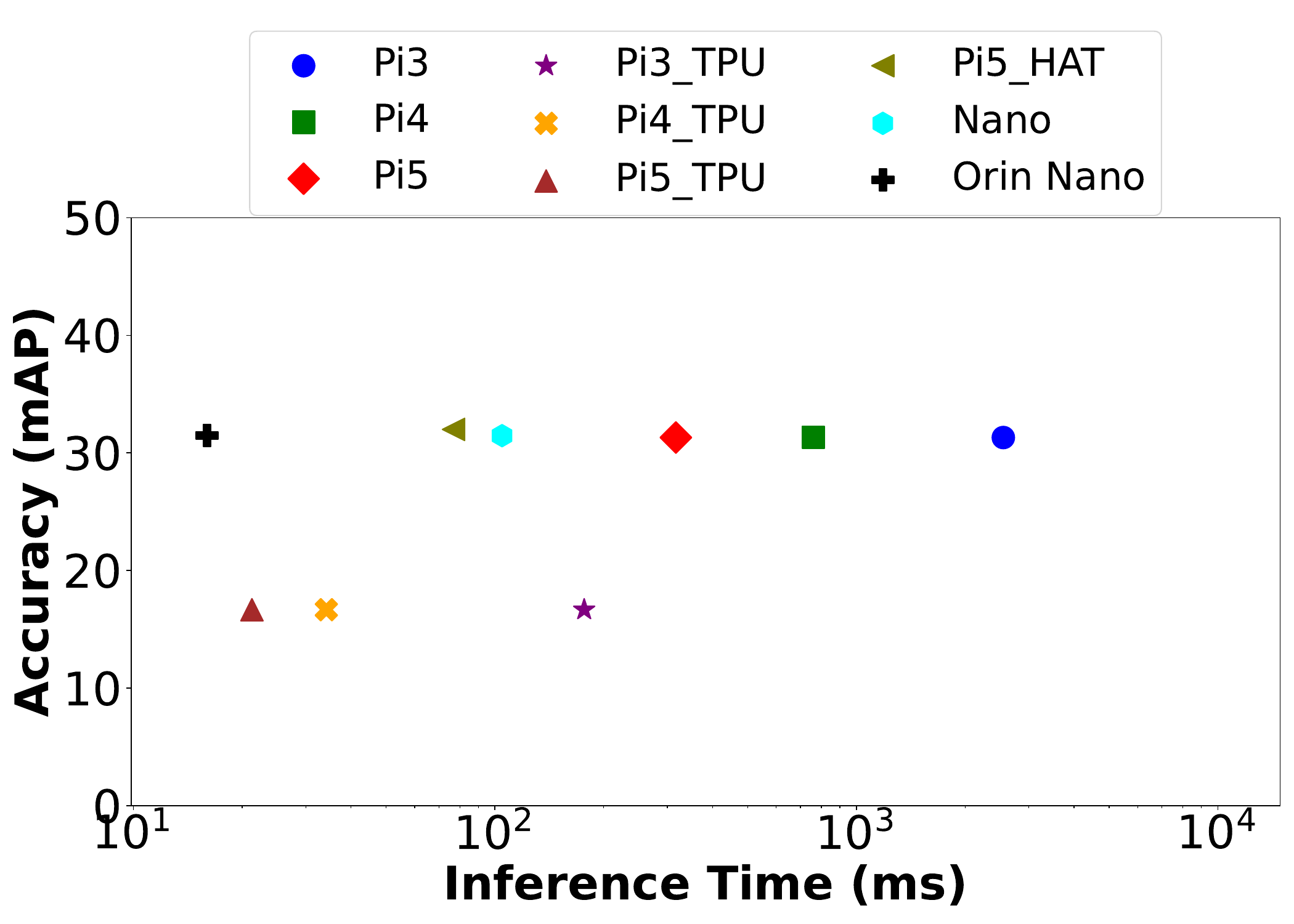}
     }
     \subfigure[Yolo8\_s]{
         \includegraphics[width=0.22\textwidth]{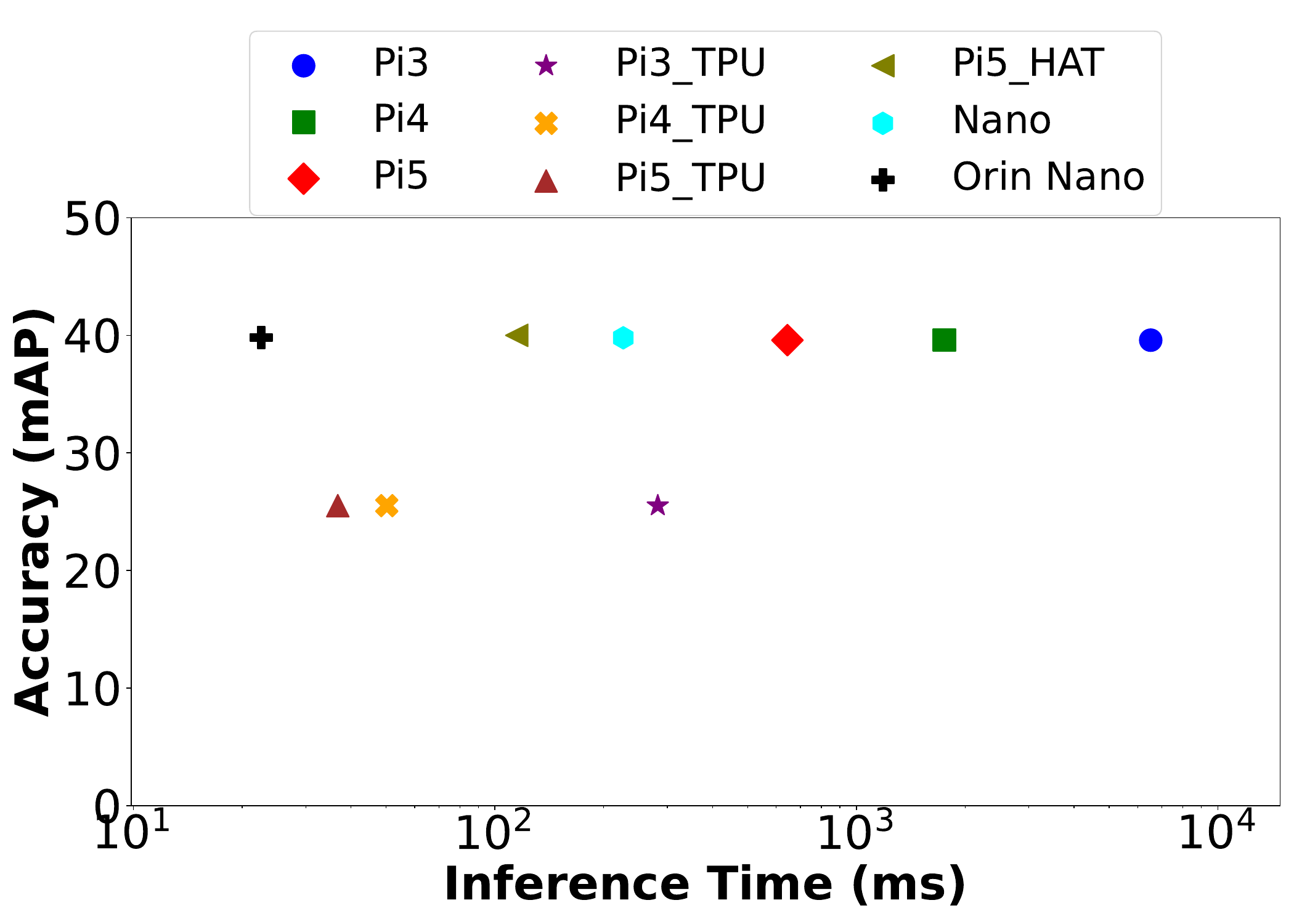}
     }
      \subfigure[Yolo8\_m]{
         \includegraphics[width=0.22\textwidth]{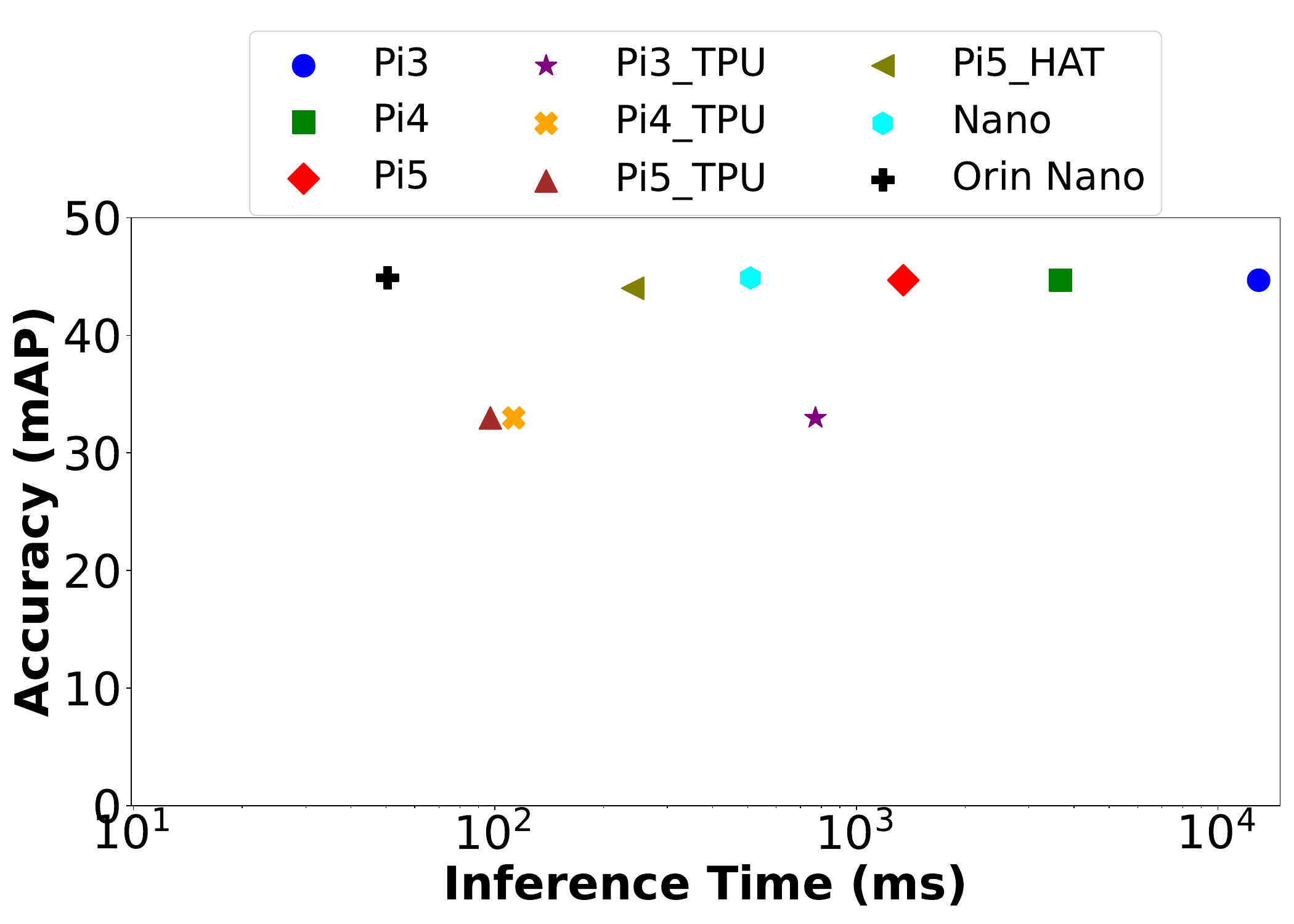}
     }
     \caption{Inference time versus accuracy for various object detection models}
     \label{fig:inference time vs accuracy}
\end{figure}

This subsection examines the trade-off between inference time and accuracy for the evaluated object detection models across edge devices, which is important for achieving low latency without substantially reducing detection quality.

For the SSD models, as shown in Fig.~\ref{fig:inference time vs accuracy}(a) and Fig.~\ref{fig:inference time vs accuracy}(b), accuracy remains largely stable across the evaluated devices, while inference time varies considerably. Raspberry Pi 5 with TPU achieves one of the lowest inference times among the Raspberry Pi platforms, whereas Raspberry Pi 3 exhibits the highest latency. Jetson Nano and Jetson Orin Nano also occupy favorable positions in the trade-off space because they combine low inference time with competitive accuracy. For the EfficientDet Lite models, shown in Fig.~\ref{fig:inference time vs accuracy}(c)--(e), a similar pattern is observed. The mAP values remain relatively stable across most devices, while the main difference lies in latency. In these plots, Jetson Orin Nano achieves the lowest inference times, while several CPU-only Raspberry Pi platforms occupy less favorable latency regions.

For the YOLOv8 models, as shown in Fig.~\ref{fig:inference time vs accuracy}(f)--(h), accuracy remains strong across most devices, but the TPU-based Raspberry Pi configurations exhibit a noticeable reduction in mAP because of the reduced input size used in that deployment setting. In contrast, Raspberry Pi 5 with AI HAT+ preserves competitive YOLOv8 accuracy while substantially reducing inference time compared with CPU-only Raspberry Pi 5. Jetson Orin Nano provides the most favorable latency-accuracy trade-off for the YOLOv8 family, combining low inference time with high mAP values, while Raspberry Pi 5 with AI HAT+ also offers a highly competitive compromise. Overall, the CPU-only Raspberry Pi platforms tend to be less favorable because they often incur high latency without a corresponding accuracy advantage.

\begin{keyinsight}
Across model families, the results show a broad trade-off between accuracy and inference time: lighter models generally achieve lower latency with lower accuracy, whereas heavier models provide higher accuracy at the cost of increased inference time. Across devices, however, this trade-off is less strict, as some accelerator-enabled platforms preserve similar accuracy for the same model while substantially reducing inference time.
\end{keyinsight}

\subsection{Energy Consumption vs Inference Time vs Accuracy} 
\begin{figure}[t]
     \centering
     \subfigure[SSD\_v1]{
         \includegraphics[width=0.22\textwidth,trim=140pt 0pt 140pt 0pt, clip]{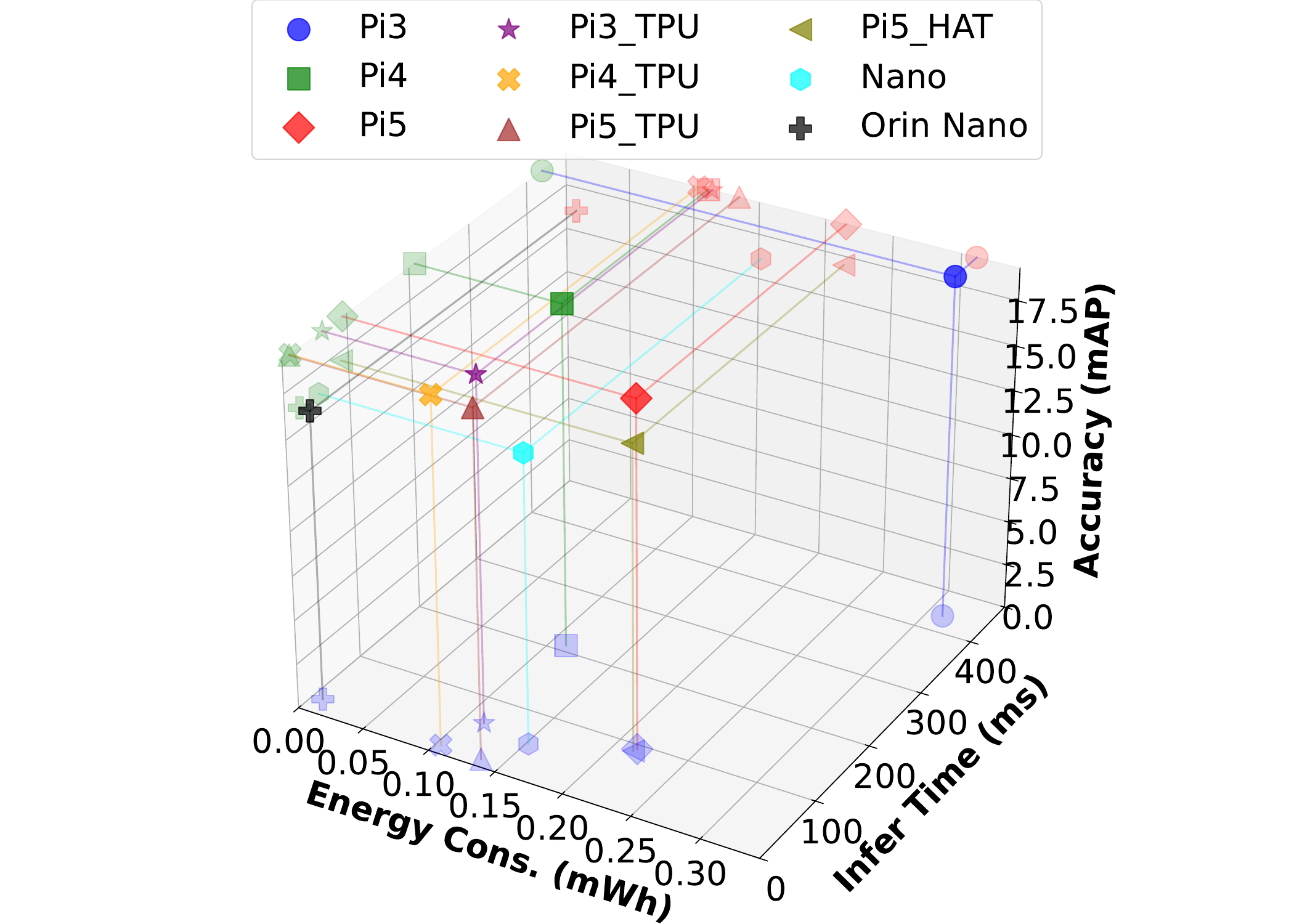}
     }
     \subfigure[SSD\_lite]{
         \includegraphics[width=0.22\textwidth,trim=140pt 0pt 140pt 0pt, clip]{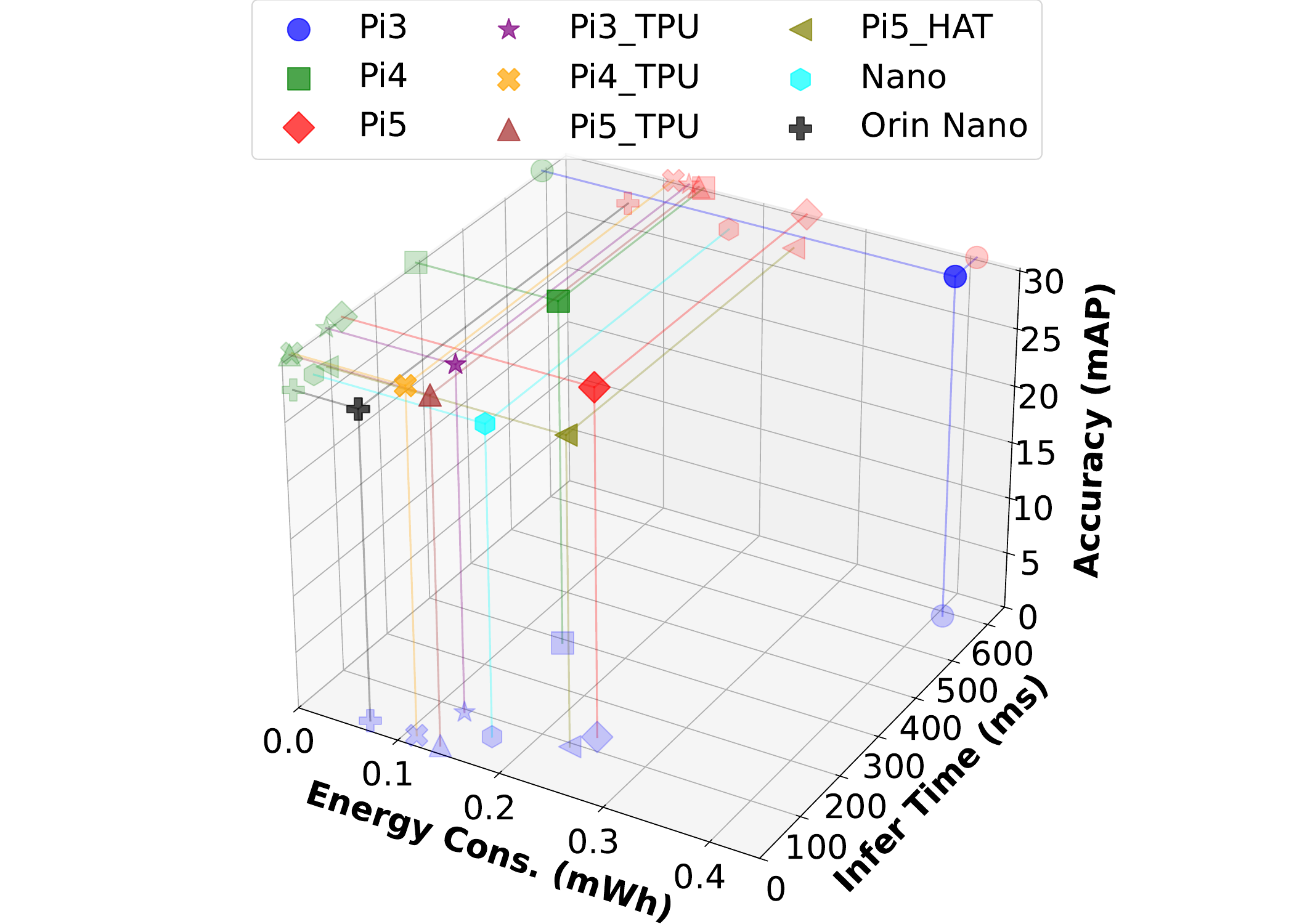}
     }
     \subfigure[Det\_lit0]{
         \includegraphics[width=0.22\textwidth,trim=140pt 0pt 140pt 0pt, clip]{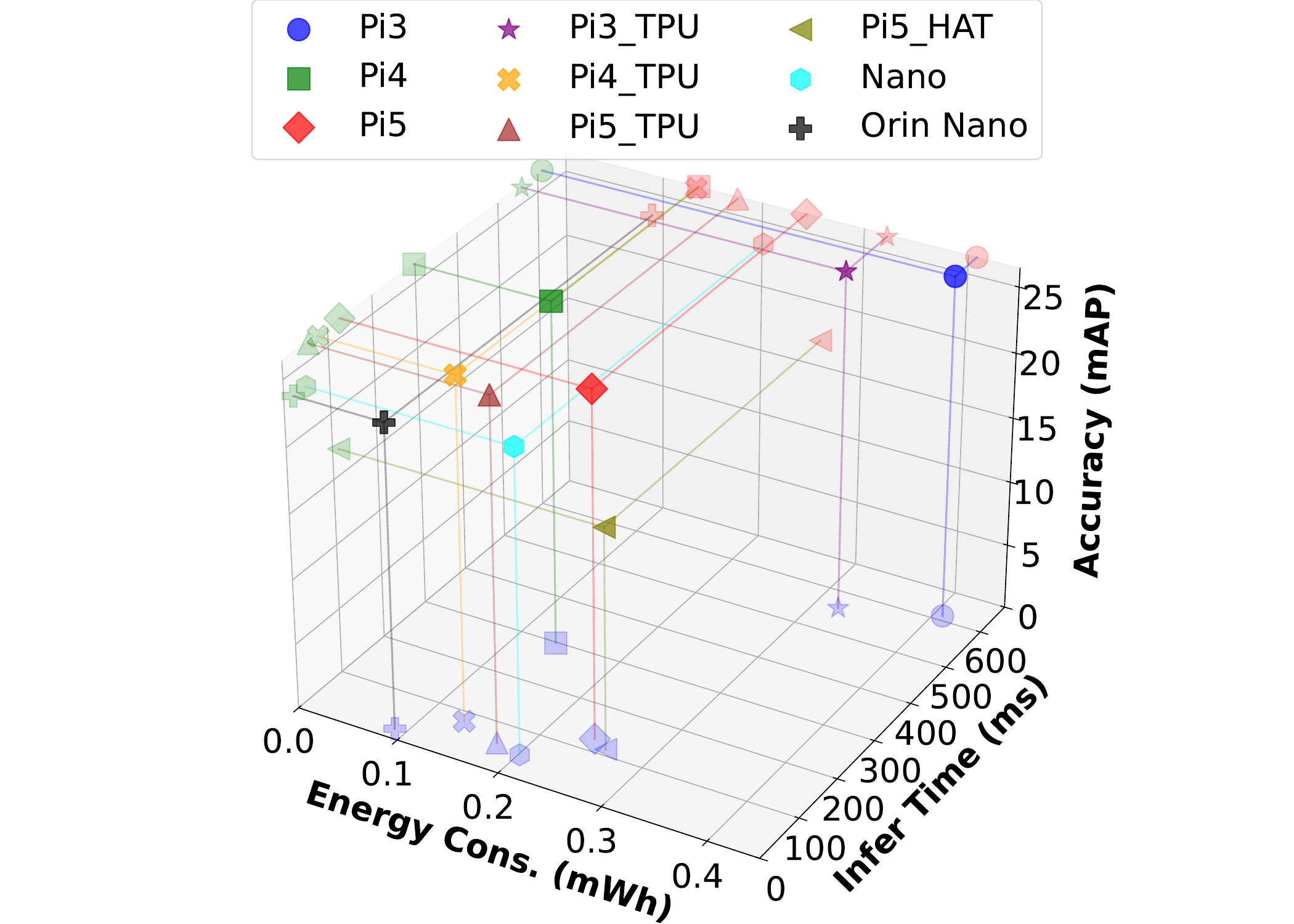}
     }
     \subfigure[Det\_lite1]{
         \includegraphics[width=0.22\textwidth,trim=140pt 0pt 140pt 0pt, clip]{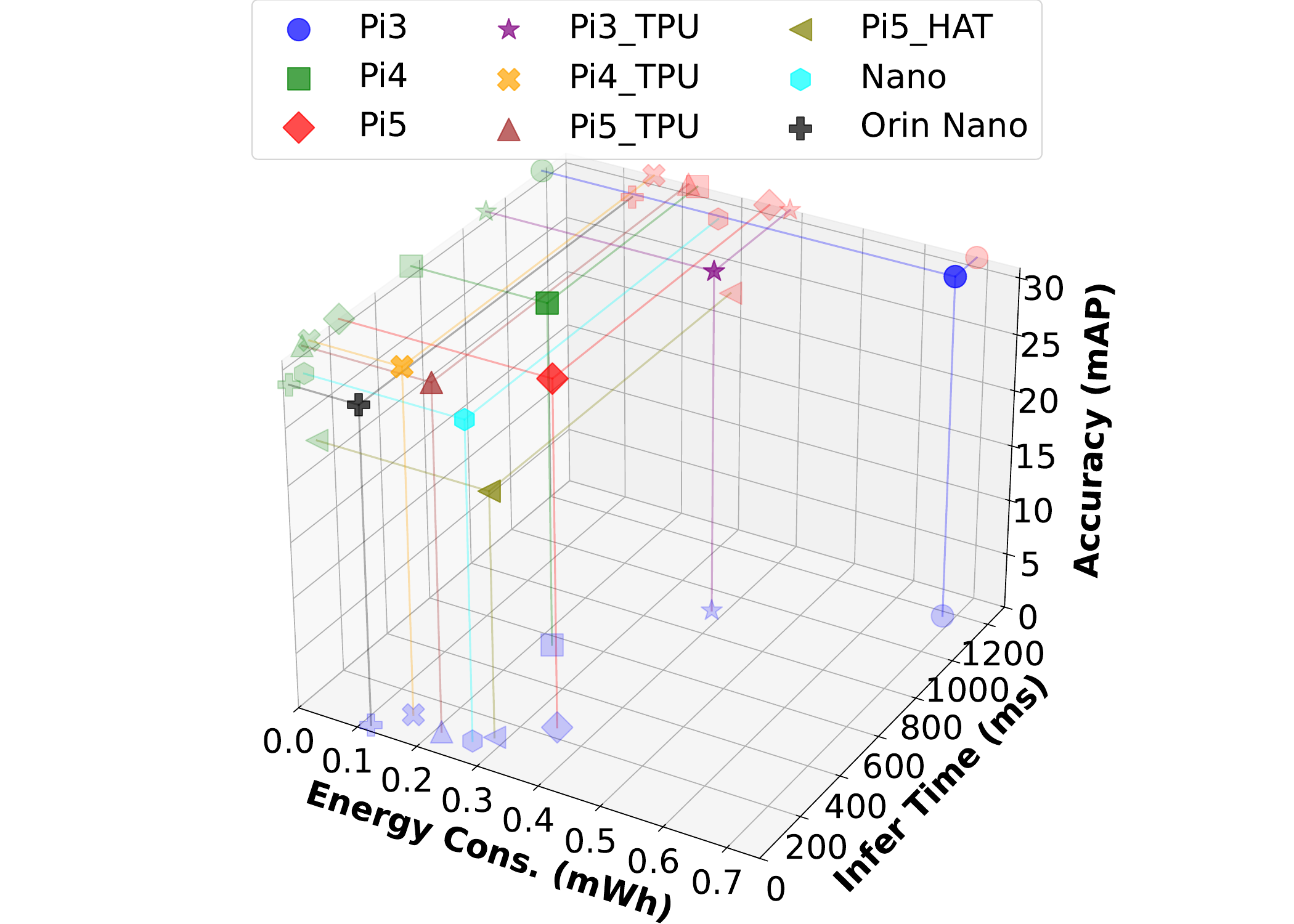}
     }
     \subfigure[Det\_lite2]{
         \includegraphics[width=0.22\textwidth,trim=140pt 0pt 140pt 0pt, clip]{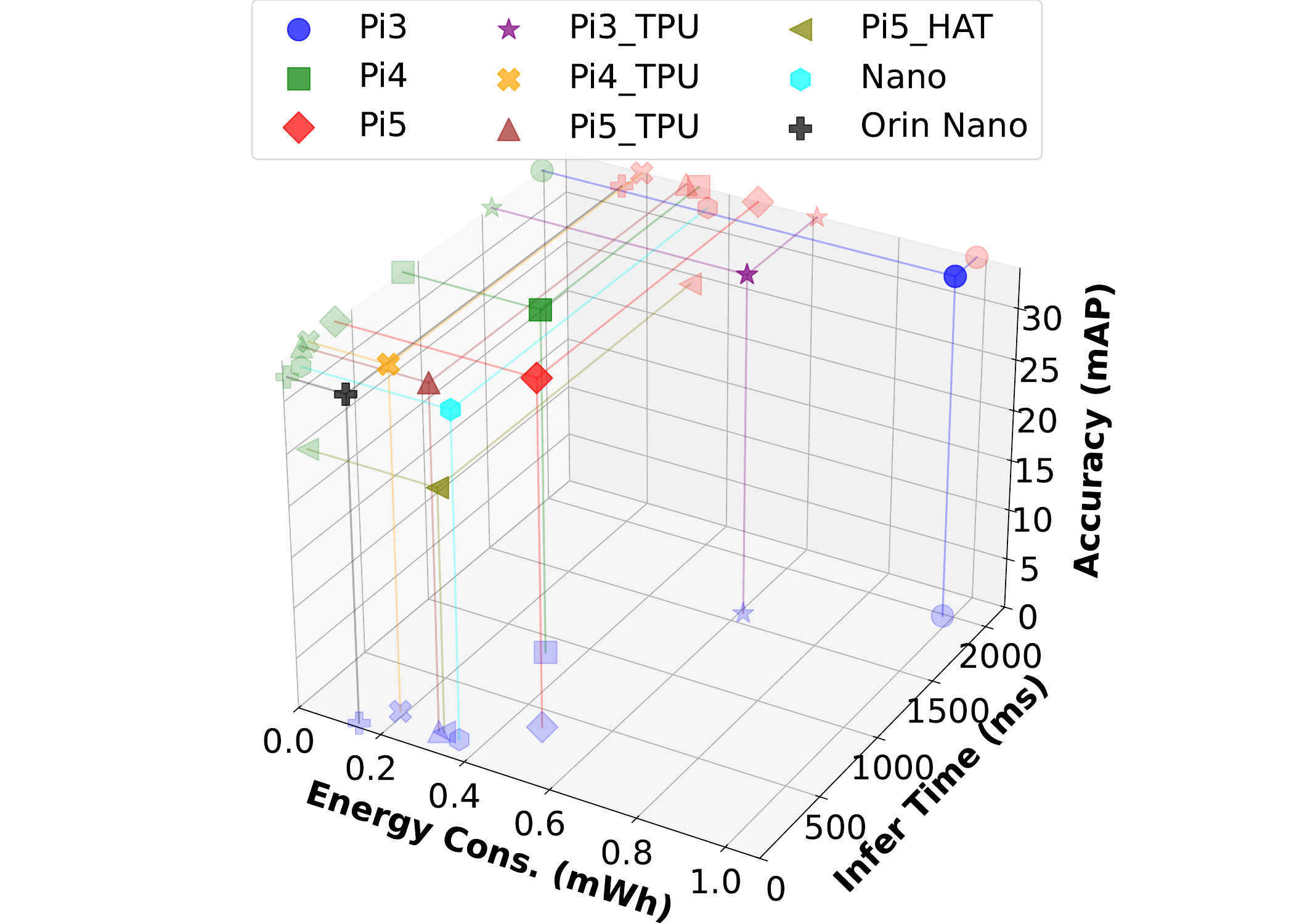}
     }
     \subfigure[Yolo8\_n]{
         \includegraphics[width=0.22\textwidth,trim=140pt 0pt 140pt 0pt, clip]{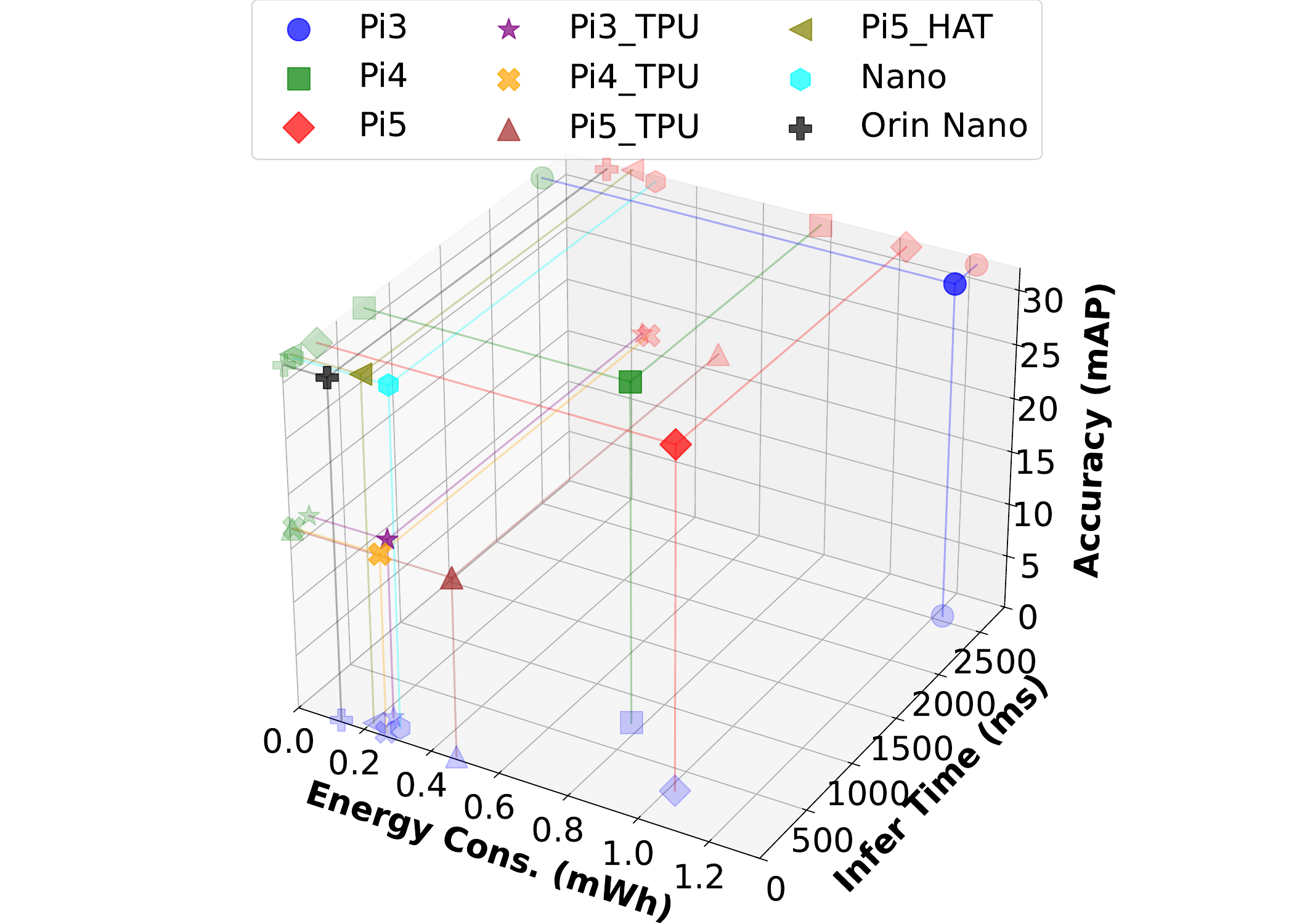}
     }
     \subfigure[Yolo8\_s]{
         \includegraphics[width=0.22\textwidth,trim=140pt 0pt 140pt 0pt, clip]{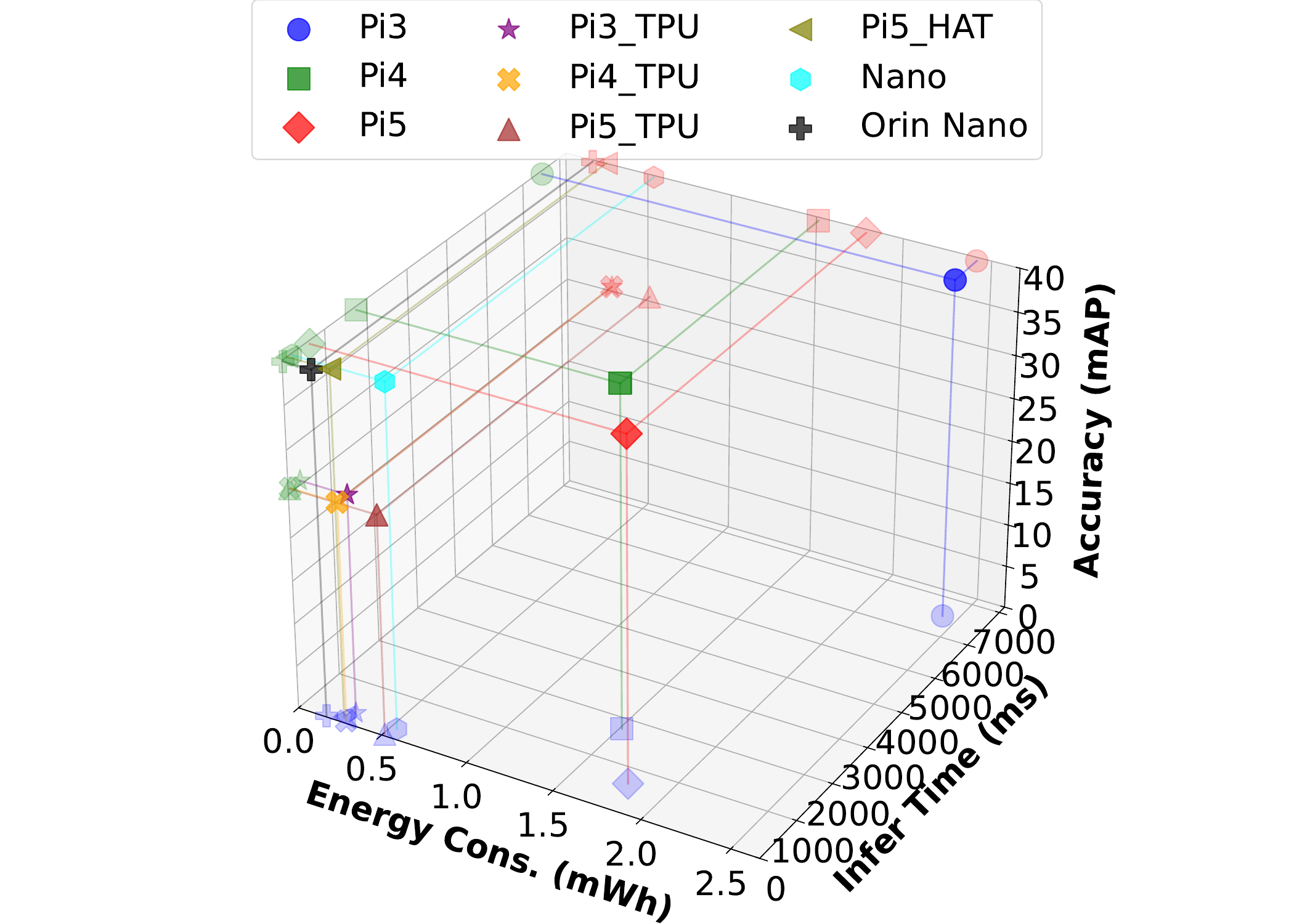}
     }
      \subfigure[Yolo8\_m]{
         \includegraphics[width=0.22\textwidth,trim=140pt 0pt 140pt 0pt, clip]{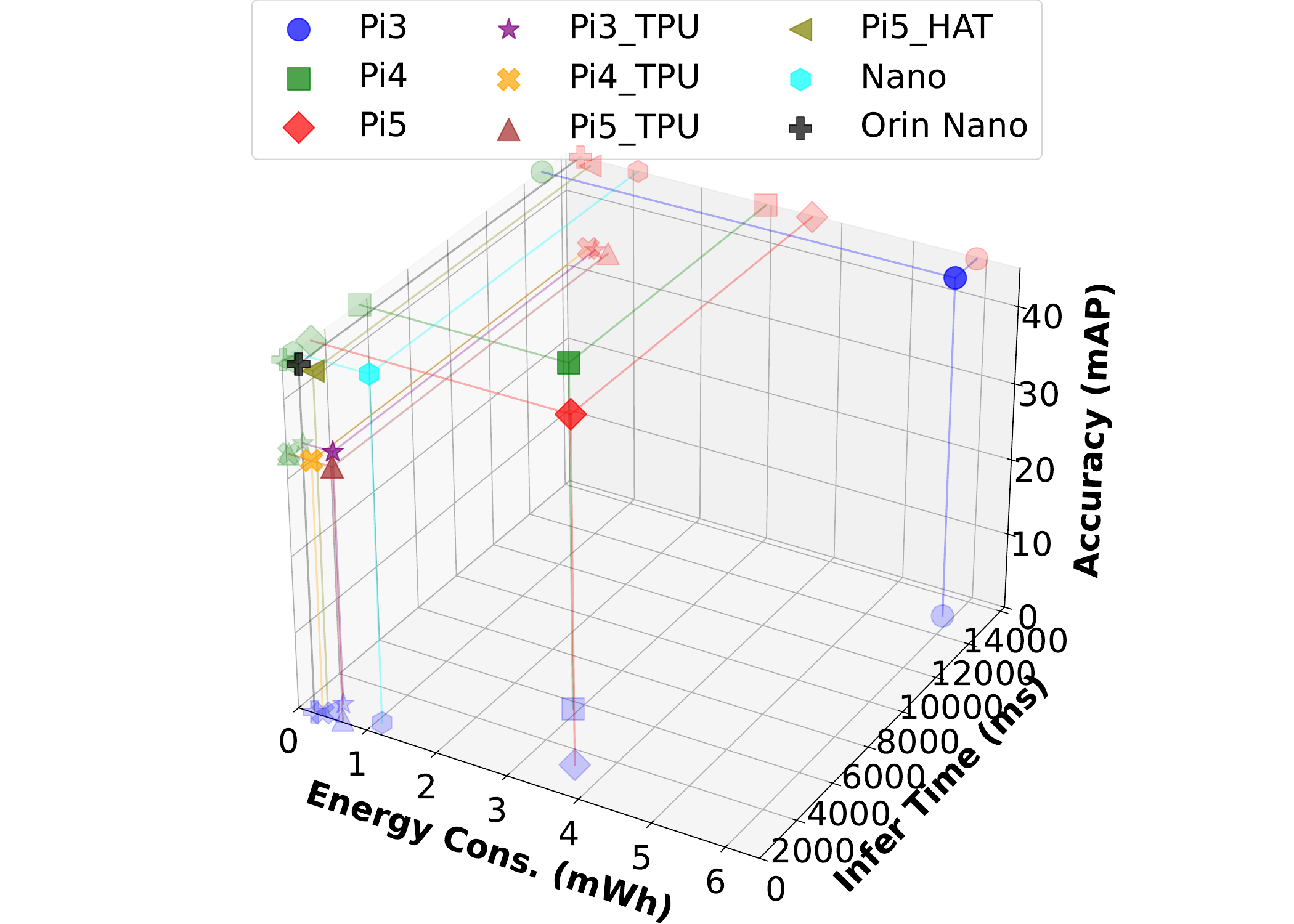}
     }
     \caption{Energy consumption per request (excluding base energy) versus inference time versus accuracy for various object detection models}
     \label{fig:energy vs inference time vs accuracy}
\end{figure}
This subsection summarizes the overall performance of the evaluated object detection models across the edge devices. Fig.~\ref{fig:energy vs inference time vs accuracy} jointly illustrates the trade-off among energy consumption, inference time, and accuracy for each model family.

For the SSD models, as shown in Fig.~\ref{fig:energy vs inference time vs accuracy}(a) and Fig.~\ref{fig:energy vs inference time vs accuracy}(b), Raspberry Pi 5 with TPU, Raspberry Pi 4 with TPU, Jetson Nano, and Orin Nano occupy the most favorable region of the trade-off space. Among these, Raspberry Pi 5 with TPU and Raspberry Pi 4 with TPU are particularly attractive because they combine very low energy consumption and inference time with the same accuracy level observed on the Raspberry Pi platforms. In contrast, Jetson Nano and Orin Nano achieve similarly strong energy and latency performance, but with a slight reduction in accuracy for SSD models.

For the EfficientDet Lite models, shown in Fig.~\ref{fig:energy vs inference time vs accuracy}(c)--(e), Jetson Nano and Orin Nano provide the most favorable overall trade-off, combining low energy consumption and low inference time with only a small decrease in mAP relative to the Raspberry Pi-based platforms. The TPU-enabled Raspberry Pi devices also provide strong latency and energy improvements, but the NVIDIA platforms remain more favorable overall in this model family. Raspberry Pi 3 consistently occupies the least favorable region because it combines higher energy consumption and longer inference time without offering an accuracy advantage.

For the YOLOv8 models, as shown in Fig.~\ref{fig:energy vs inference time vs accuracy}(f)--(h), Jetson Orin Nano and Pi 5 with AI HAT+ are the most suitable choices among the evaluated devices, providing the best overall balance among energy consumption, inference time, and accuracy. Raspberry Pi 5 with AI HAT+ also emerges as a strong option, particularly because it preserves competitive YOLOv8 accuracy while substantially improving both energy consumption and inference time relative to CPU-only Raspberry Pi 5. In contrast, TPU-based Raspberry Pi platforms achieve excellent latency and energy efficiency for YOLOv8, but their reduced accuracy makes them less attractive when all three metrics are considered jointly.

\begin{keyinsight}
The 3D analysis shows that the suitability of an object detection model depends on both the model family and the edge device type. In our experiments, Pi4 and Pi5 with TPU are most favorable for SSD, whereas Orin Nano is most favorable for EfficientDet Lite and YOLOv8.
\end{keyinsight}

\subsection{Object-Count-Based Analysis}

This subsection investigates how object count affects the performance of the evaluated object detection models. In particular, it examines whether the number of objects in an image influences accuracy, inference time, and energy consumption. 

In our experiments, meaningful variation is observed mainly in accuracy, whereas inference time and energy consumption remain largely unchanged. Therefore, the following discussion focuses on accuracy. To this end, the evaluation images are divided into five object-count groups: images containing 0, 1, 2, 3, and 4 or more objects. Images with 0 annotated objects form a special case, for which mAP is not particularly informative and is reported as 0 under our evaluation pipeline. Accordingly, the following discussion focuses primarily on the relative accuracy trends for images containing 1, 2, 3, and 4 or more objects.
\begin{figure}[t]
    \centering
    \subfigure[Raspberry Pi]{
        \includegraphics[width=0.22\textwidth]{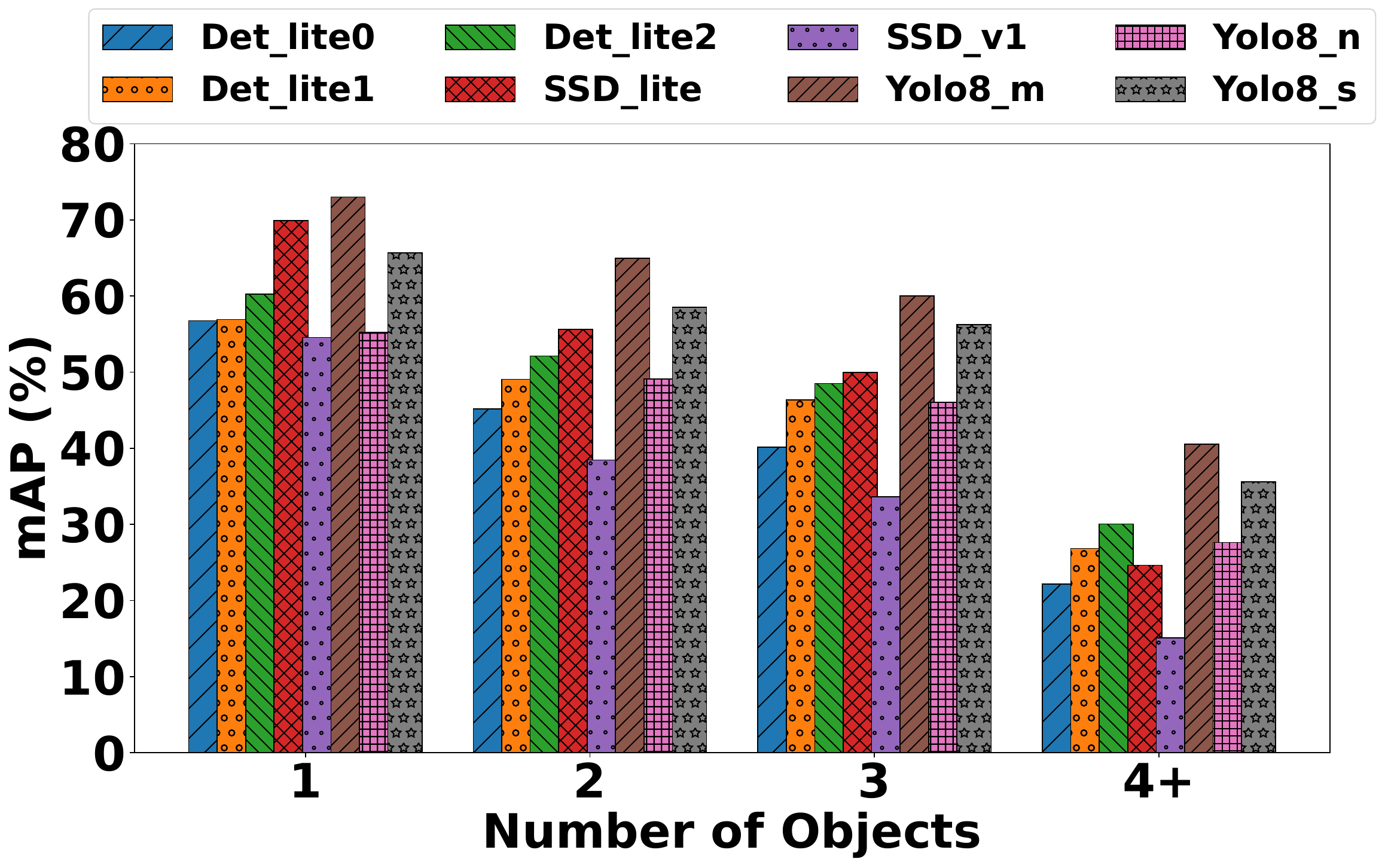}
    }
    \subfigure[Pi + TPU]{
        \includegraphics[width=0.22\textwidth]{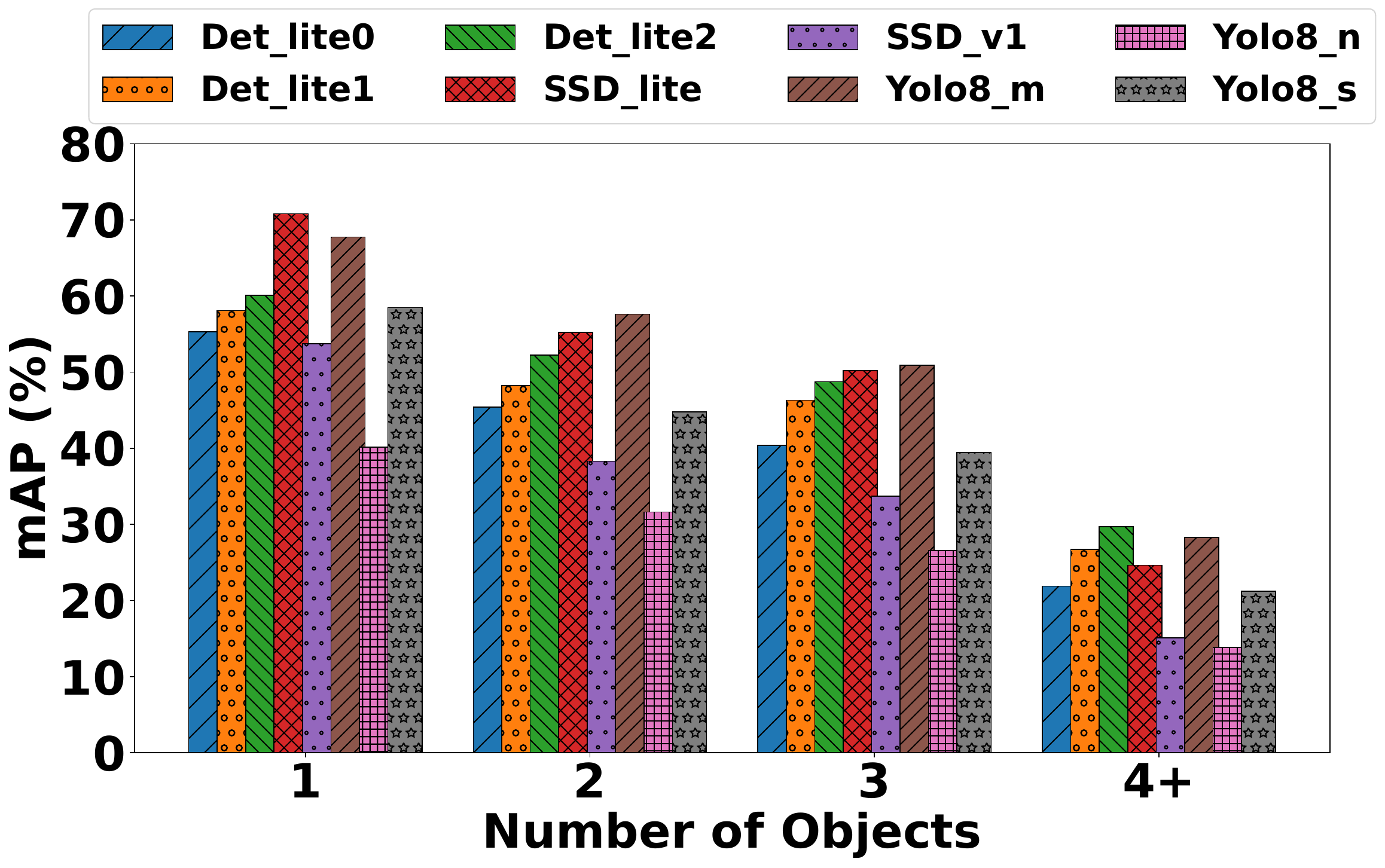}
    }
    \subfigure[Pi + AI HAT]{
        \includegraphics[width=0.22\textwidth]{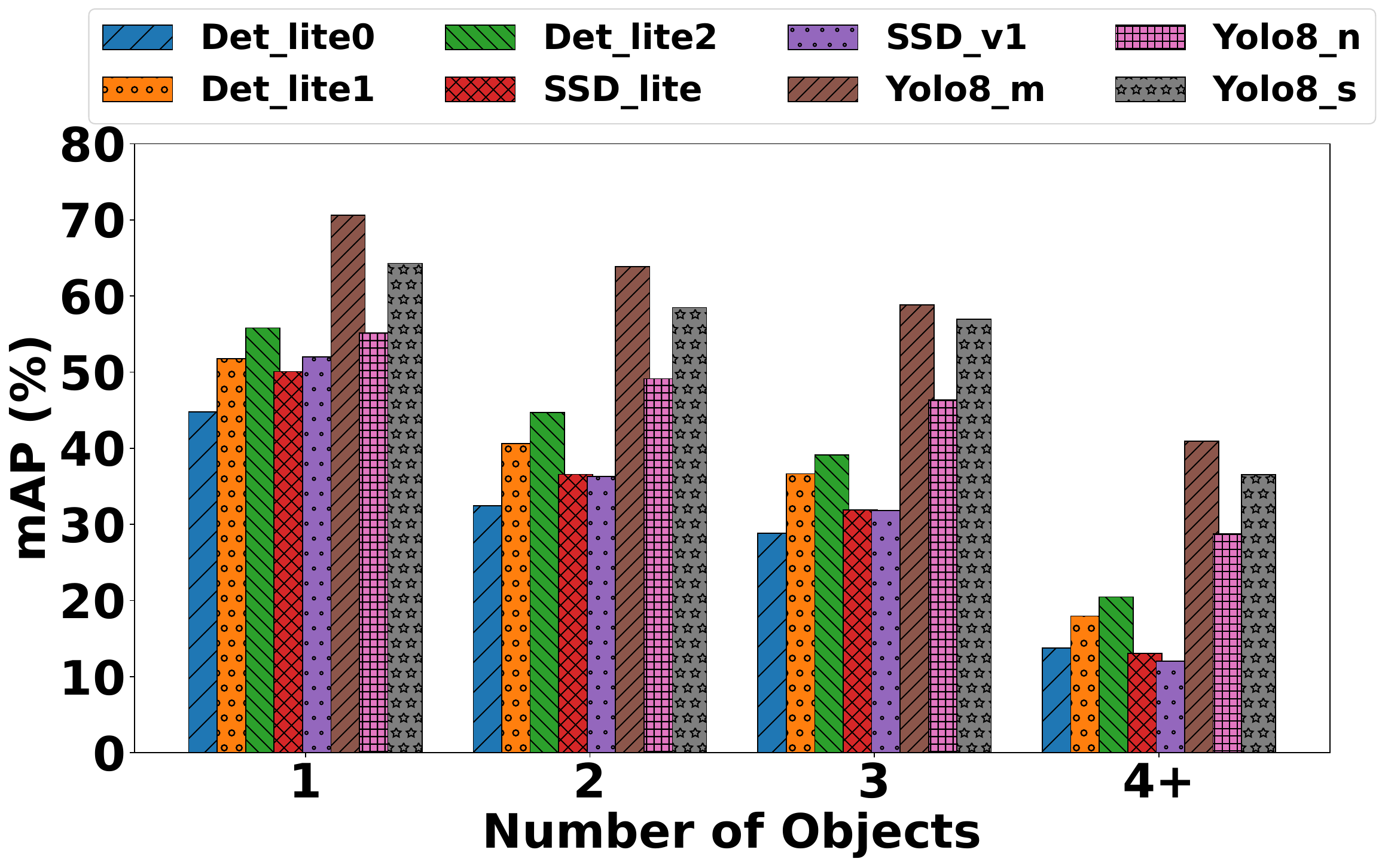}
    }
    \subfigure[NVIDIA]{
        \includegraphics[width=0.22\textwidth]{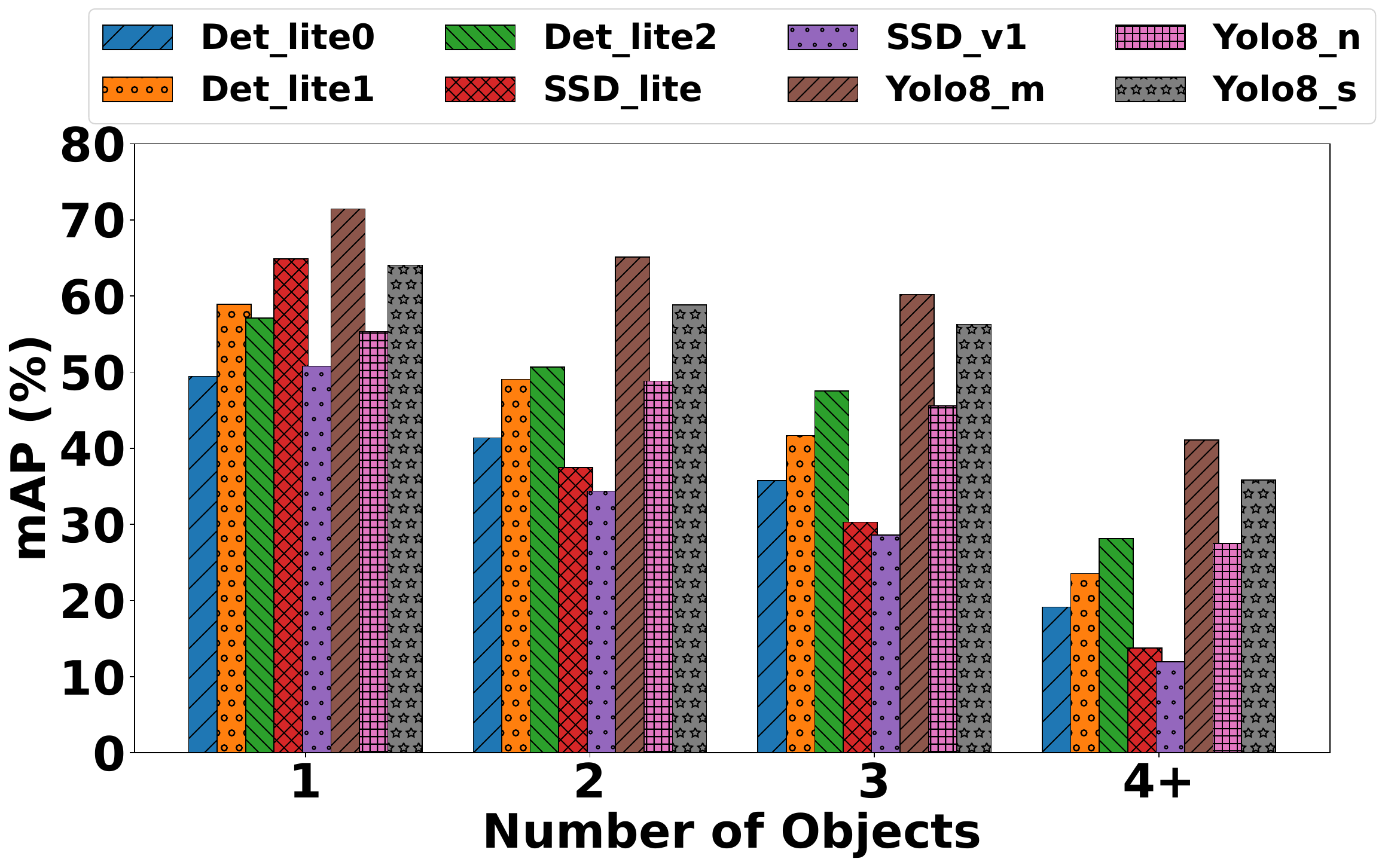}
    }

    \caption{Object-count-based mAP results for the evaluated platform categories.}
    \label{fig:mAP object count based}
\end{figure}

As shown in Fig.~\ref{fig:mAP object count based}, all platform categories exhibit the same overall trend: mAP is highest for images containing one object and decreases progressively as object count increases. Across the evaluated settings, YOLOv8\_m consistently achieves the strongest accuracy, followed by YOLOv8\_s and YOLOv8\_n, while EfficientDet Lite2 is the best performing variant within the EfficientDet family and SSD\_lite consistently outperforms SSD\_v1. Compared with the Raspberry Pi devices in Fig.~\ref{fig:mAP object count based}(a), the Raspberry Pis with TPU in Fig.~\ref{fig:mAP object count based}(b) preserve the accuracy of EfficientDet Lite and SSD relatively well, but introduce a clear accuracy reduction for YOLOv8. In Fig.~\ref{fig:mAP object count based}(c), Raspberry Pi 5 with AI HAT+ preserves YOLOv8 accuracy much more effectively than the TPU-based setting, particularly for images with higher object counts, although EfficientDet Lite and SSD show lower accuracy than on the Raspberry Pi devices. In Fig.~\ref{fig:mAP object count based}(d), the NVIDIA platforms remain highly competitive overall, preserving strong YOLOv8 performance for more complex scenes while showing slightly lower mAP for EfficientDet Lite and SSD than the Raspberry Pi devices.

To further examine the effect of scene complexity, we analyze model performance across different object counts. For images containing one or two objects, the models show relatively small performance differences, and in some cases, SSDLite achieves accuracy comparable to heavier models such as YOLOv8m. However, as the number of objects increases to four or more, SSD and SSDLite exhibit the lowest accuracy, while YOLOv8 achieves the highest among the evaluated models.

\begin{keyinsight}
Object count affects accuracy more than inference time or energy because each 
model's computation is mostly fixed for a given input size, while detection 
quality depends on separating and localizing multiple objects. In simple scenes, 
lightweight models such as SSDLite can approach YOLOv8 accuracy with lower cost, 
whereas in complex scenes the accuracy gap increases, making YOLOv8 Small or 
Medium more suitable when accuracy is prioritized. This supports scene-aware 
model selection to reduce energy without always sacrificing accuracy.
\end{keyinsight}

\section{Related Work}
\label{sec:related work}

This section reviews the most relevant studies on object detection for edge computing devices and positions our work within the existing literature. Table~\ref{comparision of related work} provides a structured comparison of prior studies across hardware platforms, model families, and evaluation criteria. Although prior work has reported valuable results on selected models and platforms, a unified evaluation across multiple detector families, heterogeneous edge devices, and complementary metrics remains limited.

Existing studies have explored edge-based object detection from different perspectives. Cantero et al.~\cite{Cantero2022}, Kamath and Renuka~\cite{Kamath2021}, Chen et al.~\cite{Chen2018}, and Puchtler and Peinl~\cite{Puchtler2020} investigate the deployment of SSD, EfficientDet, or SSD-MobileNet variants on embedded devices, focusing mainly on inference speed, quantization, accuracy, or accelerator-level benchmarking within a limited set of models and hardware platforms. Kang and Somtham~\cite{Kang2022}, Feng et al.~\cite{Feng2022}, Santos et al.~\cite{Santos2024}, Baller et al.~\cite{Baller2021}, Bulut et al.~\cite{Bulut2023}, and Appavu~\cite{Appavu2025} benchmark edge devices using object detection models and report metrics such as latency, FPS, accuracy, memory usage, hardware utilization, or energy consumption, but their analyses are typically restricted to YOLO-based models or a narrow set of devices. 

More recent work broadens the evaluated models or hardware, but important gaps remain. Zagitov et al.~\cite{Zagitov2024}, Galliera and Suri~\cite{Galliera2022}, Magalhães et al.~\cite{magalhaes2023}, and Lema et al.~\cite{Lema2024} compare a wider range of detectors or accelerators, yet they do not cover the same diverse range of edge devices evaluated in our study. Shakya et al.~\cite{Shakya2025}, Meimetis et al.~\cite{Meimetis2025}, and Such\'y and Tur\v{c}an\'ik~\cite{Suchy2026} provide recent deployment-oriented analyses of lightweight YOLO models, UAV-oriented edge detectors, and large YOLOv8/RT-DETR models, respectively. In addition, Mittal~\cite{Mittal2024} and Zhai et al.~\cite{zhai2024} survey lightweight object detection and edge deployment for visual detection, covering detector architectures, optimization methods, deployment frameworks, and edge hardware platforms. However, these works either focus on specific detector families or provide surveys rather than a unified experimental comparison. 

Overall, existing studies do not provide a unified comparison across multiple detector families, heterogeneous edge devices, and complementary performance dimensions. Most focus on a narrow set of models or platforms, often emphasize latency or accuracy alone, and only inconsistently consider energy consumption. In addition, the effect of scene complexity on model behavior remains largely underexplored in edge benchmarking. In contrast, our study jointly evaluates SSD, EfficientDet Lite, and YOLOv8 across CPU-, TPU-, NPU-, and GPU-based edge devices, while systematically analyzing inference time, energy consumption, accuracy, and object-count-based behavior in a single experimental framework.

\begin{table*}[t]
\caption{Comparison of Existing Studies Across Edge Devices, Models, and Evaluation Criteria.}
\label{comparision of related work}
\scriptsize
\centering
\begin{tabular}{|l|c|c|c|c|c|c|c|c|c|c|c|}
\hline
\textbf{Study} & \textbf{CPU} & \textbf{TPU} & \textbf{GPU} & \textbf{NPU} & \textbf{YOLOv8} & \textbf{EfficientDet} & \textbf{SSD} & \textbf{Infer Time} & \textbf{Energy} & \textbf{mAP} & \textbf{Object-Count-Based} \\
\hline
Cantero et al.~\cite{Cantero2022}  & \checkmark & \checkmark &  &  &  & \checkmark & \checkmark & \checkmark &  &  &  \\
Kamath and Renuka~\cite{Kamath2021}  & \checkmark &  &  &  &  & \checkmark &  & \checkmark &  & \checkmark &  \\
Kang and Somtham~\cite{Kang2022}  &  & \checkmark & \checkmark &  &  &  & \checkmark & \checkmark & \checkmark & \checkmark &  \\
Baller et al.~\cite{Baller2021}  & \checkmark & \checkmark & \checkmark &  &  &  & \checkmark & \checkmark & \checkmark & \checkmark &  \\
Bulut et al.~\cite{Bulut2023}  &  &  & \checkmark &  &  &  &  & \checkmark & \checkmark & \checkmark &  \\
Chen et al.~\cite{Chen2018}  & \checkmark &  &  &  &  &  & \checkmark & \checkmark &  & \checkmark &  \\
Zagitov et al.~\cite{Zagitov2024}  & \checkmark &  & \checkmark &  & \checkmark & \checkmark & \checkmark & \checkmark &  & \checkmark &  \\
Galliera and Suri~\cite{Galliera2022}  &  & \checkmark & \checkmark &  &  &  &  & \checkmark &  & \checkmark &  \\
Magalhães et al.~\cite{magalhaes2023}  &  & \checkmark & \checkmark &  &  &  &  & \checkmark & \checkmark & \checkmark &  \\
Lema et al.~\cite{Lema2024} &  & \checkmark & \checkmark &  &  &  &  & \checkmark & \checkmark & \checkmark &  \\
Feng et al.~\cite{Feng2022} & \checkmark &  & \checkmark &  &  &  &  & \checkmark & \checkmark & \checkmark &  \\
Appavu~\cite{Appavu2025} & \checkmark & \checkmark & \checkmark &  &  &  & \checkmark & \checkmark & \checkmark & \checkmark &  \\
Santos et al.~\cite{Santos2024} & \checkmark &  & \checkmark &  &  &  &  & \checkmark &  &  &  \\
Meimetis et al.~\cite{Meimetis2025} &  &  & \checkmark &  &  &  &  & \checkmark & \checkmark & \checkmark &  \\
Puchtler and Peinl~\cite{Puchtler2020} & \checkmark & \checkmark & \checkmark &  &  &  & \checkmark & \checkmark & \checkmark & \checkmark &  \\
Suchý and Turčaník~\cite{Suchy2026} & \checkmark &  & \checkmark & \checkmark & \checkmark &  &  & \checkmark & \checkmark & \checkmark &  \\
Shakya et al.~\cite{Shakya2025} &  &  & \checkmark &  & \checkmark &  &  & \checkmark & \checkmark & \checkmark &  \\
Mittal~\cite{Mittal2024} &  &  &  &  &  &  &  &  &  &  &  \\
Zhai et al.~\cite{zhai2024} &  &  &  &  &  &  &  &  &  &  &  \\
\textbf{Our Work} & \checkmark & \checkmark & \checkmark & \checkmark & \checkmark & \checkmark & \checkmark & \checkmark & \checkmark & \checkmark & \checkmark \\
\hline
\end{tabular}
\end{table*}

\section{Conclusions and Future Direction}
\label{sec:conclusions}

In this paper, we benchmarked YOLOv8, EfficientDet Lite, and SSD variants across heterogeneous edge devices, including Raspberry Pi platforms with and without accelerators, Raspberry Pi 5 with AI HAT+, and NVIDIA Jetson devices. The evaluation considered mAP, inference time, energy consumption, and the effect of scene complexity based on object count.

The results reveal non-trivial model-device trade-offs. SSD MobileNet V1 achieves low inference time and energy consumption but lower accuracy, whereas YOLOv8 Medium provides the highest accuracy at greater computational cost. Accelerator effects are model-dependent: TPU-enabled Raspberry Pi devices improve SSD and EfficientDet Lite efficiency but reduce YOLOv8 accuracy due to deployment constraints, while Jetson Orin Nano offers the best overall balance across most model families. Object-count analysis further shows that lightweight and heavyweight models perform similarly on simple scenes, but the accuracy gap widens as scene complexity increases, highlighting the need to consider input characteristics in edge deployment decisions.

Future work will further examine the effects of quantization and reduced precision, such as FP16 and INT8, and extend scene analysis beyond object count to include object classes and other visual characteristics.

\bibliographystyle{IEEEtran}
\bibliography{main}

@Article{Balasubramaniam2022,
  author  = {Balasubramaniam, Abhishek and Pasricha, Sudeep},
  journal = {arXiv preprint arXiv:2201.07706},
  title   = {Object detection in autonomous vehicles: Status and open challenges},
  year    = {2022},
}

@InProceedings{Redmon2016,
  author    = {Redmon, Joseph and Divvala, Santosh and Girshick, Ross and Farhadi, Ali},
  booktitle = {Proceedings of the IEEE conference on computer vision and pattern recognition},
  title     = {You only look once: Unified, real-time object detection},
  year      = {2016},
  pages     = {779--788},
}

@InProceedings{Liu2016,
  author       = {Liu, Wei and Anguelov, Dragomir and Erhan, Dumitru and Szegedy, Christian and Reed, Scott and Fu, Cheng-Yang and Berg, Alexander C},
  booktitle    = {Computer Vision--ECCV 2016: 14th European Conference, Amsterdam, The Netherlands, October 11--14, 2016, Proceedings, Part I 14},
  title        = {Ssd: Single shot multibox detector},
  year         = {2016},
  organization = {Springer},
  pages        = {21--37},
}

@InProceedings{Tan2020,
  author    = {Tan, Mingxing and Pang, Ruoming and Le, Quoc V},
  booktitle = {Proceedings of the IEEE/CVF conference on computer vision and pattern recognition},
  title     = {Efficientdet: Scalable and efficient object detection},
  year      = {2020},
  pages     = {10781--10790},
}

@Article{Cantero2022,
  author    = {Cantero, David and Esnaola-Gonzalez, Iker and Miguel-Alonso, Jose and Jauregi, Ekaitz},
  journal   = {Sensors},
  title     = {Benchmarking object detection deep learning models in embedded devices},
  year      = {2022},
  number    = {11},
  pages     = {4205},
  volume    = {22},
  publisher = {MDPI},
}

@InProceedings{Kamath2021,
  author       = {Kamath, Vidya and Renuka, A},
  booktitle    = {2021 International Conference on Circuits, Controls and Communications (CCUBE)},
  title        = {Performance analysis of the pretrained efficientdet for real-time object detection on raspberry pi},
  year         = {2021},
  organization = {IEEE},
  pages        = {1--6},
}

@Article{Kang2022,
  author    = {Kang, Pilsung and Somtham, Athip},
  journal   = {Mathematics},
  title     = {An evaluation of modern accelerator-based edge devices for object detection applications},
  year      = {2022},
  number    = {22},
  pages     = {4299},
  volume    = {10},
  publisher = {MDPI},
}

@InProceedings{Baller2021,
  author       = {Baller, Stephan Patrick and Jindal, Anshul and Chadha, Mohak and Gerndt, Michael},
  booktitle    = {2021 IEEE International Conference on Cloud Engineering (IC2E)},
  title        = {DeepEdgeBench: Benchmarking deep neural networks on edge devices},
  year         = {2021},
  organization = {IEEE},
  pages        = {20--30},
}

@InProceedings{Bulut2023,
  author    = {Bulut, Anilcan and Ozdemir, Fatmanur and Bostanci, Yavuz Selim and Soyturk, Mujdat},
  booktitle = {Proceedings of the 2023 5th International Conference on Image Processing and Machine Vision},
  title     = {Performance Evaluation of Recent Object Detection Models for Traffic Safety Applications on Edge},
  year      = {2023},
  pages     = {1--6},
}

@InProceedings{Chen2018,
  author    = {Chen, Chuan-Wen and Ruan, Shanq-Jang and Lin, Chang-Hong and Hung, Chun-Chi},
  booktitle = {Proceedings of the 2018 VII International Conference on Network, Communication and Computing},
  title     = {Performance evaluation of edge computing-based deep learning object detection},
  year      = {2018},
  pages     = {40--43},
}

@Article{Zagitov2024,
  author  = {Zagitov, A and Chebotareva, E and Toschev, A and Magid, E},
  journal = {Computer},
  title   = {Comparative analysis of neural network models performance on low-power devices for a real-time object detection task},
  year    = {2024},
  number  = {2},
  volume  = {48},
}

@Article{Galliera2022,
  author    = {Galliera, Raffaele and Suri, Niranjan},
  journal   = {Procedia Computer Science},
  title     = {Object Detection at the Edge: Off-the-shelf Deep Learning Capable Devices and Accelerators},
  year      = {2022},
  pages     = {239--248},
  volume    = {205},
  publisher = {Elsevier},
}

@Article{Lema2024,
  author    = {Lema, Dar{\'\i}o G and Usamentiaga, Rub{\'e}n and Garc{\'\i}a, Daniel F},
  journal   = {Integration},
  title     = {Quantitative comparison and performance evaluation of deep learning-based object detection models on edge computing devices},
  year      = {2024},
  pages     = {102127},
  volume    = {95},
  publisher = {Elsevier},
}

@InProceedings{Lin2014,
  author       = {Lin, Tsung-Yi and Maire, Michael and Belongie, Serge and Hays, James and Perona, Pietro and Ramanan, Deva and Doll{\'a}r, Piotr and Zitnick, C Lawrence},
  booktitle    = {Computer Vision--ECCV 2014: 13th European Conference, Zurich, Switzerland, September 6-12, 2014, Proceedings, Part V 13},
  title        = {Microsoft coco: Common objects in context},
  year         = {2014},
  organization = {Springer},
  pages        = {740--755},
}

@article{magalhaes2023,
  title={Benchmarking edge computing devices for grape bunches and trunks detection using accelerated object detection single shot multibox deep learning models},
  author={Magalh{\~a}es, Sandro Costa and dos Santos, Filipe Neves and Machado, Pedro and Moreira, Ant{\'o}nio Paulo and Dias, Jorge},
  journal={Engineering Applications of Artificial Intelligence},
  volume={117},
  pages={105604},
  year={2023},
  publisher={Elsevier}
}

@misc{ultralytics2024,
  author = {Glenn Jocher and Ayush Chaurasia and Jing Qiu},
  title = {Ultralytics YOLOv8},
  version = {8.0.0},
  year = {2023},
  url = {https://github.com/ultralytics/ultralytics},
  orcid = {0000-0001-5950-6979, 0000-0002-7603-6750, 0000-0003-3783-7069},
  license = {AGPL-3.0}
}

@article{shakya2025,
  title={Small-Object Detection at the Edge: A Pareto-Efficient Benchmark of Lightweight YOLO Models on UAV and Overhead Datasets},
  author={Shakya, Bijay and El-Gayar, Omar and Kaabi, Jihene and Ahmed, Khandaker Mamun},
  journal={IEEE Access},
  volume={14},
  pages={528--548},
  year={2025},
  publisher={IEEE}
}

@article{suchy2026,
  title={Review of large YOLOv8 and RT-DETR energy efficiency on edge devices for real-time detection},
  author={Such{\`y}, Ivan and Tur{\v{c}}an{\'\i}k, Michal},
  journal={Scientific Reports},
  year={2026},
  publisher={Nature Publishing Group UK London}
}

@article{mittal2024,
  title={A comprehensive survey of deep learning-based lightweight object detection models for edge devices.},
  author={Mittal, Payal},
  journal={Artificial Intelligence Review},
  volume={57},
  number={9},
  year={2024}
}

@article{feng2022,
  title={Benchmark analysis of yolo performance on edge intelligence devices},
  author={Feng, Haogang and Mu, Gaoze and Zhong, Shida and Zhang, Peichang and Yuan, Tao},
  journal={Cryptography},
  volume={6},
  number={2},
  pages={16},
  year={2022},
  publisher={MDPI}
}

@inproceedings{appavu2025,
  title={Analysing the effect of edge-optimized deep learning models on improving low-powered iot devices real-time object detection},
  author={Appavu, Narenthirakumar},
  booktitle={2025 9th International Conference on Inventive Systems and Control (ICISC)},
  pages={1663--1669},
  year={2025},
  organization={IEEE}
}

@article{santos2024,
  title={Real-time object detection performance analysis using YOLOv7 on edge devices},
  author={Santos, Ricardo C Camara de M and Coelho, Mateus and Oliveira, Ricardo},
  journal={IEEE Latin America Transactions},
  volume={22},
  number={10},
  pages={799--805},
  year={2024},
  publisher={IEEE}
}

@article{meimetis2025,
  title={Comparative Analysis of Object Detection Models for Edge Devices in UAV Swarms},
  author={Meimetis, Dimitrios and Daramouskas, Ioannis and Patrinopoulou, Niki and Lappas, Vaios and Kostopoulos, Vassilis},
  journal={Machines},
  volume={13},
  number={8},
  pages={684},
  year={2025},
  publisher={MDPI}
}

@inproceedings{puchtler2020,
  title={Evaluation of deep learning accelerators for object detection at the edge},
  author={Puchtler, Pascal and Peinl, Ren{\'e}},
  booktitle={German Conference on Artificial Intelligence (K{\"u}nstliche Intelligenz)},
  pages={320--326},
  year={2020},
  organization={Springer}
}

@article{zhai2024,
  title={Edge deployment of deep networks for visual detection: a review},
  author={Zhai, Haozhou and Du, Jinwei and Ai, Yuhui and Hu, Tianjiang},
  journal={IEEE Sensors Journal},
  volume={25},
  number={11},
  pages={18662--18683},
  year={2024},
  publisher={IEEE}
}

@inproceedings{alqahtani2026ecore,
  title={Ecore: Energy-conscious optimized routing for deep learning models at the edge},
  author={Alqahtani, Daghash K and Rodriguez, Maria A and Cheema, Muhammad Aamir and Rezatofighi, Hamid and Toosi, Adel N},
  booktitle={2026 IEEE International Conference on Pervasive Computing and Communications (PerCom)},
  pages={1--10},
  year={2026},
  organization={IEEE}
}

@article{alqahtani2026multi,
  title={Multi-Objective Load Balancing for Heterogeneous Edge-Based Object Detection Systems},
  author={Alqahtani, Daghash K and Rodriguez, Maria A and Cheema, Muhammad Aamir and Toosi, Adel N},
  journal={arXiv preprint arXiv:2603.15400},
  year={2026}
}

@inproceedings{xiong2021mobiledets,
  title={Mobiledets: Searching for object detection architectures for mobile accelerators},
  author={Xiong, Yunyang and Liu, Hanxiao and Gupta, Suyog and Akin, Berkin and Bender, Gabriel and Wang, Yongzhe and Kindermans, Pieter-Jan and Tan, Mingxing and Singh, Vikas and Chen, Bo},
  booktitle={Proceedings of the IEEE/CVF conference on computer vision and pattern recognition},
  pages={3825--3834},
  year={2021}
}

@inproceedings{sandler2018mobilenetv2,
  title={Mobilenetv2: Inverted residuals and linear bottlenecks},
  author={Sandler, Mark and Howard, Andrew and Zhu, Menglong and Zhmoginov, Andrey and Chen, Liang-Chieh},
  booktitle={Proceedings of the IEEE conference on computer vision and pattern recognition},
  pages={4510--4520},
  year={2018}
}

@article{howard2017mobilenets,
  title={Mobilenets: Efficient convolutional neural networks for mobile vision applications},
  author={Howard, Andrew G and Zhu, Menglong and Chen, Bo and Kalenichenko, Dmitry and Wang, Weijun and Weyand, Tobias and Andreetto, Marco and Adam, Hartwig},
  journal={arXiv preprint arXiv:1704.04861},
  year={2017}
}

\begin{IEEEbiography}
[{\includegraphics[width=1in,height=1.25in,clip,keepaspectratio]
{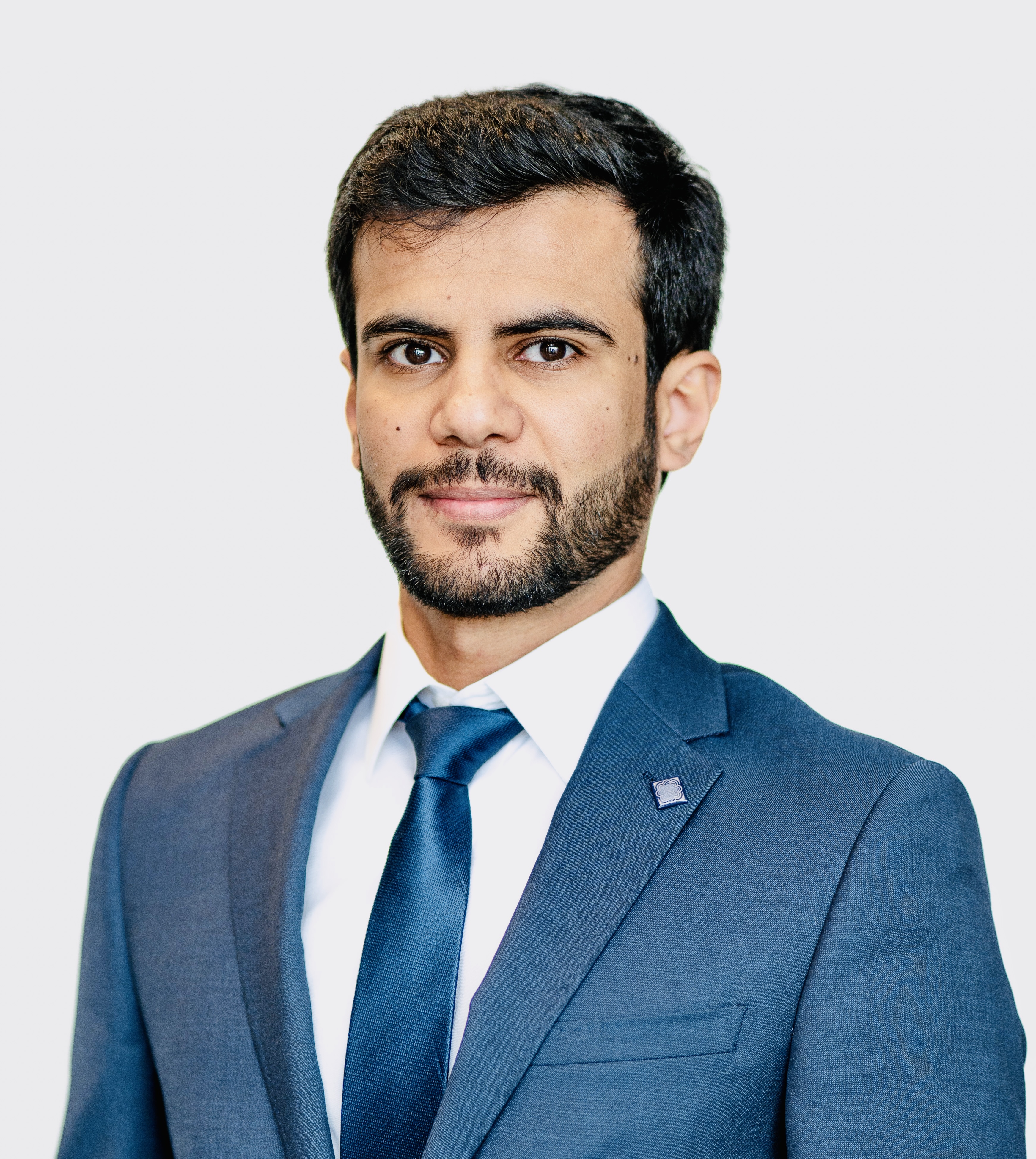}}]
{Daghash K. Alqahtani}
received the B.Sc. degree in Computer Science from King Khalid University,
Bisha, Saudi Arabia, and the Master of Information Technology degree from
Monash University, Melbourne, VIC, Australia, in 2022. He is currently
pursuing the Ph.D. degree with the School of Computing and Information
Systems, The University of Melbourne, Melbourne, VIC, Australia. His research
interests include edge computing, energy-efficient artificial intelligence,
heterogeneous computing systems, and resource management for edge intelligence.
He has published research papers in leading international conferences, including
IEEE PerCom, ICSOC, CCGrid, and IEEE EDGE.
\end{IEEEbiography}

\begin{IEEEbiography}
[{\includegraphics[width=1in,height=1.25in,clip,keepaspectratio]
{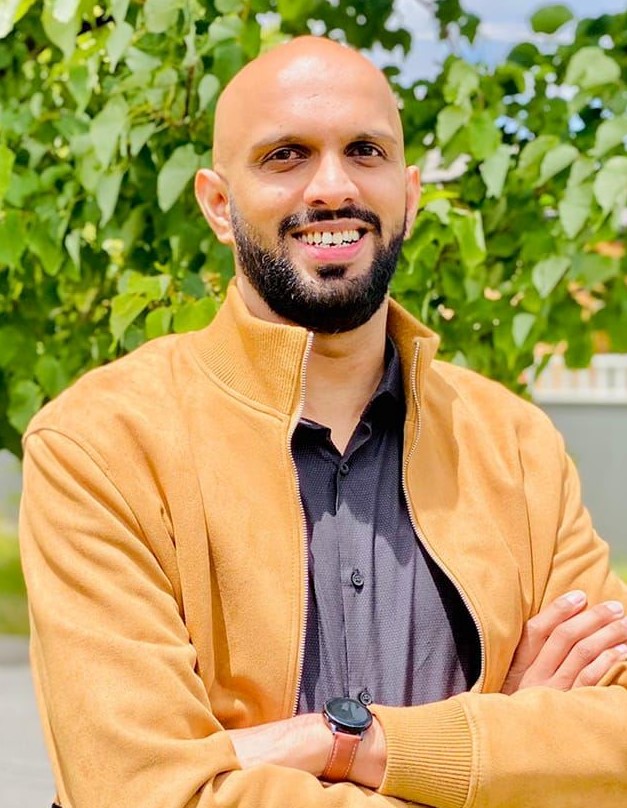}}]
{Prof. Muhammad Aamir Cheema} is Professor and Head of the Department of Software Systems and Cybersecurity at Monash University. He is also the founder of the Urban Computing Lab. He is an ACM Distinguished Speaker and the recipient of numerous prestigious awards, including the 2012 Malcolm Chaikin Prize for Research Excellence in Engineering, the 2013 ARC Discovery Early Career Researcher Award (DECRA), the 2014 Dean's Award for Excellence in Research by an Early Career Researcher, the 2018 ARC Future Fellowship, the 2018 Monash Student Association Teaching Award, the 2019 Victorian Young Tall Poppy Science Award, and 2025 Dean's Award for Equity, Diversity and Inclusion (EDI). His research has received multiple best paper recognitions, including two CiSRA Best Research Paper Awards (2009 and 2010), two "One of the Best Papers of ICDE" awards (2010 and 2012), the Best Vision Paper Award at ACM SIGSPATIAL 2024, and Best Paper Awards at ICAPS 2020, WISE 2013, and ADC 2010. 
\end{IEEEbiography}

\begin{IEEEbiography}
[{\includegraphics[width=1in,height=1.25in,clip,keepaspectratio]
{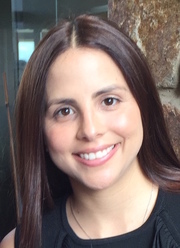}}]
{Dr. Maria Rodriguez Read} is a Lecturer in the School of Computing and Information Systems at the University of Melbourne. Her research focuses on intelligent and automated resource management for large-scale distributed computing environments, including cloud data centres, high-performance computing clusters, federated clouds, and edge infrastructures. Her work draws on optimisation, machine learning, and reinforcement learning to address challenges in the management of complex computing systems. She has published more than 50 peer-reviewed articles and regularly serves on technical program committees and as a reviewer for leading conferences and journals in distributed computing and cloud systems.
\end{IEEEbiography}

\begin{IEEEbiography}
[{\includegraphics[width=1in,height=1.25in,clip,keepaspectratio]
{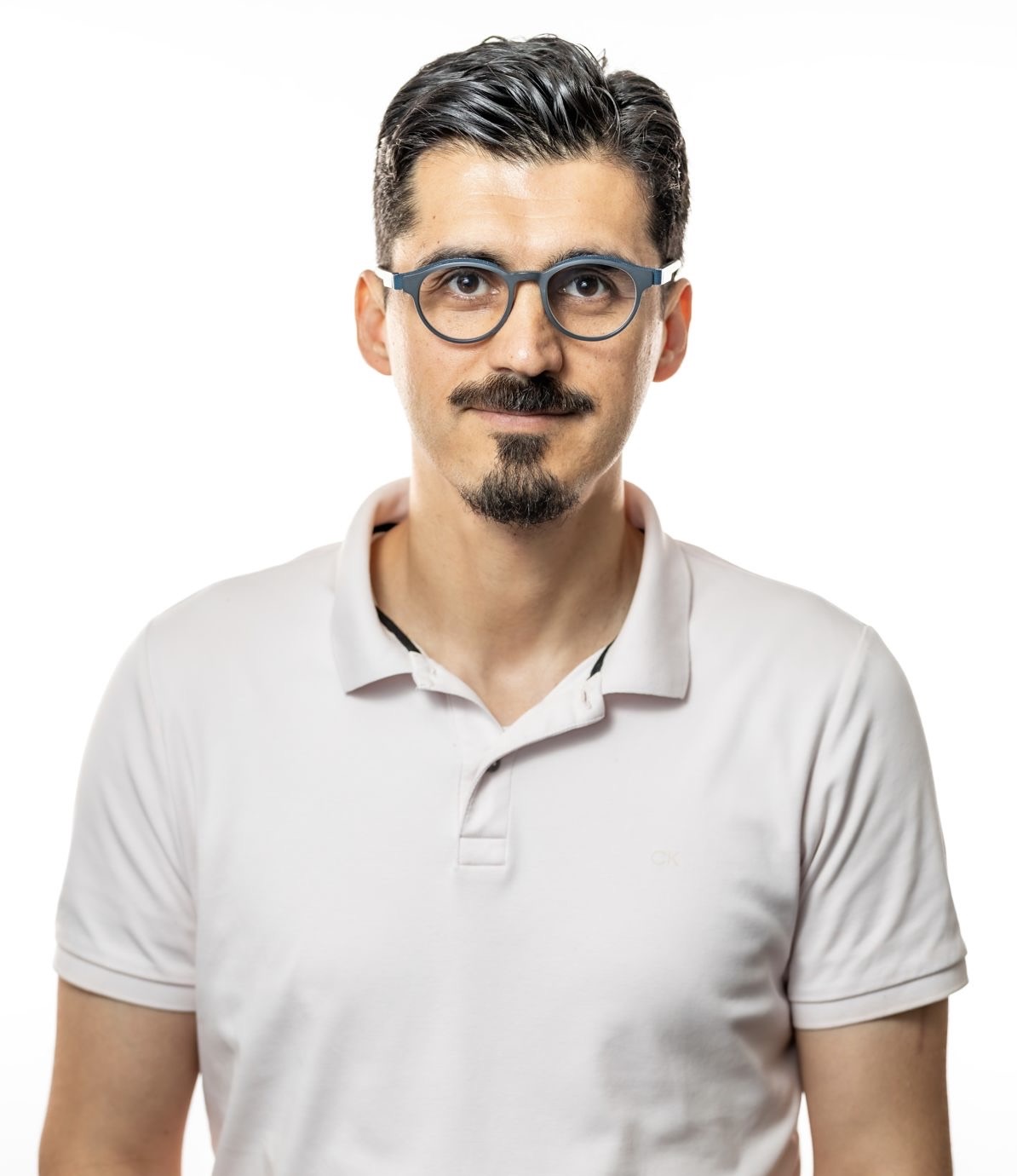}}]
{Dr. Adel N. Toosi} is an Associate Professor in Computer Systems at The University of Melbourne and Director of the Distributed Systems and Network Applications (DisNet) Laboratory. His research focuses on sustainable cloud, edge, and AI systems that reduce the environmental impact of digital infrastructure. He has published more than 90 peer-reviewed papers in leading venues, with over 7,000 citations and an h-index of 45. Dr. Toosi has developed widely used open-source research platforms, including iContinuum, WattEdge, Con-Pi, and AutoScaleSim, and has built real-world prototypes such as solar-powered micro data centres and renewable-driven agricultural systems. His work has received multiple international awards, including the ACM SIGSPATIAL 2024 Best Vision Paper Award and Best Paper awards at UCC and ICSOC. He has secured more than \$1.8 million in competitive research funding and serves on the editorial boards of Future Generation Computer Systems and IEEE Transactions on Services Computing. More information: \url{https://adelnadjarantoosi.info/}.
\end{IEEEbiography}

\end{document}